\pdfoutput=1

\documentclass[11pt]{article}

\usepackage[preprint]{acl}

\usepackage{times}
\usepackage{latexsym}

\usepackage[T1]{fontenc}

\usepackage[utf8]{inputenc}

\usepackage{microtype}

\usepackage{inconsolata}

\usepackage{graphicx}
\usepackage{xcolor}
\usepackage{subcaption}
\usepackage{booktabs}
\usepackage{multirow}
\usepackage{colortbl}
\title{Towards Building Large Scale Datasets and State-of-the-Art \\Automatic Speech Translation Systems for 14 Indian Languages}

\newif\ifcomments  
\commentstrue      

\ifcomments
    \newcommand{\todo}[1]{{\color{red} todo: #1}}
    \newcommand{\safi}[1]{{\color{teal} safi: #1}}
    \newcommand{\sparsh}[1]{{\color{purple} sparsh: #1}} 
    \newcommand{\sd}[1]{{\color{blue} SD: #1}}
    \newcommand{\mk}[1]{{\color{green} MK: #1}}
    \newcommand{\ak}[1]{{\color{orange} AK: #1}}    
    \newcommand{\raj}[1]{{\color{pink} Raj: #1}}
    \newcommand{\as}[1]{{\color{blue} AS: #1}}
\else
    \newcommand{\todo}[1]{}
    \newcommand{\safi}[1]{}
    \newcommand{\sparsh}[1]{}
    \newcommand{\sd}[1]{}
    \newcommand{\mk}[1]{}
    \newcommand{\ak}[1]{}    
    \newcommand{\raj}[1]{}
    \newcommand{\as}[1]{}
    
\fi

\author{
\textbf{Ashwin Sankar}$^{1,2}$\footnotemark[1] \quad
\textbf{Sparsh Jain}$^1$\thanks{~~Equal Contribution.} \quad
\textbf{Nikhil Narasimhan}$^{1}$ \quad
\textbf{Devilal Choudhary}$^4$\thanks{~~Work done during internship at AI4Bharat.} \\ 
\textbf{Dhairya Suman}$^{3}$\footnotemark[2] \quad
\textbf{Mohammed Safi Ur Rahman Khan}$^{1,2}$ \\
\textbf{Anoop Kunchukuttan}$^{1,5}$ \quad
\textbf{Mitesh M Khapra}$^{1,2}$ \quad
\textbf{Raj Dabre}$^{1,2,6,7}$\thanks{~~Work done while at NICT, Japan.}\thanks{~~Corresponding Author: \href{mailto:raj.dabre@cse.iitm.ac.in}{raj.dabre@cse.iitm.ac.in}}
  \\
    $^{1}$Nilekani Centre at AI4Bharat \quad
    $^{2}$Indian Institute of Technology, Madras \\
    $^{3}$Indian Institute of Technology, Delhi \quad $^{4}$Delhi Technological University \\
    $^{5}$Microsoft \quad
    $^{7}$Indian Institute of Technology, Bombay 
 \\
 \raisebox{-0.15em}{\includegraphics[height=0.9em]{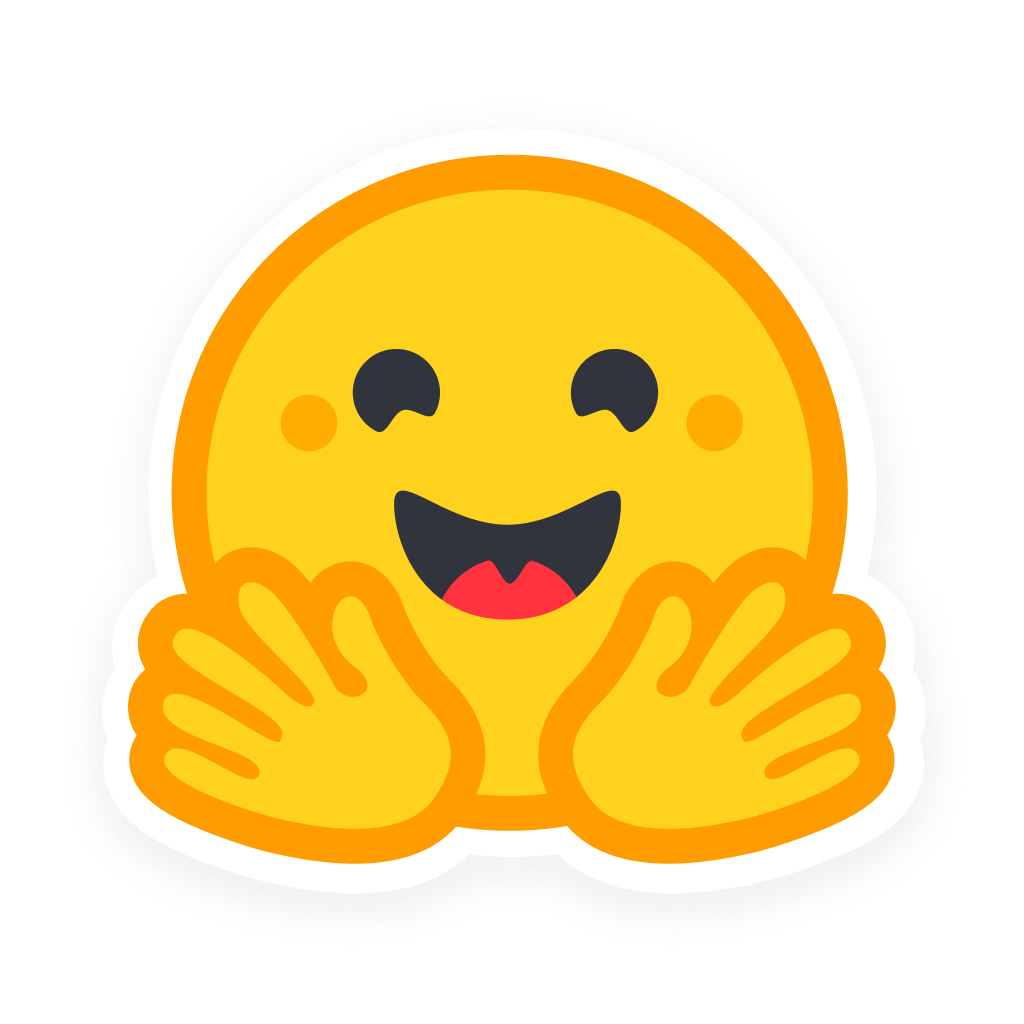}}~\small{\href{https://huggingface.co/collections/ai4bharat/bhasaanuvaad-672b3790b6470eab68b1cb87}{huggingface.co/BhasaAnuvaad}}
 \quad \quad
 \raisebox{-0.25em}{\includegraphics[height=1.1em]{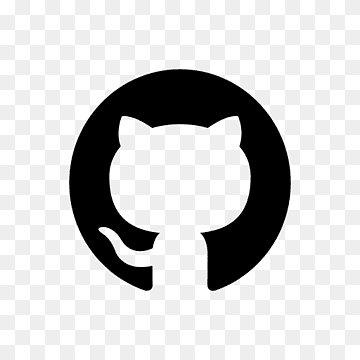}}~\small{\href{https://github.com/AI4Bharat/BhasaAnuvaad}{github.com/BhasaAnuvaad}}
}

\begin{document}
\newcommand{\fleurs}{\textsc{FLEURS}}
\newcommand{\flores}{\textsc{FLORES}}
\newcommand{\ourtestset}{\textsc{Indic-Spontaneous-Synth}}
\newcommand{\indicvoices}{\textsc{IndicVoices}}
\newcommand{\svarah}{\textsc{Svarah}}

\newcommand{\llamalarge}{\textsc{llama-3.1-405b-instruct}}
\newcommand{\seamless}{\textsc{SeamlessM4T}}
\newcommand{\azure}{\textsc{AZURE}}
\newcommand{\whisper}{\textsc{Whisper}}
\newcommand{\conformer}{\textsc{Indic-Conformer}}
\newcommand{\indicwhisper}{\textsc{Indic-Whisper}}
\newcommand{\indictrans}{\textsc{IndicTrans2}}

\newcommand{\dataset}{\textsc{BhasaAnuvaad}}

\definecolor{mygreen}{HTML}{d9ead3}
\definecolor{lightorange}{HTML}{fce5cd}
\definecolor{myorange}{HTML}{ffcd99}
\definecolor{myred}{HTML}{f4cccc}
\definecolor{mymagenta}{HTML}{ead1dc}
\definecolor{myblue}{HTML}{90cafe}
\definecolor{lightblue}{HTML}{cfe2f3}
\definecolor{mygray}{HTML}{efefef}
\definecolor{customgreen}{HTML}{57bb8a}
\definecolor{customyellow}{HTML}{deb53b}
\definecolor{customred}{HTML}{c43d3d}
\definecolor{customblue}{HTML}{3480b9}
\maketitle
\begin{abstract}
Speech translation for Indian languages remains a challenging task due to the scarcity of large-scale, publicly available datasets that capture the linguistic diversity and domain coverage essential for real-world applications. Existing datasets cover a fraction of Indian languages and lack the breadth needed to train robust models that generalize beyond curated benchmarks. To bridge this gap, we introduce \textsc{BhasaAnuvaad}, the largest speech translation dataset for Indian languages, spanning over $44$ thousand hours of audio and 17 million aligned text segments across 14 Indian languages and English. Our dataset is built through a threefold methodology: (a) aggregating high-quality existing sources, (b) large-scale web crawling to ensure linguistic and domain diversity, and (c) creating synthetic data to model real-world speech disfluencies. Leveraging \textsc{BhasaAnuvaad}, we train \textsc{IndicSeamless}, a state-of-the-art speech translation model for Indian languages that performs better than existing models. \textit{Our experiments demonstrate improvements in the translation quality, setting a new standard for Indian language speech translation.} We will release all the code, data and model weights in the open-source, with permissive licenses to promote accessibility and collaboration.

\end{abstract}

\section{Introduction}


Automatic Speech Translation (AST) has become crucial to break language barriers and enable communication across languages and cultures. Traditionally, speech translation systems have relied on cascaded architectures, where Automatic Speech Recognition (ASR) is followed by Machine Translation (MT). However, recent advances have led to more integrated end-to-end (E2E) models~\cite{babu2021xlsr, mms, whisper, communication2023seamlessm4t} that directly translate speech from one language into text in another. Additionally, Audio-LLM-based systems have emerged~\cite{wu2023decoderonly, chu2023qwenaudio, fathullah2023audiochatllama,gaido2024speechtranslationspeechfoundation} further advancing the field. 
Despite these advancements, progress has been concentrated on English and other European languages, while low- and mid-resource languages, including Indian languages, have remained largely underrepresented. The primary challenge is the lack of training data, which leads to suboptimal models.

\begin{figure*}[t]
    \centering    
    \includegraphics[width=0.8\textwidth]{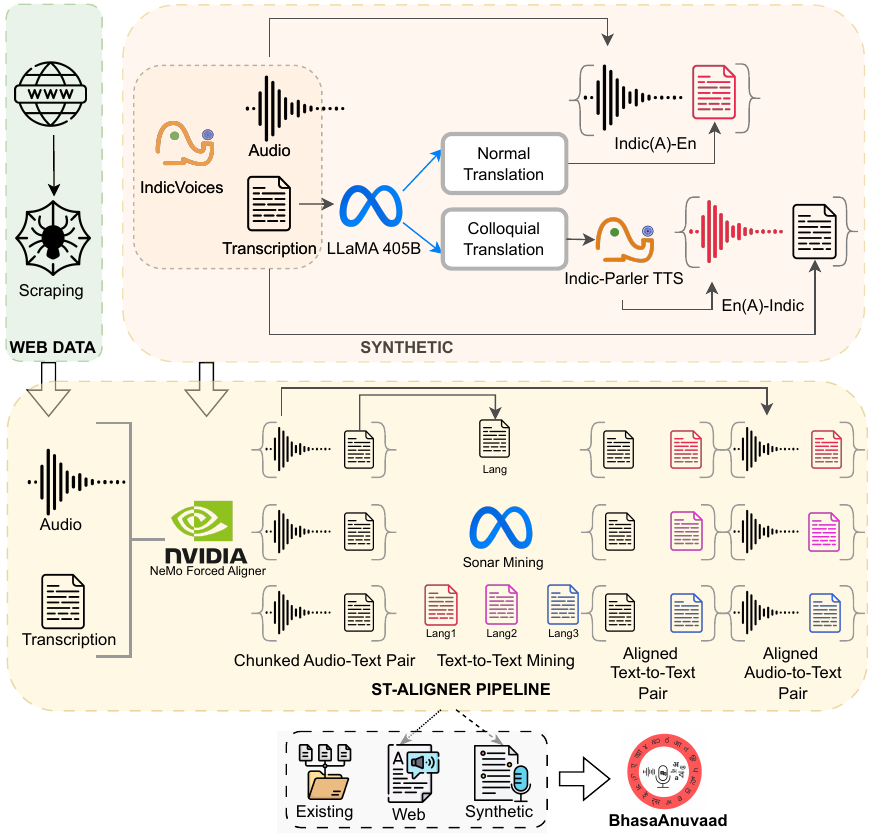}
    \caption{Overview of the creation of BhasaAnuvaad.}
    \label{fig:pipeline_overview}
\end{figure*}

India’s linguistic diversity, with 22 officially recognized languages and numerous dialects, presents significant challenges for speech translation, including diverse phonetic systems~\cite{mujadia-sharma-2023-towards}, frequent code-switching~\cite{shankar2024costa}, and syntactic variations. However, the primary bottleneck remains the scarcity of large-scale, high-quality datasets, aggravated by the limited web presence of Indian languages. To bridge this gap, we introduce \dataset~, a large-scale speech translation dataset comprising of over 44K hours of speech translation data covering 14 Indian languages alongside English, making it the most comprehensive speech translation resource for the region to date. The included languages - Assamese, Bengali, Gujarati, Hindi, Kannada, Malayalam, Marathi, Nepali, Odia, Punjabi, Sindi, Tamil, Telugu, and Urdu - span both medium and low resource languages. Our dataset construction follows a three-pronged approach to ensure high-quality and diverse data collection: \textbf{(1) Aggregating High-Quality Existing Sources:} We incorporate publicly available speech-text pairs from prior works, refining them with improved alignment techniques to enhance usability for speech translation tasks. \textbf{(2) Large-Scale Web Crawling:} To expand linguistic and domain coverage, we employ an automated web-mining pipeline that we call ST-ALIGNER (Figure~\ref{fig:pipeline_overview}) to extract speech-text pairs from diverse online sources, including government portals, educational materials, spiritual discourses, and multilingual podcasts. This enables us to capture naturally occurring speech across varied contexts, improving real-world applicability. \textbf{(3) Synthetic Data Generation:} To address gaps in speech diversity, we leverage the \textsc{IndicVoices} ASR dataset and use \textsc{Llama-3.1-405B-Instruct} for synthetic translations into 13 languages, while generating English speech using a TTS model~\cite{sankar2025parlertts}.

We also build \textsc{BhasaAnuvaad-test} by randomly sampling 20 minutes of data for each language pair from \dataset. Additionally, we introduce \textsc{IndicSeamless}, a finetuned \textsc{SeamlessM4T-V2-Large} model on \dataset~ that surpasses existing AST setups - both cascaded and end-to-end - on established benchmark. We observe an overall average improvement of $10$ chrf++ scores over previous approaches on \textsc{BhasaAnuvaad-test} demonstrating the efficacy of \dataset.

In summary, our contributions are: \textbf{(i)} the release of \dataset, the \textit{largest} Indic-language speech translation corpus, comprising over 44 thousand hours of data across 14 Indian languages and English, \textbf{(ii)} a speech-translation data mining pipeline, ST-ALIGNER \textbf{(iii)} an \textit{evaluation} of popular AST baselines on~\citet{fleurs} and \textsc{BhasaAnuvaad-test}, a test split from \dataset~, and \textbf{(iv)} the release of state-of-the-art speech translation model for 14 Indian languages.  All code and datasets developed as part of this work will be made publicly available to support future research in AST for Indic languages.

\section{Related Work}
\noindent\textbf{Cascaded and End-to-End AST. } Cascaded Automatic Speech Translation (AST) has been explored extensively in early research~\cite{di-gangi-etal-2019-robust, cheng-etal-2019-breaking, bahar-etal-2020-start}, connecting Automatic Speech Recognition (ASR) and Machine Translation (MT) systems in sequence. These systems suffer from cascaded error propagation~\cite{sperber-paulik-2020-speech, beck-etal-2019-neural}, loss of prosodic information~\cite{bentivogli2021cascade}, and increased computational complexity~\cite{lee-etal-2022-direct}. In contrast, end-to-end (E2E) models can avoid this~\cite{weiss17_interspeech, 9415093}, but suffer from lack of good quality data~\cite{sethiya2023endtoend}. With the rise of modern Large Language Models (LLMs), recent works have started exploring the feasibility of using Audio-based LLMs (or Speech LLMs), which combine Speech Foundation models (SFMs) and LLMs~\cite{fathullah2023audiochatllama, wu2023decoderonly, huang2023speech,gaido2024speechtranslationspeechfoundation} .

\noindent \textbf{Datasets for AST.} The collection of speech-translation datasets for training end-to-end models has proven to be a persistent challenge. While efforts like SpeechMatrix~\cite{duquenne2022speechmatrix} and SeamlessAlign~\cite{communication2023seamlessm4t} attempt to mine data from the web, the resulting datasets are often noisy. Most research has focused on curating data using comparable corpora from domains like Education~\cite{duriskova2024khan, song2019coursera}, TED talks~\cite{salesky2021multilingual}, Parliament Debates~\cite{koehn-2005-europarl}, among others. Additionally, with the rise of generative models, recent works have also started exploring the generation of large amounts of synthetic speech translation data in a flexible and cost-effective manner~\cite{ye2023gigast10000hourpseudospeech},~\cite{bamfo-odoom-etal-2024-synthetic}. Although the use of synthetic data has been widely studied in text-based machine translation~\cite{sennrich-etal-2016-improving, gala2023indictrans2}, its application in Speech translation, particularly for low-resource languages like Indic languages, remains underexplored, a gap which we fill.

\noindent \textbf{Indian Languages AST.} Unlike English, the shortage of parallel datasets has hindered the progress in Indian Languages~\cite{mujadia-sharma-2023-towards, mhaskar2023vaktasetu}. Multilingual speech translation models like \textsc{mSLAM}~\cite{bapna2022mslam}, \textsc{Whisper}~\cite{whisper}, \textsc{Seamless}~\cite{communication2023seamlessm4t} have made strides in addressing this gap, but their effectiveness in low-resource and real-world spontaneous speech settings remains largely unexplored. 
Furthermore, speech-based benchmarks for Indian languages are still underdeveloped~\cite{sethiya2023endtoend}, with \textsc{Fleurs}~\cite{fleurs} being the primary option. In this paper, we push the boundaries of AST for Indian languages with the help of the large-scale dataset we construct.

\section{\textsc{BhasaAnuvaad}}

\textsc{BhasaAnuvaad} is the largest spoken translation dataset for any Indian language, totaling approximately 44,400 hours of speech and text data. Its construction involved a multi-step approach, as illustrated in Figure \ref{fig:pipeline_overview}, combining (i) Aggregating existing datasets, (ii) Mining parallel speech and text data from comparable sources, and (iii) Synthetically generated speech/text parallel data. A language wise composition of this dataset is illustrated in Figure \ref{fig:radar_plot}. A detailed breakdown of language and dataset-wise statistics is provided in Appendix \ref{sec:data_stats} in Tables~\ref{tab:en-in} \& \ref{tab:in-en}.

\subsection{Aggregating Existing Datasets} We begin by aggregating several existing datasets in Indian Languages, which form the foundation of \textsc{BhasaAnuvaad}. These include:

\begin{figure*}[!t]
    \centering
    \includegraphics[width=\linewidth]{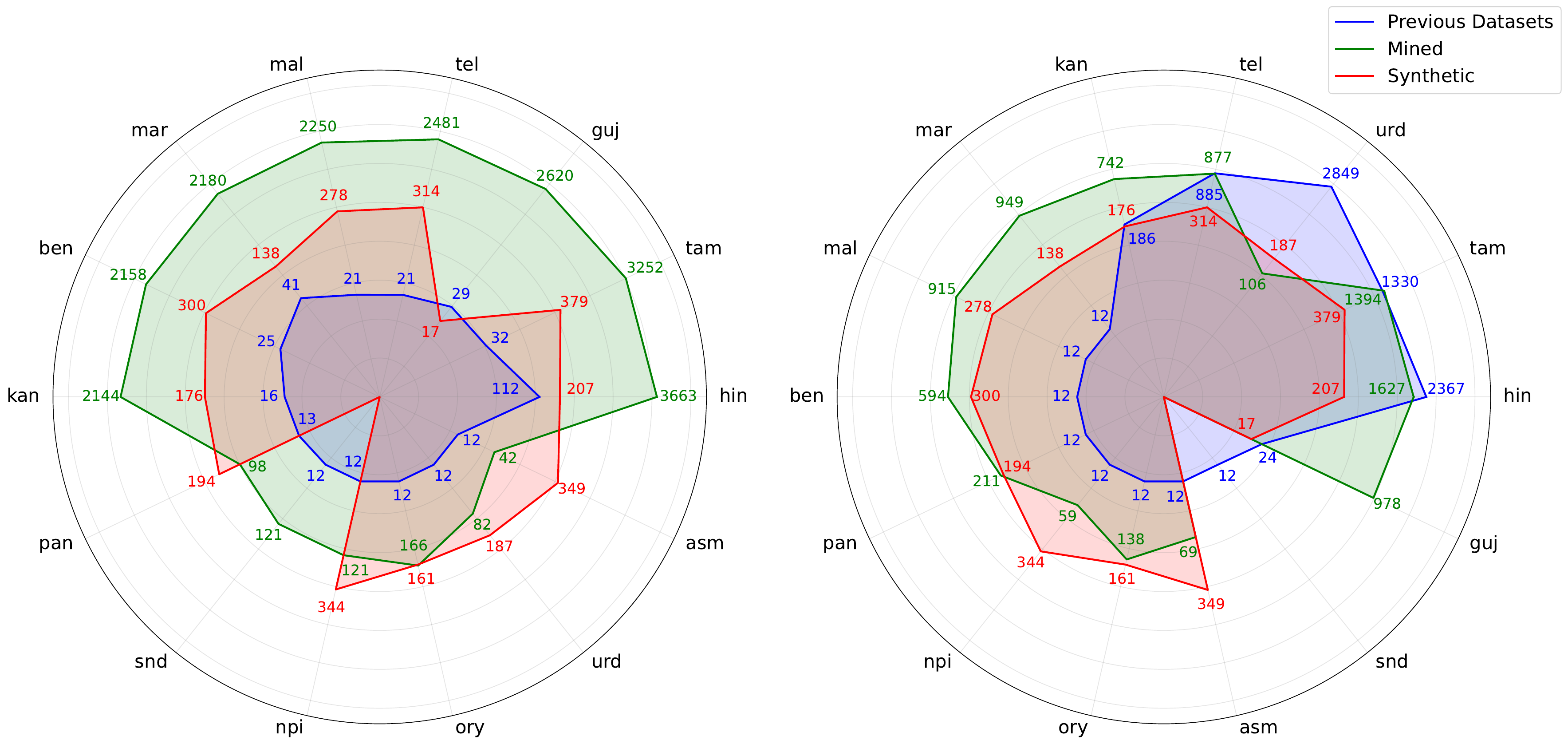}
    \caption{ Distribution of dataset hours in log scale for En-Indic (left) and Indic-En (right) language pairs across three data sources: Previous Datasets (blue), Curated \& Mined (green), and Synthetic (red).}
    \label{fig:radar_plot}
\end{figure*}

\noindent \textbf{Indic-TEDST}: The Indic-TEDST~\cite{sethiya2024indic} dataset includes the audio, transcripts, and translations of popular TED Talks in nine widely spoken Indian languages - Bengali, Gujarati, Hindi, Kannada, Malayalam, Marathi, Punjabi, Tamil, and Telugu - totaling approximately 203 hours of English-to-Indic data.

\noindent \textbf{FLEURS}: The FLEURS-train dataset~\cite{fleurs} is a multilingual speech corpus covering over 100 languages for training automatic speech recognition (ASR) and speech translation (AST) models. FLEURS was previously the largest source of data, covering 14 Indian languages, making it a valuable resource.

\noindent \textbf{CVSS}: The CVSS~\cite{jia2022cvss} corpus is a multilingual speech-to-speech translation dataset derived from CoVoST 2~\cite{wang2020covost2massivelymultilingual}, which is sampled from the Common Voice dataset. Spanning 21 languages, it includes only one Indian language, Tamil, with approximately 3 hours of data.

\noindent \textbf{Khan Academy Corpus}: The Khan Academy corpus~\cite{duriskova2024khan} provides data for the education domain in 29 languages. It includes 3 Indian languages, with 105 hours of audio.

\noindent \textbf{SeamlessAlign:} 
We recreate the SeamlessAlign data introduced by \citet{communication2023seamlessm4t} to get around 7500 hours of data for 5 Indic languages.

\subsection{Mining from comparable sources}
We explore the extraction of parallel speech-text data from various online sources containing high-quality content across multiple domains. While many sources provide transcripts in different languages, these are often misaligned with the corresponding audio. To address this issue, we develop an automated pipeline that aligns transcripts with their corresponding source audios, to produce high-quality parallel speech-text data.

\subsubsection{ST-Aligner}
\label{sec:st_aligner}
Our pipeline consists of four stages: normalization, punctuation restoration, audio-transcript alignment, and parallel text alignment.

\noindent \textbf{1. Normalization:} To ensure compatibility, all audio files are first converted to a mono channel and resampled at 16 kHz, as ASR models operate on 16 kHz audio. The correspoding transcripts are then cleaned by removing commonly found noise like extraneous symbols, extra spaces, and line breaks. We also replace the terminal punctuation marks with a non-standard character to aid sentence segmentation during alignment.

\noindent \textbf{2. Punctuation Restoration:} Many subtitles and transcripts lack proper punctuation, which can hinder accurate alignment. We conducted a small experiment with indic-punct~\cite{gupta2022indicpunctautomaticpunctuationrestoration}, a punctuation restoration model, and found that it did not perform well in spoken contexts. Given this limitation, we opt for the \textsc{Llama-3.1-405B-Instruct} model, which demonstrated superior performance in restoring punctuation and effectively structuring the text for alignment (Figure \ref{fig:Prompt_Punctuation}).

\noindent \textbf{3. Audio-Transcript Alignment:} The cleaned transcripts and audio are processed using the Nemo Forced Aligner (NFA)\footnote{\url{https://github.com/NVIDIA/NeMo/tree/main/tools/nemo_forced_aligner}}, which generates token-level, word-level, and segment-level timestamps based on CTC-based ASR models. Given the linguistic diversity of our dataset, we utilize \textsc{IndicConformer} ASR models for Indian languages and Nvidia FastConformer\footnote{\url{https://catalog.ngc.nvidia.com/orgs/nvidia/teams/nemo/models/stt_en_fastconformer_hybrid_large_pc}} for English, ensuring that the alignment is optimized for the phonetic and acoustic characteristics of each language.

Long-form audio presents additional challenges, as ASR models and forced aligners struggle with excessive sequence lengths, often resulting in degraded alignment accuracy. To address this, we employ Silero Voice Activity Detection (Silero VAD)~\cite{sileroVAD}, a lightweight yet highly effective neural model that detects speech boundaries and segments long recordings into smaller, more manageable chunks. This pre-segmentation step not only improves computational efficiency but also enhances alignment precision by reducing errors that arise from excessive length mismatches between audio and transcript.

\noindent \textbf{4. Parallel Text Alignment:} Once audio segments are paired with transcripts, we align them with their corresponding target language text. We generate sentence embeddings for both the source and target texts using SONAR~\cite{Duquenne:2023:sonar_arxiv}. Next, we compute cosine similarity scores between these embeddings to identify the most likely parallel sentence pairs. 

\subsubsection{Sources of Mining Corpora}
\label{sec:web_mined}

We apply the pipeline described in Section~\ref{sec:st_aligner} on the following sources to extract translation pairs.

\noindent \textbf{Spoken Tutorial:} Spoken tutorial, is an initiative under the National Mission on Education under the Ministry of Human Resource Development, Government of India to enable the translation of various vocational educational content in different languages. This dataset covers Indic-En translations with 499.04 hours of educational tutorials in 12 Indian languages.

\noindent \textbf{UGCE Resources:} The University Grants Commission translates a wide range of university-level courses into multiple Indian languages. Originally delivered in English (En audio), these courses have their transcripts translated into 8 additional languages, resulting in over 430 hours of aligned audio-text data for each language pair.

\noindent \textbf{VaaniPedia:} Vaanipedia is a comprehensive collection of religious and spiritual audio-text content in 14 Indian languages, comprising over 121 hours of aligned audio shared across each language pair.

\noindent \textbf{Mann ki Baat:} Mann Ki Baat is a monthly radio broadcast by the Prime Minister of India that addresses various national issues. Each episode is transcribed and manually translated into 13 Indian languages. These translations are also manually recorded, resulting in over 435 hours of speech data across all language pairs, covering both Indic-to-English and English-to-Indic directions. 
\textit{Notably, Mann Ki Baat is the only source in this dataset that includes Manipuri, which is provided in Bengali script (rather than the Meitei script). While tables and figures do not explicitly highlight it, we are releasing the Manipuri data in Bengali script as sourced from Mann Ki Baat.}

\noindent \textbf{WordProject:} This dataset consists of Bible audiobooks sourced from the Word Project website in 12 Indian languages, totaling approximately 1,612 hours of speech data across both Indic-En and En-Indic directions.

\noindent \textbf{NPTEL:} The National Programme on Technology Enhanced Learning (NPTEL) is an Indian e-learning platform for university-level science, technology, engineering, and mathematics (STEM) subjects. It is the largest e-repository in the world, offering courses in engineering, basic sciences, and selected humanities and management subjects\footnote{\url{https://en.wikipedia.org/wiki/National_Programme_on_Technology_Enhanced_Learning}}.  These courses are primarily recorded in English, but they are transcribed and re-recorded in multiple languages manually.  This is the largest component of our dataset, contributing over 22,500 hours of spoken content across 10 Indian languages. It covers a wide range of STEM subjects and includes both En-XX and XX-En translations, making it one of the most significant sources in our collection.

While these datasets are high quality, they are not diverse with majority of the content being from Educational domain. Furthermore, the audio styles they cover lack the diversity necessary for models to adapt effectively to real-world, where background noise, spontaneous speech, and varying accents are common. To overcome these limitations, we augment our dataset by using synthetic data generation as described below.

\subsection{Synthetic Data Generation}
\label{sec:synthetic}
To ensure that we have audio diversity across different domains, demographics, and languages, we use \textsc{IndicVoices}~\cite{javed-etal-2024-indicvoices}
which is the largest collection of natural and spontaneous speech for Indian languages containing over 4,000 hours of transcribed speech.  Its extensive language diversity makes it well-suited for training models capable of performing effectively in real-world scenarios.


\noindent \textbf{Synthetic Translations:} The \textsc{IndicVoices} dataset consists of spontaneous speech, which includes disfluencies, repetitions, and informal constructions that pose challenges for direct text-based translation. Since \textsc{IndicVoices} also provides normalized transcripts, where such artifacts are already addressed, we use these instead of verbatim transcripts for translation. This ensures cleaner and more structured inputs while retaining the essence of conversational speech. 

To generate high-quality English translations, we first apply punctuation restoration and translate the unsegmented version of the transcripts using \textsc{Llama-3.1-405B-Instruct} (Figure \ref{fig:Prompt_Punctuation}). By pairing the original Indic audio with these translated transcripts, we construct XX-En pairs suitable for speech translation tasks. We then utilize \textsc{ST-Aligner} to align the translations at the segment level, preserving synchronization between the audio and textual representations. 


\begin{figure*}
    \begin{minipage}{\textwidth}
        \centering
        \begin{subfigure}[b]{0.48\textwidth}
            \centering
            \includegraphics[width=\textwidth]{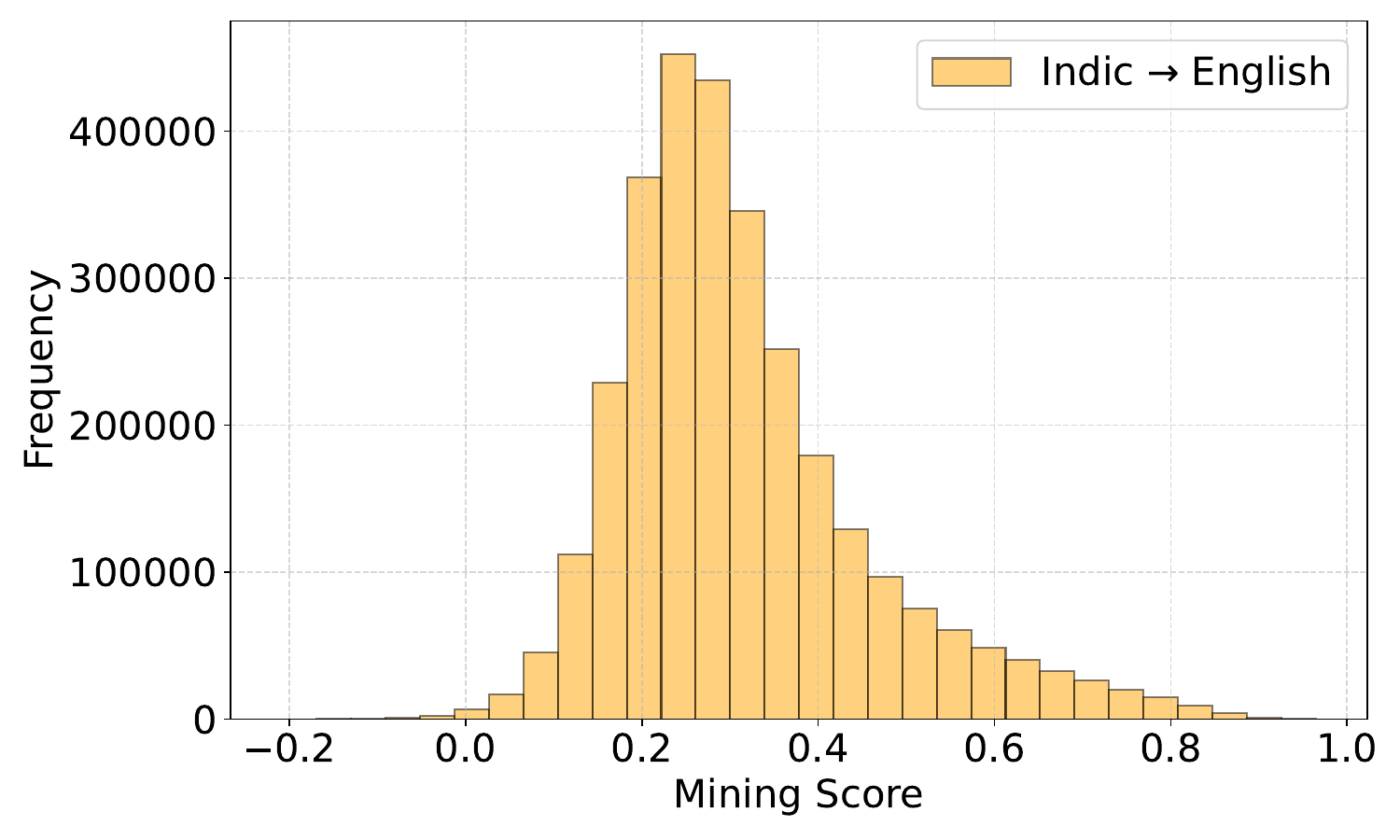}
            \caption{Seamless Align}
            \label{fig:seamless_urdu}
        \end{subfigure}
        \hfill
        \begin{subfigure}[b]{0.48   \textwidth}
            \centering
            \includegraphics[width=\textwidth]{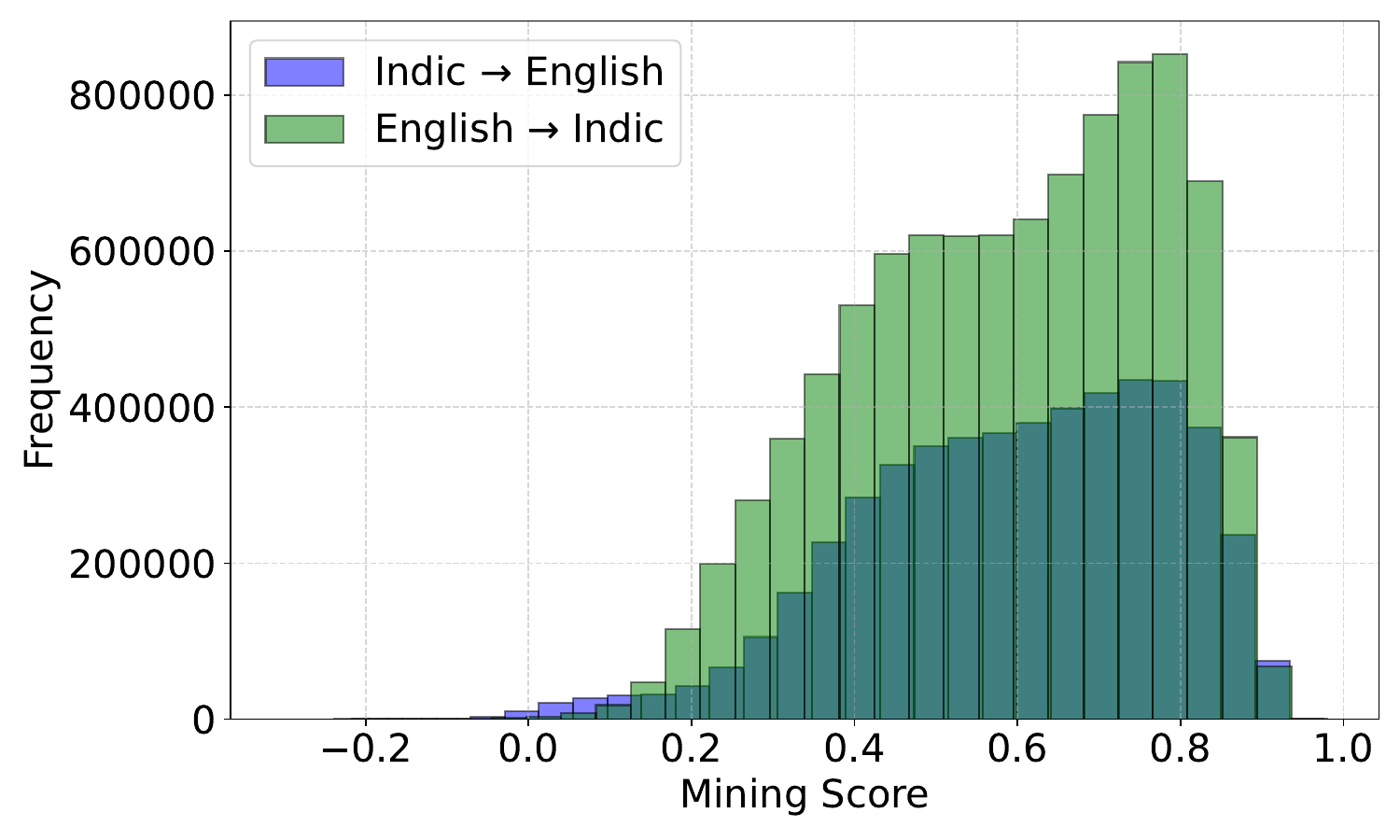}
            \caption{\dataset}
            \label{fig:dataset_urdu}
        \end{subfigure}
    \end{minipage}
    \caption{Distribution of SONAR mining scores for XX $\rightarrow$ En sentence pairs from the SeamlessAlign (left) and \dataset~ (right) datasets. \dataset~ consistently exhibits a higher concentration of high-quality alignments, with mining scores skewed toward the upper end of the scale. In contrast, SeamlessAlign distributions are centered around lower mining scores, indicating overall lower alignment quality. This comparison underscores the improved precision of alignments in \dataset~.}
    \label{fig:mining_scores_comparison}
\end{figure*}

Unlike specialized translation models such as \textsc{IndicTrans2}~\cite{gala2023indictrans2}, which are optimized for formal text and often struggle with colloquial or spontaneous speech, \textsc{Llama-3.1-405B-Instruct} offers a \textit{steerable} alternative. Its instruction-following capability enables more flexible adaptation to informal and conversational registers through prompting with examples and contextual cues.

To validate this, we conducted a small-scale internal human evaluation on a subset of the \textsc{IndicVoices} dataset. \textsc{Llama-3.1-405B-Instruct} consistently outperformed \textsc{IndicTrans2}, particularly in preserving the nuances and fluency of informal spoken language. This makes it well-suited for real-world applications, where such variability is common. In a broader multilingual evaluation, we assessed the translation quality of \textsc{Llama-3.1-405B-Instruct} across 1,000 documents per language. The model achieved an average human-rated score of 8 out of 10, indicating adequacy for our use case.

\noindent \textbf{Synthetic Speech Generation:}
To overcome the time and cost limitations of manual voice recordings, we augment our dataset synthetic speech generation. We begin with verbatim transcripts in \textsc{IndicVoices}, which preserve natural speech phenomena such as disfluencies, false starts, repetitions, and pauses. These are translated into colloquial English using the \textsc{Llama-3.1-405B-Instruct} model, producing outputs that reflect spontaneous conversational patterns, including hesitations and filler words (Figure~\ref{fig:colloquial_translation}).

To synthesize English speech with prosodic and pronunciation features characteristic of Indian-accented speakers, we use \textsc{Indic Parler TTS}~\cite{sankar2025parlertts}, a state-of-the-art text-to-speech model fine-tuned for Indian languages. The model offers fine-grained prosodic control, enabling realistic variation in rhythm, intonation, and articulation.

By pairing the synthesized English speech with the normalized Indic transcripts from \textsc{IndicVoices}, we construct a high-quality En-XX dataset. This resource enhances both linguistic diversity and domain robustness, supporting more effective training and evaluation for real-world speech translation applications.

\subsection{Quality Control}
\label{sec:quality_control}

To ensure the reliability of the aligned speech-text pairs, we adopt a two-fold evaluation strategy; one for audio-text alignment quality and the other for translation quality.

\noindent \textbf{Alignment Quality:}
We quantify the accuracy of audio-transcript alignment by computing Levenshtein distance-based similarity scores between the cleaned reference transcripts and ASR-predicted outputs, following the methodology of~\citet{kaushalShrutilipi}. This metric enables us to identify and filter out poorly aligned segments, thus maintaining high fidelity in timestamped pairs that are critical for training downstream speech and translation models.

\noindent \textbf{Translation Quality:}
To evaluate the quality of the aligned parallel text, we compute cosine similarity scores between sentence embeddings of the source transcript and its corresponding translation. These embeddings are obtained using the SONAR model~\cite{Duquenne:2023:sonar_arxiv}, which provides robust cross-lingual sentence representations. 
\emph{While the dataset is not filtered based on these metrics in the public release, we include both the alignment and translation quality scores alongside the data. This allows users to apply their own thresholds or filtering strategies depending on the specific needs of their applications.}

\noindent \textbf{Comparison with Prior Web-Mined Corpora:}
Figure~\ref{fig:mining_scores_comparison} compares SONAR-based mining scores for XX~$\rightarrow$En sentence pairs from \textsc{BhasaAnuvaad} and the SeamlessAlign corpus. SeamlessAlign shows a concentration of low scores, indicating noisy or weak alignments, whereas \textsc{BhasaAnuvaad} yields a markedly higher proportion of high-quality pairs, with scores skewed toward the upper end of the distribution. Language-wise alignment score distributions are shown in Figures~\ref{fig:lang_dist_1} and~\ref{fig:lang_dist_2}, with comprehensive plots for all languages provided in Figures~\ref{fig:mining_histogram_ba_1}, \ref{fig:mining_histogram_ba_2}, and \ref{fig:mining_histogram_ba_3}. These results underscore the effectiveness of our mining pipeline in producing cleaner and more semantically aligned sentence pairs.

\section{Speech Translation Models}
In this study, we primarily consider the Unified (or end-to-end) and the Cascaded Speech translation paradigms across various models. While recent advancements in Audio-LLMs (or SLLMs) introduce new paradigms by integrating Speech Foundation Models with LLMs for Automatic Speech Translation (AST), we do not include them in this study due to the lack of support for Indian languages.

\setlength\fboxsep{1pt}

\subsection{Our Model}

\noindent \colorbox{lightblue}{\seamless~(SD\textsubscript{BA})}{\raisebox{-0.4em}{\includegraphics[height=1.4em]{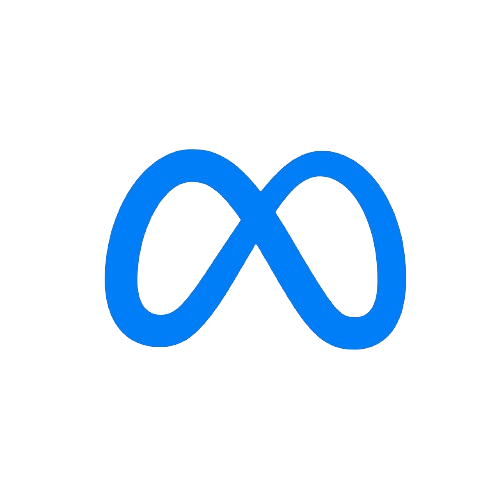}}} We fine-tune the \textsc{SeamlessM4T-v2-large} model on a filtered subset of \dataset~(SD\textsubscript{BA}; IndicSeamless), using a high-quality filtered subset to ensure effective adaptation to Indian languages. Training is conducted on 8 × A100 GPUs for 430{,}000 steps with an effective per-device batch size of 64. We employ the Adam 8-bit optimizer with a learning rate of $1 \times 10^{-5}$ and 1{,}000 warmup steps. To mitigate overfitting, we apply early stopping with a patience of 10.

\begin{table*}[!t]
\centering
\fontsize{8pt}{10pt}\selectfont
\centering
\setlength{\tabcolsep}{6pt} 
\renewcommand{\arraystretch}{1} 
\begin{tabular}{@{}ccccccccccccc@{}}
\toprule
\cellcolor[HTML]{FFFFFF}{\color[HTML]{31333F} }                                & \multicolumn{6}{c}{\textbf{Direct}}                                                                                                                                                                                                                      & \multicolumn{6}{c}{\textbf{Cascaded}}                                                                                                                                                         \\ \cmidrule(lr){2-7} \cmidrule(lr){8-13}
\cellcolor[HTML]{FFFFFF}{\color[HTML]{31333F} }                                & \multicolumn{2}{c}{\textbf{SD}}                                                                         & \multicolumn{2}{c}{\textbf{SD\textsubscript{BA}}}                           & \multicolumn{2}{c}{\textbf{AZURE}}                                           & \multicolumn{2}{c}{\textbf{SC}}                               & \multicolumn{2}{c}{\textbf{W + IT2}}                          & \multicolumn{2}{c}{\textbf{SC + IT2}}                         \\ \cmidrule(lr){2-3} \cmidrule(lr){4-5} \cmidrule(lr){6-7} \cmidrule(lr){8-9} \cmidrule(lr){10-11} \cmidrule(lr){12-13}
\multirow{-3}{*}{\cellcolor[HTML]{FFFFFF}{\color[HTML]{31333F} \textit{lang}}} & \textbf{FL}                                     & \textbf{BA}                                         & \textbf{FL}               & \textbf{BA}                   & \textbf{FL}                              & \textbf{BA}                   & \textbf{FL}               & \textbf{BA}                   & \textbf{FL}               & \textbf{BA}                   & \textbf{FL}               & \textbf{BA}                   \\ \midrule
\cellcolor[HTML]{FFFFFF}{\color[HTML]{0D0D0D} \textit{asm}}                    & \cellcolor[HTML]{F6FBF9}{\color[HTML]{0D0D0D} 37.90} & \cellcolor[HTML]{FFFFFF}{\color[HTML]{0D0D0D} 32.60} & \cellcolor[HTML]{EDF8F3}39.30 & \cellcolor[HTML]{FFEEBE}42.20 & \cellcolor[HTML]{F6FBF9}37.90 & \cellcolor[HTML]{FFFCF3}36.40 & \cellcolor[HTML]{FFFFFF}36.20 & \cellcolor[HTML]{FFFFFD}35.30 & \cellcolor[HTML]{E5F5ED}40.80 & \cellcolor[HTML]{FFEBB3}43.40 & \cellcolor[HTML]{E6F5EE}40.50 & \cellcolor[HTML]{FFF3D2}40.00 \\
\cellcolor[HTML]{FFFFFF}{\color[HTML]{0D0D0D} \textit{ben}}                    & \cellcolor[HTML]{BEE5D2}{\color[HTML]{0D0D0D} 47.50} & \cellcolor[HTML]{FFFFFF}{\color[HTML]{0D0D0D} 31.00} & \cellcolor[HTML]{B5E2CC}48.90 & \cellcolor[HTML]{FFECB6}43.10 & \cellcolor[HTML]{C1E6D4}47.00 & \cellcolor[HTML]{FFFCF4}36.30 & \cellcolor[HTML]{C4E7D6}46.40 & \cellcolor[HTML]{FFFBEE}36.90 & \cellcolor[HTML]{A9DDC3}51.00 & \cellcolor[HTML]{FFEDBC}42.50 & \cellcolor[HTML]{ADDEC6}50.40 & \cellcolor[HTML]{FFF6DD}38.80 \\
\cellcolor[HTML]{FFFFFF}{\color[HTML]{0D0D0D} \textit{guj}}                    & \cellcolor[HTML]{B5E2CC}{\color[HTML]{0D0D0D} 48.90} & \cellcolor[HTML]{FFFFFF}{\color[HTML]{0D0D0D} 33.80} & \cellcolor[HTML]{AEDFC7}50.10 & \cellcolor[HTML]{FFE59C}46.00 & \cellcolor[HTML]{B7E2CD}48.60 & \cellcolor[HTML]{FFF8E3}38.20 & \cellcolor[HTML]{B8E3CE}48.40 & \cellcolor[HTML]{FFF7E0}38.50 & \cellcolor[HTML]{A3DABF}52.10 & \cellcolor[HTML]{FFE59E}45.80 & \cellcolor[HTML]{A5DBC1}51.70 & \cellcolor[HTML]{FFF2CC}40.70 \\
\cellcolor[HTML]{FFFFFF}{\color[HTML]{0D0D0D} \textit{hin}}                    & \cellcolor[HTML]{96D5B6}{\color[HTML]{0D0D0D} 54.20} & \cellcolor[HTML]{FFFFFF}{\color[HTML]{0D0D0D} 25.00} & \cellcolor[HTML]{91D3B3}55.10 & \cellcolor[HTML]{FFD666}51.90 & \cellcolor[HTML]{9DD8BB}53.10 & \cellcolor[HTML]{FFE6A1}45.40 & \cellcolor[HTML]{96D5B6}54.20 & \cellcolor[HTML]{FFE6A0}45.60 & \cellcolor[HTML]{8CD1AF}56.00 & \cellcolor[HTML]{FFDB79}49.90 & \cellcolor[HTML]{8ED1B0}55.70 & \cellcolor[HTML]{FFE8A7}44.80 \\
\cellcolor[HTML]{FFFFFF}{\color[HTML]{0D0D0D} \textit{kan}}                    & \cellcolor[HTML]{B5E2CC}{\color[HTML]{0D0D0D} 48.90} & \cellcolor[HTML]{FFF8E4}{\color[HTML]{0D0D0D} 38.00} & \cellcolor[HTML]{B0DFC8}49.90 & \cellcolor[HTML]{FFDA75}50.30 & \cellcolor[HTML]{B5E2CC}48.90 & \cellcolor[HTML]{FFF0C6}41.40 & \cellcolor[HTML]{BCE4D0}47.80 & \cellcolor[HTML]{FFEBB3}43.40 & \cellcolor[HTML]{A3DABF}52.10 & \cellcolor[HTML]{FFDA72}50.60 & \cellcolor[HTML]{A7DCC2}51.40 & \cellcolor[HTML]{FFE6A0}45.60 \\
\cellcolor[HTML]{FFFFFF}{\color[HTML]{0D0D0D} \textit{mal}}                    & \cellcolor[HTML]{B5E1CC}{\color[HTML]{0D0D0D} 49.00} & \cellcolor[HTML]{FFF5D8}{\color[HTML]{0D0D0D} 39.40} & \cellcolor[HTML]{A8DCC3}51.20 & \cellcolor[HTML]{FFD970}50.80 & \cellcolor[HTML]{ADDEC6}50.30 & \cellcolor[HTML]{FFEAB0}43.80 & \cellcolor[HTML]{BBE4D0}47.90 & \cellcolor[HTML]{FFE9AC}44.20 & \cellcolor[HTML]{94D4B4}54.70 & \cellcolor[HTML]{FFDC7B}49.60 & \cellcolor[HTML]{98D6B8}53.90 & \cellcolor[HTML]{FFE18D}47.60 \\
\cellcolor[HTML]{FFFFFF}{\color[HTML]{0D0D0D} \textit{mar}}                    & \cellcolor[HTML]{D1EDDF}{\color[HTML]{0D0D0D} 44.10} & \cellcolor[HTML]{FFFFFF}{\color[HTML]{0D0D0D} 29.30} & \cellcolor[HTML]{C8E9D9}45.70 & \cellcolor[HTML]{FFECB5}43.20 & \cellcolor[HTML]{D6EFE3}43.30 & \cellcolor[HTML]{FFFEFB}35.50 & \cellcolor[HTML]{DCF1E7}42.30 & \cellcolor[HTML]{FFFBF0}36.70 & \cellcolor[HTML]{BEE5D2}47.50 & \cellcolor[HTML]{FFEBB3}43.40 & \cellcolor[HTML]{BFE6D3}47.20 & \cellcolor[HTML]{FFF4D6}39.60 \\
\cellcolor[HTML]{FFFFFF}{\color[HTML]{0D0D0D} \textit{npi}}                    & \cellcolor[HTML]{AFDFC7}{\color[HTML]{0D0D0D} 50.00} & \cellcolor[HTML]{FFFFFF}{\color[HTML]{0D0D0D} 28.00} & \cellcolor[HTML]{ADDEC6}50.40 & \cellcolor[HTML]{FFE9AA}44.40 & \cellcolor[HTML]{C2E6D4}46.80 & \cellcolor[HTML]{FFFFFF}26.70 & \cellcolor[HTML]{BFE6D3}47.20 & \cellcolor[HTML]{FFFEFA}35.60 & \cellcolor[HTML]{99D6B8}53.70 & \cellcolor[HTML]{FFECB8}42.90 & \cellcolor[HTML]{9DD8BB}53.10 & \cellcolor[HTML]{FFFCF4}36.30 \\
\cellcolor[HTML]{FFFFFF}{\color[HTML]{0D0D0D} \textit{ory}}                    & \cellcolor[HTML]{C5E8D7}{\color[HTML]{0D0D0D} 46.20} & \cellcolor[HTML]{FFFFFF}{\color[HTML]{0D0D0D} 31.90} & \cellcolor[HTML]{B8E2CE}48.50 & \cellcolor[HTML]{FFF3D2}40.00 & \cellcolor[HTML]{C8E9D9}45.70 & \cellcolor[HTML]{FFFFFF}29.70 & \cellcolor[HTML]{D6EFE2}43.40 & \cellcolor[HTML]{FFFEFA}35.60 & \cellcolor[HTML]{C2E6D4}46.80 & \cellcolor[HTML]{FFFAEB}37.30 & \cellcolor[HTML]{C3E7D6}46.50 & \cellcolor[HTML]{FFFFFE}35.20 \\
\rowcolor[HTML]{FFFFFF} 
{\color[HTML]{0D0D0D} \textit{pan}}                                            & \cellcolor[HTML]{AFDFC7}{\color[HTML]{0D0D0D} 50.00} & {\color[HTML]{0D0D0D} 27.00}                         & \cellcolor[HTML]{A7DCC2}51.40 & 34.40                         & \cellcolor[HTML]{BAE3CF}48.10 & 28.50                         & \cellcolor[HTML]{B6E2CC}48.80 & 28.40                         & \cellcolor[HTML]{A6DBC1}51.50 & 34.40                         & \cellcolor[HTML]{AADDC4}50.90 & 29.80                         \\
\cellcolor[HTML]{FFFFFF}{\color[HTML]{0D0D0D} \textit{snd}}                    & \cellcolor[HTML]{C0E6D3}{\color[HTML]{0D0D0D} 47.10} & {\color[HTML]{0D0D0D} -----}                        & \cellcolor[HTML]{B5E2CC}48.90 & -----                         & \cellcolor[HTML]{C9E9D9}----- & -----                         & \cellcolor[HTML]{C7E9D8}45.90 & -----                         & \cellcolor[HTML]{DCF1E7}42.30 & -----                         & \cellcolor[HTML]{DEF2E8}42.00 & -----                         \\
\cellcolor[HTML]{FFFFFF}{\color[HTML]{0D0D0D} \textit{tam}}                    & \cellcolor[HTML]{ABDDC5}{\color[HTML]{0D0D0D} 50.60} & \cellcolor[HTML]{FFFFFF}{\color[HTML]{0D0D0D} 33.80} & \cellcolor[HTML]{A2D9BE}52.30 & \cellcolor[HTML]{FFF2CE}40.50 & \cellcolor[HTML]{ACDEC5}50.50 & \cellcolor[HTML]{FFEDB9}42.80 & \cellcolor[HTML]{B2E0C9}49.50 & \cellcolor[HTML]{FFF0C7}41.20 & \cellcolor[HTML]{9CD7BA}53.30 & \cellcolor[HTML]{FFDE83}48.80 & \cellcolor[HTML]{9ED8BB}53.00 & \cellcolor[HTML]{FFE8A7}44.80 \\
\cellcolor[HTML]{FFFFFF}{\color[HTML]{0D0D0D} \textit{tel}}                    & \cellcolor[HTML]{A3DABF}{\color[HTML]{0D0D0D} 52.00} & \cellcolor[HTML]{FFFFFF}{\color[HTML]{0D0D0D} 33.20} & \cellcolor[HTML]{9BD7B9}53.50 & \cellcolor[HTML]{FFE08C}47.80 & \cellcolor[HTML]{B0DFC8}49.90 & \cellcolor[HTML]{FFFCF1}36.60 & \cellcolor[HTML]{A7DCC2}51.30 & \cellcolor[HTML]{FFF3D1}40.10 & \cellcolor[HTML]{90D2B2}55.30 & \cellcolor[HTML]{FFE49A}46.20 & \cellcolor[HTML]{95D4B5}54.50 & \cellcolor[HTML]{FFEEBE}42.20 \\
\cellcolor[HTML]{FFFFFF}{\color[HTML]{0D0D0D} \textit{urd}}                    & \cellcolor[HTML]{BFE6D3}{\color[HTML]{0D0D0D} 47.20} & \cellcolor[HTML]{FFF1C8}{\color[HTML]{0D0D0D} 41.10} & \cellcolor[HTML]{BFE5D2}47.30 & \cellcolor[HTML]{FFDC7A}49.70 & \cellcolor[HTML]{C8E9D9}45.80 & \cellcolor[HTML]{FFF1C8}41.10 & \cellcolor[HTML]{C3E7D5}46.60 & \cellcolor[HTML]{FFE6A0}45.50 & \cellcolor[HTML]{B6E2CC}48.80 & \cellcolor[HTML]{FFDC7B}49.60 & \cellcolor[HTML]{B6E2CC}48.80 & \cellcolor[HTML]{FFE59B}46.10 \\ \midrule
\multicolumn{1}{l}{\textit{avg.}}                                                           & \multicolumn{1}{r}{48.10}                            & \multicolumn{1}{r}{33.20}                            & \multicolumn{1}{r}{49.46}     & \multicolumn{1}{r}{44.75}     & \multicolumn{1}{r}{47.38}     & \multicolumn{1}{r}{37.11}     & \multicolumn{1}{r}{46.85}     & \multicolumn{1}{r}{39.33}     & \multicolumn{1}{r}{50.42}     & \multicolumn{1}{r}{44.84}     & \multicolumn{1}{r}{49.97}     & \multicolumn{1}{r}{40.90}    \\ \bottomrule
\end{tabular}
\caption{chrF++ ($\uparrow$) scores for different direct and cascaded speech translation models evaluated on the \textsc{Fleurs-Test} (FL; \textcolor{customgreen}{green}) and \textsc{BhasaAnuvaad-Test} (BA; \textcolor{customyellow}{yellow}) datasets in $En \rightarrow XX$ direction. The colors represent a heatmap, where darker shades indicating better translation performance.}
\label{tab:consolidated-en-xx}
\end{table*}

\subsubsection{Data}

We construct the seed dataset by retaining only those audio-text pairs that meet the following quality thresholds: a cosine similarity-based mining score $\sigma \geq 0.6$, and a Levenshtein distance-based alignment score $\tau \geq 0.8$ , following~\citet{kaushalShrutilipi} for the alignment criterion, from the newly collected sources. These thresholds are chosen to ensure that only highly reliable and accurately aligned data contribute to model fine-tuning.

From the filtered data, we construct the evaluation set---\textsc{BhasaAnuvaad-test} (Table~
\ref{tab:bhasaanuvaad_test})---by randomly sampling approximately 20 minutes of speech per language for each translation direction (En~$\rightarrow$~XX and XX~$\rightarrow$~En). This setup provides a balanced test set across all covered languages, while preserving diversity in speaker accents, linguistic complexity, and domain coverage. The remaining filtered data, excluding the test portion, is used for training the model.

\subsection{Baselines}
\noindent \textbf{Unified Systems: }~In this paradigm, a single end-to-end speech translation model is used. The unified AST models considered in our study are as follows:

\noindent \colorbox{lightblue}{\seamless~(SD)} {\raisebox{-0.4em}{\includegraphics[height=1.4em]{figures/meta.png}}}~\cite{communication2023seamlessm4t}~is a foundational multilingual and multitask model supporting speech-to-speech (S2S), speech-to-text (S2T), text-to-speech (T2S), text-to-text (T2T) translation, and automatic speech recognition (ASR) for up to 100 languages. For our experiments, we use the \textsc{Seamless-M4T-v2-large model}.

\noindent \colorbox{lightblue}{\azure}~{\raisebox{-0.2em}{\includegraphics[height=1em]{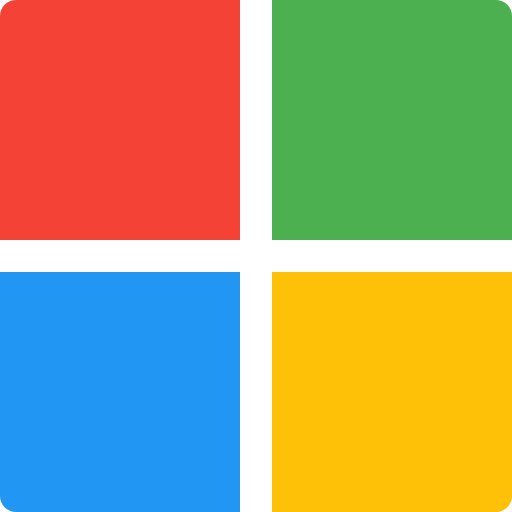}}}~is a closed-source speech translation system developed by Microsoft, providing cloud-based automatic speech recognition (ASR) and speech-to-text (S2T) translation services. While details about its architecture and training data remain proprietary, Azure's speech translation API supports multiple languages and is widely used in commercial applications. For our study, we evaluate its performance against other baselines and SD\textsubscript{BA}.

\begin{table*}[h]
\fontsize{8pt}{10pt}\selectfont
\centering
\begin{tabular}{@{}crrrrrrrrrrrr@{}}
\toprule
\cellcolor[HTML]{FFFFFF}{\color[HTML]{31333F} }                                & \multicolumn{6}{c}{\textbf{Direct}}                                                                                                                                                                                                                      & \multicolumn{6}{c}{\textbf{Cascaded}}                                                                                                                                                         \\ \cmidrule(lr){2-7} \cmidrule(lr){8-13}
\cellcolor[HTML]{FFFFFF}{\color[HTML]{31333F} }                                & \multicolumn{2}{c}{\textbf{SD}}                                                                           & \multicolumn{2}{c}{\textbf{SD\textsubscript{BA}}}                           & \multicolumn{2}{c}{\textbf{AZURE}}                                           & \multicolumn{2}{c}{\textbf{SC}}                               & \multicolumn{2}{c}{\textbf{IC + IT2}}                          & \multicolumn{2}{c}{\textbf{SC + IT2}}                         \\ \cmidrule(lr){2-3} \cmidrule(lr){4-5} \cmidrule(lr){6-7} \cmidrule(lr){8-9} \cmidrule(lr){10-11} \cmidrule(lr){12-13}
\multirow{-3}{*}{\cellcolor[HTML]{FFFFFF}{\color[HTML]{31333F} \textit{lang}}} & \textbf{FL}                                     & \textbf{BA}                                         & \textbf{FL}               & \textbf{BA}                   & \textbf{FL}                              & \textbf{BA}                   & \textbf{FL}               & \textbf{BA}                   & \textbf{FL}               & \textbf{BA}                   & \textbf{FL}               & \textbf{BA}                   \\ \midrule
\textit{asm}                        & \cellcolor[HTML]{FFFFFF}44.00 & \cellcolor[HTML]{FFEFC3}46.90 & \cellcolor[HTML]{D8F0E4}49.70 & \cellcolor[HTML]{FFE6A0}53.80 & \cellcolor[HTML]{FFFFFF}38.20 & \cellcolor[HTML]{FFF7E0}41.20 & \cellcolor[HTML]{F5FBF8}46.30 & \cellcolor[HTML]{FFECB6}49.40 & \cellcolor[HTML]{EDF8F3}47.20 & \cellcolor[HTML]{FFE9AB}51.50 & \cellcolor[HTML]{EEF8F3}47.10 & \cellcolor[HTML]{FFEAAF}50.70 \\
\textit{ben}                        & \cellcolor[HTML]{E2F4EB}48.50 & \cellcolor[HTML]{FFF2CB}45.20 & \cellcolor[HTML]{ABDDC5}55.00 & \cellcolor[HTML]{FFE18D}57.40 & \cellcolor[HTML]{F6FCF9}46.10 & \cellcolor[HTML]{FFF8E3}40.50 & \cellcolor[HTML]{CBEADB}51.20 & \cellcolor[HTML]{FFEFC1}47.20 & \cellcolor[HTML]{C0E6D4}52.50 & \cellcolor[HTML]{FFECB6}49.50 & \cellcolor[HTML]{BAE3CF}53.30 & \cellcolor[HTML]{FFEDB9}48.80 \\
\textit{guj}                        & \cellcolor[HTML]{BEE5D2}52.80 & \cellcolor[HTML]{FFE9AD}51.20 & \cellcolor[HTML]{80CCA7}60.20 & \cellcolor[HTML]{FFDD80}60.00 & \cellcolor[HTML]{FCFEFD}45.40 & \cellcolor[HTML]{FFFFFF}32.70 & \cellcolor[HTML]{A4DAC0}55.90 & \cellcolor[HTML]{FFE59E}54.20 & \cellcolor[HTML]{9AD6B9}57.10 & \cellcolor[HTML]{FFE59C}54.50 & \cellcolor[HTML]{96D5B6}57.50 & \cellcolor[HTML]{FFE59D}54.40 \\
\textit{hin}                        & \cellcolor[HTML]{D0ECDF}50.60 & \cellcolor[HTML]{FFEDBA}48.60 & \cellcolor[HTML]{89D0AD}59.10 & \cellcolor[HTML]{FFDE83}59.40 & \cellcolor[HTML]{CAEADA}51.40 & \cellcolor[HTML]{FFEEC0}47.50 & \cellcolor[HTML]{BBE4D0}53.20 & \cellcolor[HTML]{FFE9AD}51.20 & \cellcolor[HTML]{9FD8BC}56.50 & \cellcolor[HTML]{FFE395}55.80 & \cellcolor[HTML]{9FD8BC}56.50 & \cellcolor[HTML]{FFE6A0}53.70 \\
\textit{kan}                        & \cellcolor[HTML]{EAF7F0}47.60 & \cellcolor[HTML]{FFEEBF}47.60 & \cellcolor[HTML]{ADDEC6}54.80 & \cellcolor[HTML]{FFE290}56.90 & \cellcolor[HTML]{EEF8F3}47.10 & \cellcolor[HTML]{FFFBF0}38.10 & \cellcolor[HTML]{D5EEE2}50.10 & \cellcolor[HTML]{FFEAAF}50.80 & \cellcolor[HTML]{C5E8D7}52.00 & \cellcolor[HTML]{FFE8A6}52.50 & \cellcolor[HTML]{C5E8D7}52.00 & \cellcolor[HTML]{FFE8A8}52.20 \\
\textit{mal}                        & \cellcolor[HTML]{E3F4EC}48.40 & \cellcolor[HTML]{FFF4D4}43.50 & \cellcolor[HTML]{A4DAC0}55.90 & \cellcolor[HTML]{FFE69F}54.00 & \cellcolor[HTML]{F7FCFA}46.00 & \cellcolor[HTML]{FFFFFC}35.60 & \cellcolor[HTML]{C5E8D7}52.00 & \cellcolor[HTML]{FFF0C6}46.30 & \cellcolor[HTML]{B2E0CA}54.20 & \cellcolor[HTML]{FFECB6}49.50 & \cellcolor[HTML]{B3E1CA}54.10 & \cellcolor[HTML]{FFEDBA}48.70 \\
\textit{mar}                        & \cellcolor[HTML]{E3F4EC}48.40 & \cellcolor[HTML]{FFF0C6}46.20 & \cellcolor[HTML]{A9DCC3}55.30 & \cellcolor[HTML]{FFE393}56.20 & \cellcolor[HTML]{FFFFFF}43.80 & \cellcolor[HTML]{FFFCF2}37.70 & \cellcolor[HTML]{C3E7D5}52.20 & \cellcolor[HTML]{FFEBB2}50.10 & \cellcolor[HTML]{B1E0C9}54.30 & \cellcolor[HTML]{FFE9AD}51.10 & \cellcolor[HTML]{B8E3CE}53.50 & \cellcolor[HTML]{FFEBB2}50.20 \\
{\color[HTML]{0D0D0D} \textit{npi}} & \cellcolor[HTML]{D0ECDF}50.60 & \cellcolor[HTML]{FFFFFC}35.70 & \cellcolor[HTML]{9ED8BC}56.60 & \cellcolor[HTML]{FFE8A9}51.90 & \cellcolor[HTML]{D5EEE2}50.10 & \cellcolor[HTML]{FFFFFF}31.20 & \cellcolor[HTML]{B1E0C9}54.30 & \cellcolor[HTML]{FFFAED}38.70 & \cellcolor[HTML]{A9DCC3}55.30 & \cellcolor[HTML]{FFF0C5}46.50 & \cellcolor[HTML]{A0D9BD}56.40 & \cellcolor[HTML]{FFF7E1}40.90 \\
{\color[HTML]{0D0D0D} \textit{ory}} & \cellcolor[HTML]{E1F3EA}48.60 & \cellcolor[HTML]{FFEBB2}50.10 & \cellcolor[HTML]{ACDEC6}54.90 & \cellcolor[HTML]{FFE08C}57.70 & \cellcolor[HTML]{E5F5ED}48.20 & \cellcolor[HTML]{FFF3D2}44.00 & \cellcolor[HTML]{BEE5D2}52.80 & \cellcolor[HTML]{FFE6A1}53.50 & \cellcolor[HTML]{ABDDC4}55.10 & \cellcolor[HTML]{FFE8A8}52.10 & \cellcolor[HTML]{B2E0CA}54.20 & \cellcolor[HTML]{FFE7A4}52.90 \\
{\color[HTML]{0D0D0D} \textit{pan}} & \cellcolor[HTML]{D3EDE0}50.30 & \cellcolor[HTML]{FFEBB3}50.00 & \cellcolor[HTML]{A0D9BD}56.40 & \cellcolor[HTML]{FFE18C}57.60 & \cellcolor[HTML]{FFFFFF}44.00 & \cellcolor[HTML]{FFF2CE}44.70 & \cellcolor[HTML]{BCE4D1}53.00 & \cellcolor[HTML]{FFE499}55.00 & \cellcolor[HTML]{ABDDC4}55.10 & \cellcolor[HTML]{FFE6A1}53.60 & \cellcolor[HTML]{AEDFC7}54.70 & \cellcolor[HTML]{FFE497}55.40 \\
{\color[HTML]{0D0D0D} \textit{snd}} & \cellcolor[HTML]{FFFFFF}31.10 & {\color[HTML]{0D0D0D} -----}  & \cellcolor[HTML]{FFFFFF}43.40 & {\color[HTML]{0D0D0D} -----}  & {\color[HTML]{0D0D0D} -----}  & {\color[HTML]{0D0D0D} -----}  & \cellcolor[HTML]{FFFFFF}36.20 & {\color[HTML]{0D0D0D} -----}  & \cellcolor[HTML]{FFFFFF}39.30 & {\color[HTML]{0D0D0D} -----}  & \cellcolor[HTML]{FFFFFF}38.50 & {\color[HTML]{0D0D0D} -----}  \\
{\color[HTML]{0D0D0D} \textit{tam}} & \cellcolor[HTML]{FFFFFF}44.80 & \cellcolor[HTML]{FFF6DE}41.60 & \cellcolor[HTML]{CDEBDC}51.00 & \cellcolor[HTML]{FFE8A8}52.20 & \cellcolor[HTML]{FCFEFD}45.40 & \cellcolor[HTML]{FFFCF4}37.30 & \cellcolor[HTML]{E5F5ED}48.10 & \cellcolor[HTML]{FFF3D0}44.30 & \cellcolor[HTML]{D0ECDE}50.70 & \cellcolor[HTML]{FFEFC1}47.30 & \cellcolor[HTML]{D3EDE0}50.30 & \cellcolor[HTML]{FFEFC3}46.80 \\
{\color[HTML]{0D0D0D} \textit{tel}} & \cellcolor[HTML]{EAF7F1}47.50 & \cellcolor[HTML]{FFF3CF}44.50 & \cellcolor[HTML]{B1E0C9}54.30 & \cellcolor[HTML]{FFE499}55.00 & \cellcolor[HTML]{E4F4EC}48.30 & \cellcolor[HTML]{FFF9E9}39.40 & \cellcolor[HTML]{C9E9D9}51.50 & \cellcolor[HTML]{FFEEC0}47.50 & \cellcolor[HTML]{A6DBC1}55.60 & \cellcolor[HTML]{FFE9AD}51.20 & \cellcolor[HTML]{B1E0C9}54.40 & \cellcolor[HTML]{FFEBB2}50.10 \\
{\color[HTML]{0D0D0D} \textit{urd}} & \cellcolor[HTML]{EAF7F0}47.60 & \cellcolor[HTML]{FFEDB9}48.90 & \cellcolor[HTML]{B1E0C9}54.40 & \cellcolor[HTML]{FFDC7A}61.10 & \cellcolor[HTML]{EAF7F0}47.60 & \cellcolor[HTML]{FFF5D8}42.80 & \cellcolor[HTML]{D3EDE0}50.30 & \cellcolor[HTML]{FFE9AD}51.20 & \cellcolor[HTML]{CBEADB}51.30 & \cellcolor[HTML]{FFE7A5}52.70 & \cellcolor[HTML]{BBE4D0}53.10 & \cellcolor[HTML]{FFE7A4}53.00 \\ \midrule
\multicolumn{1}{l}{\textit{avg.}}                & 47.20                         & 46.15                         & 54.36                         & 56.40                         & 46.28                         & 39.44                         & 50.51                         & 49.18                         & 52.59                         & 51.37                         & 52.54                         & 50.60    \\ \bottomrule
\end{tabular}
\caption{chrF++ ($\uparrow$) scores for different direct and cascaded speech translation models evaluated on the \textsc{Fleurs-Test} (FL; \textcolor{customgreen}{green}) and \textsc{BhasaAnuvaad-Test} (BA; \textcolor{customyellow}{yellow}) datasets in $XX \rightarrow En$ direction. The colors represent a heatmap, where darker shades indicating better translation performance.}
\label{tab:consolidated-xx-en}
\end{table*}

\setlength\fboxsep{1pt}
\noindent \textbf{Cascaded Systems: }~In this paradigm, separately trained Automatic Speech Recognition (ASR) and Machine Translation (MT) models are combined in a pipeline for performing Automatic Speech Translation (AST). The different ASR and NMT models used in our experiments are listed below.

\noindent \colorbox{lightorange}{\seamless(SC)}{\raisebox{-0.4em}{\includegraphics[height=1.4em]{figures/meta.png}}}\cite{communication2023seamlessm4t}~where we integrate the ASR and NMT capabilities of the \seamless~model in a sequential pipeline.


\noindent \colorbox{lightorange}{\whisper~(W)} {\raisebox{-0.2em}{\includegraphics[height=1.0em]{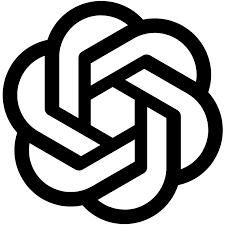}}}~\cite{whisper}~is a multilingual model trained for automatic speech recognition (ASR) and speech translation (AST). For our experiments, we use the \textsc{Whisper-large-v3} model which has been trained on 1 million hours of weakly labeled and 4 million hours of pseudo-labeled audio.


\noindent \colorbox{lightorange}{\conformer~(IC)\footnote{\url{https://ai4bharat.iitm.ac.in/areas/model/ASR/IndicConformer}}} {\raisebox{-0.4em}{\includegraphics[height=1.4em]{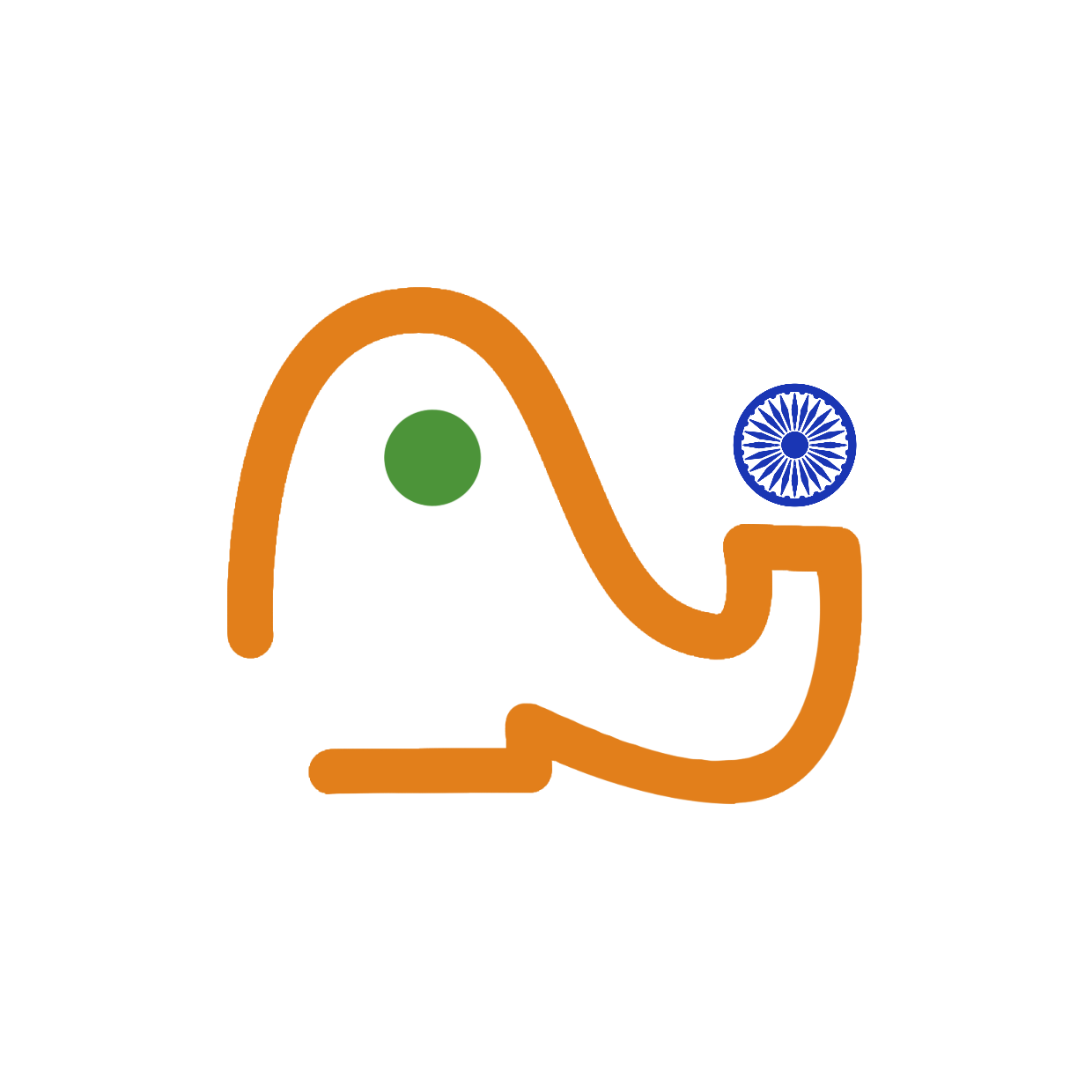}}}\cite{bhogale2025towards} which is a suite of ASR models based on the Conformer architecture, models supporting all 22 scheduled Indian languages.

\noindent \colorbox{lightorange}{\indictrans~(IT2)} {\raisebox{-0.4em}{\includegraphics[height=1.4em]{figures/ai4b.png}}}\cite{gala2023indictrans2}~which is an open-source transformer-based multilingual NMT model supporting translations across all 22 scheduled Indic languages. For our experiments, we use the \textsc{IndicTrans2-1B} model.


\noindent In our experiments, we combine the \textsc{IndicTrans2-1B} model and the \textsc{Seamless-M4T V2 Large} translation model with various ASR models to perform automatic speech translation (AST). Specifically, the translation capabilities of \textsc{IndicTrans2-1B} and \textsc{Seamless-M4T V2 Large} (T2TT) are integrated with ASR outputs from systems such as \textsc{Whisper}, and \textsc{Indic-Conformer}.

\subsection{Metrics}
\label{subsec:metrics}

In this work, we adopt chrF++~\citep{popovic2017chrf} as our primary metric, which offers strong alignment with human judgments~\cite{sai-b-etal-2023-indicmt} particularly Indian languages. We compute chrF++ scores using sacreBLEU~\cite{post-2018-call}; for Indic-En\footnote{\texttt{En-Indic sacreBLEU ChrF++ \\\tiny{signature:nrefs:1|case:mixed|eff:yes|nc:6|nw:2|space:no|version:2.3.1}}} evaluation, we use the standard \texttt{mteval-v13a} tokenizer, while for En-Indic, we apply Indic-specific tokenizers from IndicNLP~\cite{kunchukuttan2020indicnlp} and Urduhack\footnote{\url{https://github.com/urduhack/urduhack}} before scoring, ensuring linguistically informed segmentation. We also report additional metrics such as BLEU~\cite{papineni2002bleu} and XCOMET~\cite{guerreiro-etal-2024-xcomet} scores in Appendix Tables \ref{tab:comet_xx_en} \& \ref{tab:comet_en_xx} and \ref{tab:bleu_en_xx} \& \ref{tab:bleu_xx_en} respectively for both the directions.

\section{Results}

We evaluate the translation performance of various models in both En~$\rightarrow$XX and XX$\rightarrow$~En directions using chrF++ scores on the \textsc{Fleurs-Test} and \textsc{BhasaAnuvaad-Test} (Table \ref{tab:bhasaanuvaad_test}) datasets. Our analysis focuses on the comparative effectiveness of direct and cascaded models, highlighting the impact of fine-tuning and architectural choices.

\subsection{Direct vs. Cascaded Systems}

A comparison between direct and cascaded approaches on the \textsc{Fleurs-Test} set reveals that their performance remains closely matched across most languages. In En~$\rightarrow$XX translation, the average chrF++ score difference between the two paradigms is marginal, suggesting that direct speech translation methods are increasingly closing the performance gap with cascaded pipelines. To analyze this further, we compare the strongest direct model, SD\textsubscript{BA}, with the best-performing cascaded setups: W + IT2 for En$\rightarrow$XX and IC + IT2 for XX$\rightarrow$~En.

For En~$\rightarrow$ XX (Table~\ref{tab:consolidated-en-xx}), SD\textsubscript{BA} and W + IT2 achieve comparable results across most languages, underscoring the efficacy of both direct speech translation and cascaded modeling. However, an exception arises in Tamil in \textsc{BhasaAnuvaad-Test}, where W + IT2 significantly outperforms SD\textsubscript{BA}. This deviation warrants further investigation into language-specific factors that may influence direct model performance.

In the XX~$\rightarrow$~En direction, SD\textsubscript{BA} surpasses IC + IT2 across multiple languages, demonstrating strong generalization capabilities. Notably, on the \textsc{BhasaAnuvaad-Test} set, SD\textsubscript{BA} achieves significantly higher \textsc{ChrF++} scores for Nepali, Bengali, and Urdu, reinforcing its ability to handle diverse linguistic structures effectively. These results suggest that fine-tuned direct models can be competitive or even superior to cascaded systems, where cascading errors from ASR and MT modules compound and degrade translation quality.

\subsection{Performance Gains from Fine-Tuning}

Fine-tuning on the BhasaAnuvaad dataset (BA) leads to consistent improvements across both translation directions (En~$\rightarrow$XX and XX$\rightarrow$~En). SD\textsubscript{BA} achieves substantial gains over SD, particularly on the BA test set, where it outperforms the baseline by at least 10 points on average.

On the \textsc{FLEURS-Test} set, SD\textsubscript{BA} consistently outperforms SD across all languages, achieving an average improvement of over 7 points. These results highlight the effectiveness of domain-specific fine-tuning in enhancing direct speech translation performance, particularly in real-world, low-resource conditions.

\section{Conclusion}

In this work, we present \textsc{BhasaAnuvaad}, the largest publicly available Indian language speech translation dataset, comprising 44,000+ hours of speech and 17 million aligned text segments across 14 Indian languages and English. Our dataset is built using a threefold approach: \textbf{(i)} aggregating high-quality sources, \textbf{(ii)} large-scale web crawling, and \textbf{(iii)} synthetic augmentation to capture real-world disfluencies. Using \textsc{BhasaAnuvaad}, we train \textsc{IndicSeamless}, a state-of-the-art speech translation model for Indian languages. Our evaluations show that fine-tuned direct speech translation models (SD\textsubscript{BA}) match or surpass strong cascaded systems, particularly in low-resource settings where cascading errors degrade performance. By releasing all data, models, and code under open-source licenses, we aim to advance research in Indian language speech translation.



\section{Limitations}
Despite the scale and diversity of \textsc{BhasaAnuvaad}, several limitations remain, addressing which will be essential to making speech translation systems for Indian languages more robust, reliable, and truly representative of real-world usage.

First, while our dataset integrates a range of sources, spontaneous speech remains underrepresented. Real-world conversational speech often includes disfluencies, hesitations, code-switching, and speaker variability that are not fully captured in our current dataset. 

Second, domain coverage remains a challenge. While \textsc{BhasaAnuvaad} aggregates speech from multiple sources including educational content, government archives, and podcasts certain critical domains such as informal dialogues, social media discourse, medical interactions, and legal proceedings remain underrepresented. 

Third, benchmarking Indian language AST remains an open challenge. Most prior evaluations rely on \textsc{FLEURS}, which consists of acted-out, read speech rather than spontaneous speech. While useful for measuring system performance under controlled conditions, \textsc{FLEURS} fails to reflect the complexities of natural, conversational speech, where hesitations, overlaps, and real-world noise significantly impact translation quality. Our experiments indicate that models achieving strong results on \textsc{FLEURS} do not necessarily maintain similar performance on more challenging, real-world speech scenarios. 

\section{Ethics}

The code and datasets created in this work will be made available under permissible licenses. Generative AI systems were only used for assistance purely with the language of the paper, eg: paraphrasing, spell-check, polishing the author's original content and for writing boiler plate code.

\section{Acknowledgements}

We gratefully acknowledge the generous support and funding provided by Digital India Bhashini, the Centre for Development of Advanced Computing (C-DAC) Pune, Yotta, EkStep Foundation, and Nilekani Philanthropies. We also thank Pranjal Agadh Chitale for his insightful discussions and valuable feedback throughout the development of this work.

\bibliography{custom}
\appendix
\section*{Appendix}

\setcounter{table}{0}
\renewcommand{\thetable}{A\arabic{table}}
\setcounter{figure}{0}
\renewcommand{\thefigure}{A\arabic{figure}}

\section{Data Statistics}
\label{sec:data_stats}

This section outlines the datasets used for training and evaluation, covering diverse Indic languages and both En-Indic and Indic-En directions. We begin with the test set composition, followed by training data statistics.

\begin{table*}[]
\fontsize{9pt}{10pt}\selectfont
\centering
\scriptsize
\setlength{\tabcolsep}{3pt} 
\begin{tabular}{@{}lrrrrrrrrrrrrrrrr@{}}
\toprule
\multirow{3}{*}{\textit{lang}} & \multicolumn{4}{c}{\textbf{Existing}}               & \multicolumn{10}{c}{\textbf{Mined}}                                                                                                                                                                                                                 & \multicolumn{2}{c}{\textbf{Synthetic}}              \\ \cmidrule(lr){2-5}\cmidrule(lr){6-15}\cmidrule(lr){16-17}
                                   & \multicolumn{2}{c}{\textbf{Indic-TedST}} & \multicolumn{2}{c}{\textbf{Fleurs}} & \multicolumn{2}{c}{\textbf{VaaniPedia}}    & \multicolumn{2}{c}{\textbf{WordProject}} & \multicolumn{2}{c}{\textbf{UGCE Resources}} & \multicolumn{2}{c}{\textbf{Mann Ki Baat}}  & \multicolumn{2}{c}{\textbf{NPTEL}}         & \multicolumn{2}{c}{\textbf{IndicVoices}}  \\
                                   \cmidrule(lr){2-3} \cmidrule(lr){4-5} \cmidrule(lr){6-7} \cmidrule(lr){8-9} \cmidrule(lr){10-11} \cmidrule(lr){12-13} \cmidrule(lr){14-15} \cmidrule(lr){16-17}
                                   & \begin{tabular}[c]{@{}c@{}}\textbf{\#}\\\textbf{hours}\end{tabular} & \begin{tabular}[c]{@{}c@{}}\textbf{\#}\\\textbf{uttr.}\end{tabular} & \begin{tabular}[c]{@{}c@{}}\textbf{\#}\\\textbf{hours}\end{tabular} & \begin{tabular}[c]{@{}c@{}}\textbf{\#}\\\textbf{uttr.}\end{tabular} & \begin{tabular}[c]{@{}c@{}}\textbf{\#}\\\textbf{hours}\end{tabular} & \begin{tabular}[c]{@{}c@{}}\textbf{\#}\\\textbf{uttr.}\end{tabular}      & \begin{tabular}[c]{@{}c@{}}\textbf{\#}\\\textbf{hours}\end{tabular} & \begin{tabular}[c]{@{}c@{}}\textbf{\#}\\\textbf{uttr.}\end{tabular} & \begin{tabular}[c]{@{}c@{}}\textbf{\#}\\\textbf{hours}\end{tabular} & \begin{tabular}[c]{@{}c@{}}\textbf{\#}\\\textbf{uttr.}\end{tabular} & \begin{tabular}[c]{@{}c@{}}\textbf{\#}\\\textbf{hours}\end{tabular} & \begin{tabular}[c]{@{}c@{}}\textbf{\#}\\\textbf{uttr.}\end{tabular} & \begin{tabular}[c]{@{}c@{}}\textbf{\#}\\\textbf{hours}\end{tabular} & \begin{tabular}[c]{@{}c@{}}\textbf{\#}\\\textbf{uttr.}\end{tabular} & \begin{tabular}[c]{@{}c@{}}\textbf{\#}\\\textbf{hours}\end{tabular} & \begin{tabular}[c]{@{}c@{}}\textbf{\#}\\\textbf{uttr.}\end{tabular} \\ \midrule
\textbf{asm}                       & -                 & -                      & 12.0              & 1.5                       & 3.4               & 1.6                    & -                       & -                           & -                  & -                      & 28.3              & 15.7                   & 10.6              & 5.3                    & 348.9             & 23.5                   \\
\textbf{ben}                       & 13.1              & 6.5                    & 12.0              & 1.5                       & 121.0             & 52.5                   & 50.7                    & 23.9                        & 431.7              & 329.6                  & 38.3              & 21.4                   & 1516.6            & 618.8                  & 300.5             & 30.3                   \\
\textbf{guj}                       & 17.2              & 1.5                    & 12.0              & 1.5                       & 121.0             & 52.5                   & 50.7                    & 23.9                        & 431.7              & 329.4                  & 39.1              & 22.0                   & 1977.9            & 784.9                  & 16.7              & 2.8                    \\
\textbf{hin}                       & 100.4             & 52.5                   & 12.0              & 1.5                       & 121.0             & 52.5                   & 50.7                    & 23.9                        & 431.7              & 329.2                  & 37.7              & 21.3                   & 3021.6            & 1176.8                 & 206.6             & 20.4                   \\
\textbf{kan}                       & 3.7               & 1.9                    & 12.0              & 1.5                       & 23.2              & 9.9                    & 50.7                    & 23.9                        & 431.7              & 328.6                  & 36.9              & 20.8                   & 1601.4            & 638.0                  & 175.6             & 11.7                   \\
\textbf{mal}                       & 9.2               & 3.2                    & 12.0              & 1.5                       & 0.5               & 52.5                   & 50.7                    & 23.9                        & 431.7              & 329.6                  & 38.3              & 21.3                   & 1729.1            & 699.3                  & 278.1             & 22.9                   \\
\textbf{mar}                       & 29.0              & 12.7                   & 12.0              & 1.5                       & 56.5              & 24.4                   & 50.7                    & 23.9                        & 431.7              & 322.4                  & 39.0              & 22.0                   & 1601.7            & 626.3                  & 138.0             & 16.2                   \\
\textbf{npi}                       & -                 & -                      & 12.0              & 1.5                       & 121.0             & 52.5                   & -                       & -                           & -                  & -                      & -                 & -                      & -                 & -                      & 343.9             & 31.6                   \\
\textbf{ory}                       & -                 & -                      & 12.0              & 1.5                       & 79.0              & 34.0                   & 49.2                    & 23.1                        & -                  & -                      & 36.8              & 20.5                   & 0.8               & 0.3                    & 161.1             & 19.1                   \\
\textbf{pan}                       & 1.2               & 0.4                    & 12.0              & 1.5                       & 0.5               & 0.3                    & 50.7                    & 23.9                        & -                  & -                      & 37.0              & 20.5                   & 9.6               & 5.8                    & 194.1             & 18.9                   \\
\textbf{snd}                       & -                 & -                      & 12.0              & 1.5                       & 121.0             & 52.5                   & -                       & -                           & -                  & -                      & -                 & -                      & -                 & -                      & -             & -                   \\
\textbf{tam}                       & 20.1              & 10.1                   & 12.0              & 1.5                       & 121.0             & 52.5                   & 30.5                    & 14.1                        & 431.7              & 327.9                  & 37.4              & 20.9                   & 2631.9            & 1048.1                 & 378.6             & 33.8                   \\
\textbf{tel}                       & 9.1               & 2.2                    & 12.0              & 1.5                       & 121.0             & 52.5                   & 50.7                    & 23.9                        & 431.7              & 322.6                  & 37.8              & 21.4                   & 1839.9            & 741.2                  & 314.3             & 21.6                   \\
\textbf{urd}                       & -                 & -                      & 12.0              & 1.5                       & 2.3               & 1.1                    & 50.7                    & 23.9                        & -                  & -                      & 29.0              & 16.0                   & -                 & -                      & 186.6             & 19.1                   \\ \midrule
\textbf{Total}                     & 202.9             & 91.0                   & 168.0             & 21                      & 891.1             & 439.0                  & 536.2                   & 252.3                       & 3453.8             & 2619.3                 & 435.7             & 244.0                  & 15940.9           & 6344.7                 & 2694.1            & 269.0                  \\ \bottomrule
\end{tabular}
\caption{Overview of En-Indic spoken translation dataset across various Indic languages. The table details the hours of audio and the number of utterances collected from multiple sources. Number of utterances are in thousands (K)}
\label{tab:en-in}

\bigskip

\begin{tabular}{@{}lrrrrrrrrrrrrrrrr@{}}
\toprule
\multirow{3}{*}{\textit{lang}} & \multicolumn{6}{c}{\textbf{Existing}}               & \multicolumn{8}{c}{\textbf{Mined}}                                                                                                                                                                                                                 & \multicolumn{2}{c}{\textbf{Synthetic}}              \\ \cmidrule(lr){2-7}\cmidrule(lr){8-15}\cmidrule(lr){16-17}
                                   & \multicolumn{2}{c}{\begin{tabular}[c]{@{}c@{}}\textbf{CVSS}\\\textbf{\& Khan} \\ \textbf{Academy}\end{tabular}} & \multicolumn{2}{c}{\textbf{SeamlessAlign}} & \multicolumn{2}{c}{\textbf{Fleurs}} & \multicolumn{2}{c}{\textbf{WordProject}}   & \multicolumn{2}{c}{\textbf{Spokentutorial}} & \multicolumn{2}{c}{\textbf{Mann Ki Baat}}  & \multicolumn{2}{c}{\textbf{NPTEL}}         & \multicolumn{2}{c}{\textbf{IndicVoices}}  \\
                                   \cmidrule(lr){2-3} \cmidrule(lr){4-5} \cmidrule(lr){6-7} \cmidrule(lr){8-9} \cmidrule(lr){10-11} \cmidrule(lr){12-13} \cmidrule(lr){14-15} \cmidrule(lr){16-17}
                                   & \begin{tabular}[c]{@{}c@{}}\textbf{\#}\\\textbf{hours}\end{tabular} & \begin{tabular}[c]{@{}c@{}}\textbf{\#}\\\textbf{uttr.}\end{tabular} & \begin{tabular}[c]{@{}c@{}}\textbf{\#}\\\textbf{hours}\end{tabular} & \begin{tabular}[c]{@{}c@{}}\textbf{\#}\\\textbf{uttr.}\end{tabular} & \begin{tabular}[c]{@{}c@{}}\textbf{\#}\\\textbf{hours}\end{tabular} & \begin{tabular}[c]{@{}c@{}}\textbf{\#}\\\textbf{uttr.}\end{tabular}      & \begin{tabular}[c]{@{}c@{}}\textbf{\#}\\\textbf{hours}\end{tabular} & \begin{tabular}[c]{@{}c@{}}\textbf{\#}\\\textbf{uttr.}\end{tabular} & \begin{tabular}[c]{@{}c@{}}\textbf{\#}\\\textbf{hours}\end{tabular} & \begin{tabular}[c]{@{}c@{}}\textbf{\#}\\\textbf{uttr.}\end{tabular} & \begin{tabular}[c]{@{}c@{}}\textbf{\#}\\\textbf{hours}\end{tabular} & \begin{tabular}[c]{@{}c@{}}\textbf{\#}\\\textbf{uttr.}\end{tabular} & \begin{tabular}[c]{@{}c@{}}\textbf{\#}\\\textbf{hours}\end{tabular} & \begin{tabular}[c]{@{}c@{}}\textbf{\#}\\\textbf{uttr.}\end{tabular} & \begin{tabular}[c]{@{}c@{}}\textbf{\#}\\\textbf{hours}\end{tabular} & \begin{tabular}[c]{@{}c@{}}\textbf{\#}\\\textbf{uttr.}\end{tabular} \\ \midrule
\textbf{asm}                       & -                     & -                       & -                 & -                      & 12.0              & 1.5                    & -                 & -                      & 36.9               & 23.2                   & 23.0              & 13.9                   & 9.3               & 5.7                    & 443.6             & 26.0                   \\
\textbf{ben}                       & -                     & -                       & -                 & -                      & 12.0              & 1.5                    & 117.6             & 28.9                   & 49.9               & 31.1                   & 38.5              & 20.7                   & 388.4             & 149.1                  & 450.6             & 33.2                   \\
\textbf{guj}                       & 12.0                  & 10.0                    & -                 & -                      & 12.0              & 1.5                    & 82.1              & 29.0                   & 63.6               & 37.6                   & 37.4              & 21.3                   & 795.0             & 300.9                  & 30.2              & 3.0                    \\
\textbf{hin}                       & -                     & -                       & 2355.4            & 1021.5                 & 12.0              & 1.5                    & 88.5              & 29.5                   & 63.6               & 37.9                   & 40.9              & 21.3                   & 1433.9            & 477.5                  & 282.9             & 21.6                   \\
\textbf{kan}                       & \multicolumn{1}{l}{}  & -                       & 174.2             & 68.6                   & 12.0              & 1.5                    & 108.7             & 29.5                   & 25.0               & 15.4                   & 38.5              & 21.1                   & 569.9             & 219.3                  & 199.5             & 12.2                   \\
\textbf{mal}                       & -                     & -                       & -                 & -                      & 12.0              & 1.5                    & 134.4             & 55.5                   & 24.4               & 15.4                   & 35.6              & 24.1                   & 721.0             & 320.2                  & 378.4             & 24.9                   \\
\textbf{mar}                       & \multicolumn{1}{l}{}  & -                       & -                 & -                      & 12.0              & 1.5                    & 98.8              & 29.2                   & 68.9               & 41.3                   & 42.1              & 22.7                   & 738.7             & 289.7                  & 201.7             & 17.5                   \\
\textbf{npi}                       & -                     & -                       & -                 & -                      & 12.0              & 1.5                    & -                 & -                      & 58.8               & 35.4                   & -                 & -                      & -                 & -                      & 479.1             & 34.0                   \\
\textbf{ory}                       & -                     & -                       & -                 & -                      & 12.0              & 1.5                    & 97.1              & 28.5                   & 6.5                & 3.5                    & 34.0              & 20.0                   & -                 & -                      & 219.6             & 20.0                   \\
\textbf{pan}                       & -                     & -                       & -                 & -                      & 12.0              & 1.5                    & 113.2             & 1.1                    & 19.8               & 11.4                   & 78.2              & 8.2                    & -                 & -                      & 231.4             & 19.9                   \\
\textbf{snd}                       & -                     & -                       & -                 & -                      & 12.0              & 1.5                    & -                 & -                      & -                  & -                      & -                 & -                      & -                 & -                      & -                 & -                      \\ 
\textbf{tam}                       & 96.0                  & 80.0                    & 1222.4            & 478.8                  & 12.0              & 1.5                    & 51.5              & 18.9                   & 69.1               & 39.9                   & 41.2              & 19.9                   & 1231.8            & 538.8                  & 454.5             & 35.5                   \\
\textbf{tel}                       & -                     & -                       & 873.3             & 329.5                  & 12.0              & 1.5                    & 90.4              & 29.2                   & 12.5               & 6.5                    & 35.3              & 26.3                   & 738.4             & 302.1                  & 377.5             & 22.8                   \\
\textbf{urd}                       & -                     & -                       & 2837.1            & 1107.7                 & 12.0              & 1.5                    & 94.3              & 29.4                   & -                  & -                      & 11.9              & 7.0                    & -                 & -                      & 239.2             & 17.4                   \\ \midrule
\textbf{Total}                     & 108.0                 & 90.0                    & 7462.3            & 3006.0                 & 156.0             & 21                   & 1076.7            & 308.6                  & 499.0              & 298.7                  & 456.4             & 226.5                  & 6626.5            & 2603.3                 & 3988.2            & 288.0                  \\ \bottomrule
\end{tabular}
\caption{Overview of Indic-En spoken translation dataset across various Indic languages. The table details the hours of audio and the number of utterances collected from multiple sources. Number of utterances are in thousands (K)}
\label{tab:in-en}
\end{table*}

\subsection{\textsc{BhasaAnuvaad-test}}
The \textsc{BhasaAnuvaad-test} set (Table~\ref{tab:bhasaanuvaad_test}) contains 26.22 hours of En-Indic and 28.18 hours of Indic-En speech translation data, covering 13 Indian languages. Most languages have balanced data in both directions, with minor variations. Notably, mni and npi have smaller amounts of test data, with mni lacking Indic-En coverage. This test set provides a reliable benchmark for evaluating spoken translation performance across diverse Indic languages.
\begin{table}[!h]
\fontsize{10pt}{12pt}\selectfont
\centering
\setlength{\tabcolsep}{3pt} 
\renewcommand{\arraystretch}{1} 
\begin{tabular}{@{}crr@{}}
\toprule
\multicolumn{1}{l}{\textit{lang}}                             & \textbf{En-XX}                                       & \textbf{XX-En} \\ \midrule
\textit{asm}                                                  & 1.51                                                 & 2.01           \\
\textit{ben}                                                  & 2.51                                                 & 2.50            \\
\textit{guj}                                                  & 2.51                                                 & 2.51           \\
\textit{hin}                                                  & 2.50                                                 & 2.50            \\
\textit{kan}                                                  & 2.51                                                 & 2.50            \\
\textit{mal}                                                  & 2.13                                                 & 2.50            \\
\textit{mar}                                                  & 2.51                                                 & 2.51           \\
\textit{npi}                                                  & 0.50                                                 & 0.64           \\
\textit{ory}                                                  & 1.50                                                 & 2.01           \\
\textit{pan}                                                  & 1.02                                                 & 1.98           \\
\textit{tam}                                                  & 2.51                                                 & 2.51           \\
\textit{tel}                                                  & 2.51                                                 & 2.51           \\
\textit{urd}                                                  & 1.50                                                 & 1.50
\\ \midrule
\cellcolor[HTML]{FFFFFF}{\color[HTML]{0D0D0D} \textit{tot.}} & \cellcolor[HTML]{FFFFFF}{\color[HTML]{0D0D0D} 26.22} & 28.18 \\ \bottomrule
\end{tabular}
\caption{Statistics of data obtained across each language in both Indic-En and En-Indic directions for  \textsc{BhasaAnuvaad-test} set. For each language and in both directions, we show the number of hours.}
\label{tab:bhasaanuvaad_test}
\end{table}


\subsection{Training Data}

\noindent \textbf{En-Indic:} The dataset combines 202.9 hours (91K utterances) of existing En-Indic speech data, 3453.8 hours (2619.3K utterances) of mined data, and 2694.1 hours (269K utterances) of synthetic data. Mined sources like NPTEL contribute the largest share with 15940.9 hours and 6344.7K utterances, while synthetic data from IndicVoices helps balance low-resource languages. This diverse collection ensures comprehensive coverage for spoken translation across multiple Indic languages.

\noindent \textbf{Indic-En}
The dataset includes 108 hours (90K utterances) of existing Indic-En speech data, 1076.7 hours (308.6K utterances) of mined data, and 3988.2 hours (288K utterances) of synthetic data. Among the mined sources, NPTEL contributes the largest portion with 6626.5 hours and 2603.3K utterances. The synthetic IndicVoices dataset further strengthens the collection with 3988.2 hours and 288K utterances, ensuring better representation across low-resource Indic languages. This combination provides a balanced and diverse dataset for advancing Indic-English spoken translation.

\section{Prompt Design for Punctuation Restoration}

\begin{figure}[t]
\centering
   \includegraphics[width=0.5\textwidth]{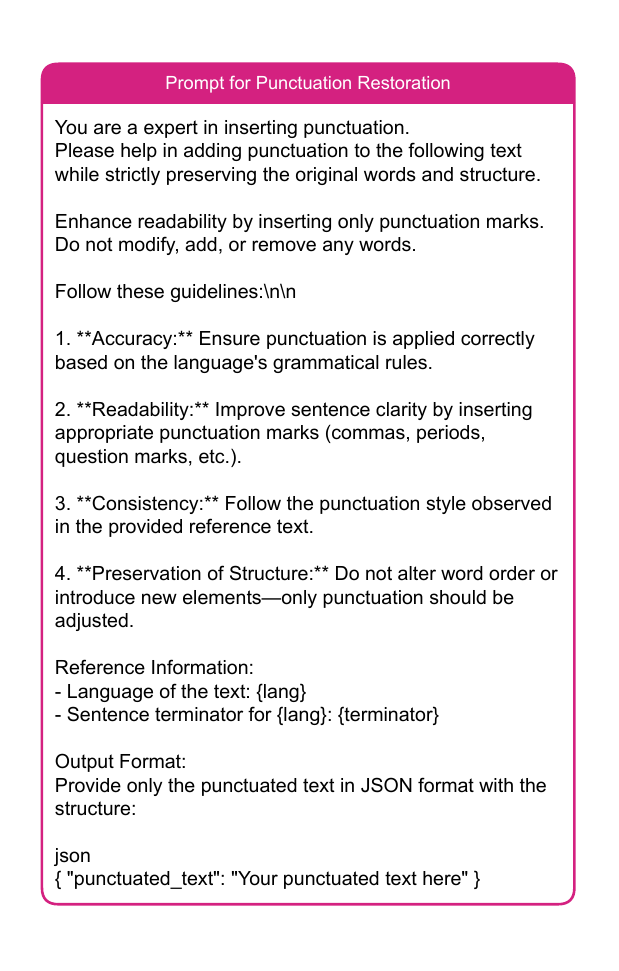}
    \caption{Prompt For Punctuation Restoration}
    \label{fig:Prompt_Punctuation}
\end{figure}

To address the lack of punctuation in automatic speech recognition (ASR) outputs and enhance downstream readability, we design a structured prompt for \textbf{punctuation restoration} using \textsc{Llama-3.1-405B-Instruct}. As shown in Figure~\ref{fig:Prompt_Punctuation}, the prompt guides the model to insert only punctuation marks into a given raw text while \textit{strictly preserving the original word order and structure}.

\begin{figure*}[!t]
    \centering
    \begin{minipage}[b]{0.46\textwidth}
        \centering
        \includegraphics[width=0.9\textwidth]{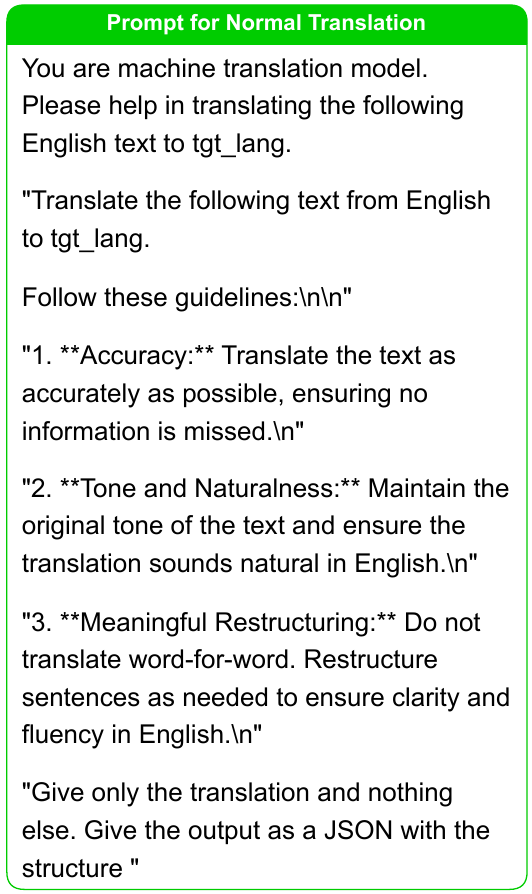}
        \caption{Prompt For Normal Translation}
        \label{fig:normal_translation}
    \end{minipage}
    \hfill
    \begin{minipage}[b]{0.48\textwidth}
        \centering
        \includegraphics[width=0.9\textwidth]{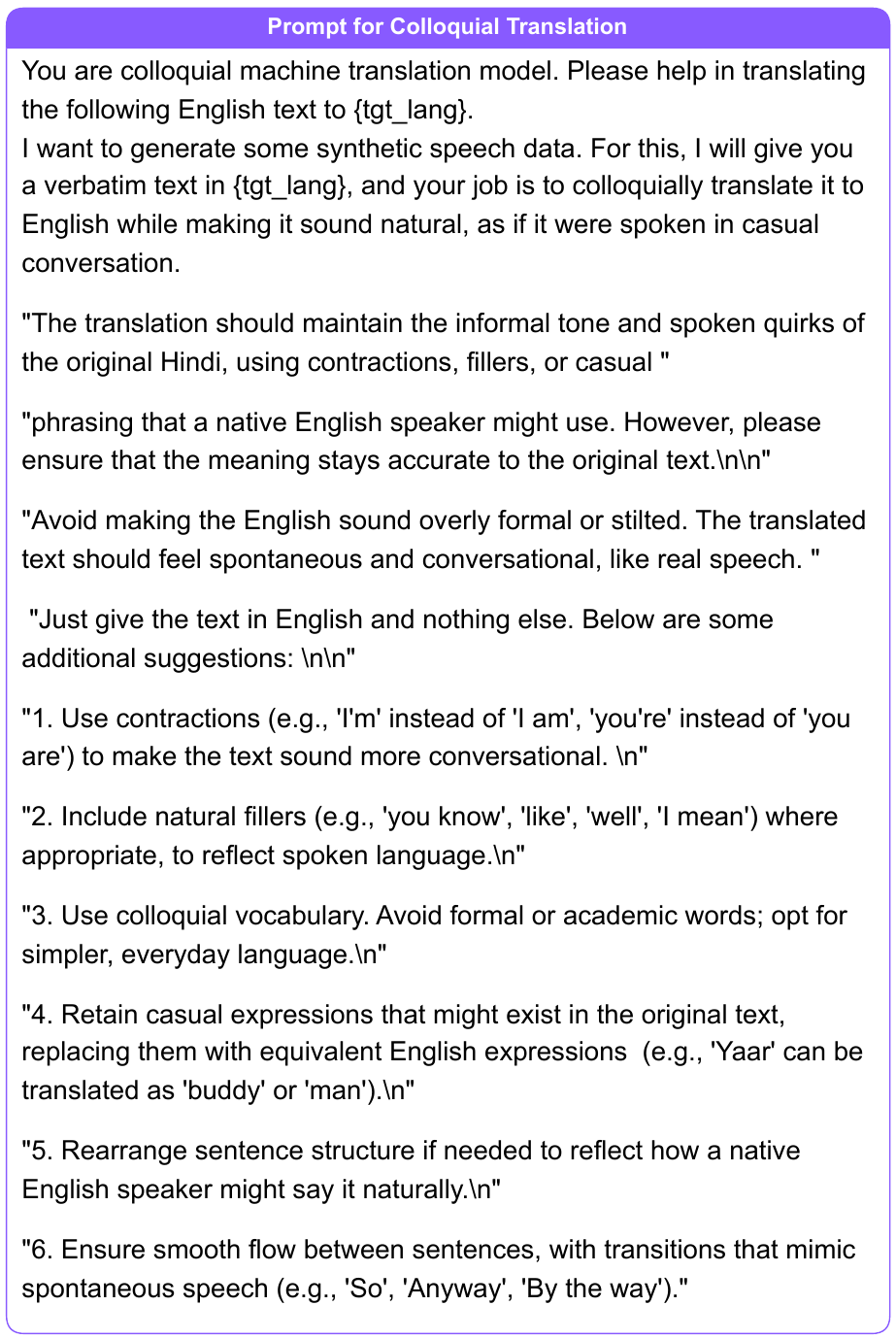}
        \caption{Prompt For Colloquial Translation}
        \label{fig:colloquial_translation}
    \end{minipage}
\end{figure*}

The prompt emphasizes four key principles: 
\begin{enumerate}
    \item \textbf{Accuracy}: Punctuation must conform to the grammatical rules of the target language.
    \item \textbf{Readability}: Insert appropriate marks (e.g., commas, periods, question marks) to enhance sentence clarity.
    \item \textbf{Consistency}: Follow the punctuation style observed in reference texts or previous examples.
    \item \textbf{Structure Preservation}: Do not add, delete, or reorder words—only punctuation should be inserted.
\end{enumerate}

To support multilingual usage, we provide language-specific reference metadata, including the expected sentence terminator. The model is instructed to return output in a clean JSON format, making it suitable for direct integration into data processing pipelines.

This prompt is especially beneficial for restoring structure in spontaneous speech transcripts, facilitating improved performance in downstream tasks such as machine translation, summarization, and language modeling where well-punctuated input is essential.

\section{Prompts Used for Synthetic Data Generation:}

\textbf{Prompt for Normal Translation} (Figure \ref{fig:normal_translation}) is structured to generate neutral, grammatically correct translations faithful to the original content. It emphasizes three key principles: (1) accuracy, ensuring no information is lost, (2) natural tone, and (3) meaningful restructuring, which encourages fluent and idiomatic English rather than literal renderings. This prompt is used to simulate conventional text translation, making the output suitable for benchmark comparison with traditional MT systems.

In contrast, \textbf{Prompt for Colloquial Translation} (Figure \ref{fig:colloquial_translation}) is tailored for generating synthetic translations that mimic spontaneous, conversational speech. It guides the model to incorporate disfluencies, fillers, contractions, and informal vocabulary, reflecting the characteristics of real-world spoken language. The prompt explicitly encourages natural phrasing and flexibility in sentence structure while preserving the core meaning. This helps in constructing datasets that better model speech patterns and are particularly valuable for training or evaluating speech-to-speech translation systems in informal contexts.

\section{Analysis of Mining Score Distributions}

\begin{figure*}[h!]
    \centering
    
    \begin{minipage}{\textwidth}
        \centering
        \begin{subfigure}[b]{0.48\textwidth}
            \centering
            \includegraphics[width=\textwidth]{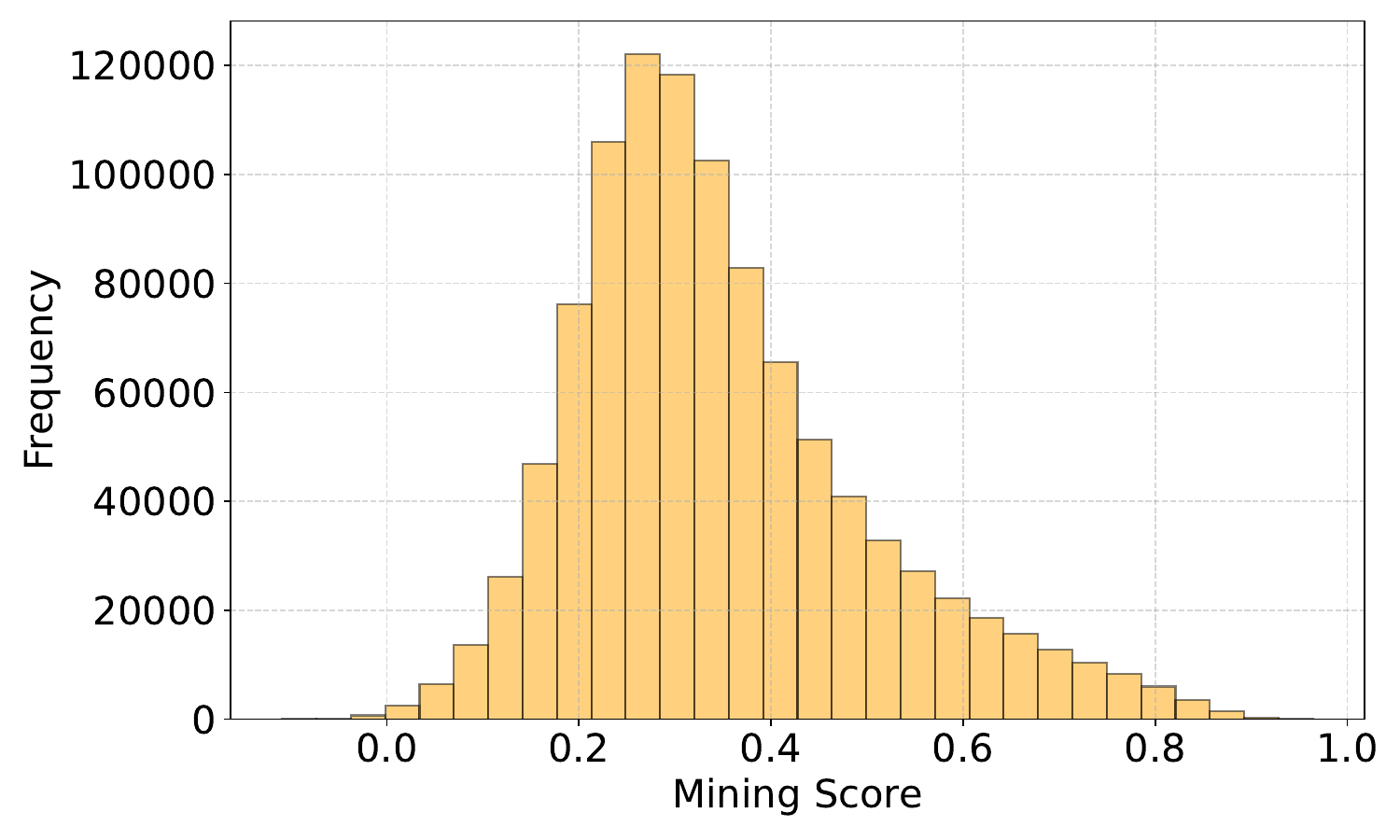}
            \caption{Seamless Align - Hindi}
            \label{fig:seamless_hindi}
        \end{subfigure}
        \hfill
        \begin{subfigure}[b]{0.48\textwidth}
            \centering
            \includegraphics[width=\textwidth]{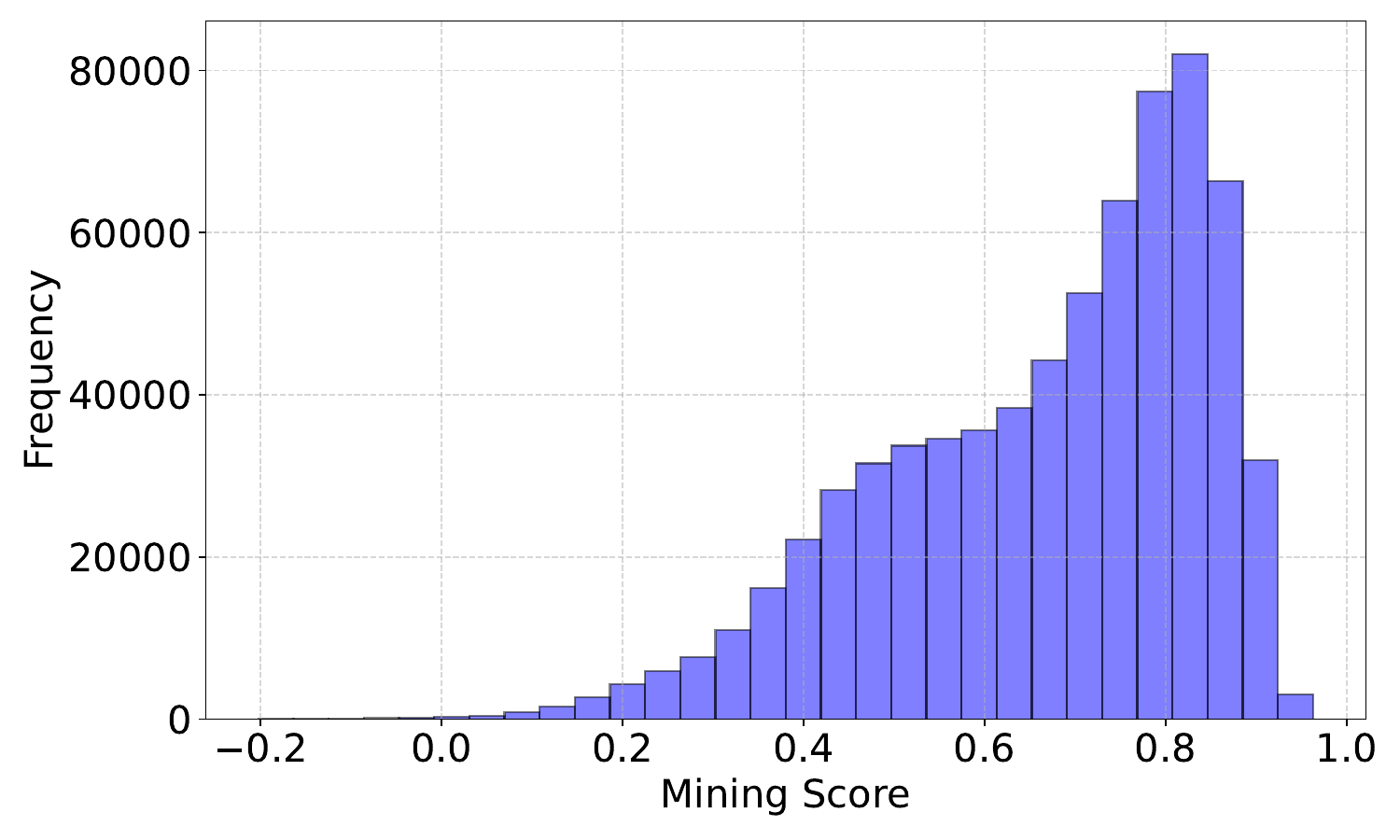}
            \caption{\dataset - Hindi}
            \label{fig:dataset_hindi}
        \end{subfigure}
    \end{minipage}
    
    
    \begin{minipage}{\textwidth}
        \centering
        \begin{subfigure}[b]{0.48\textwidth}
            \centering
            \includegraphics[width=\textwidth]{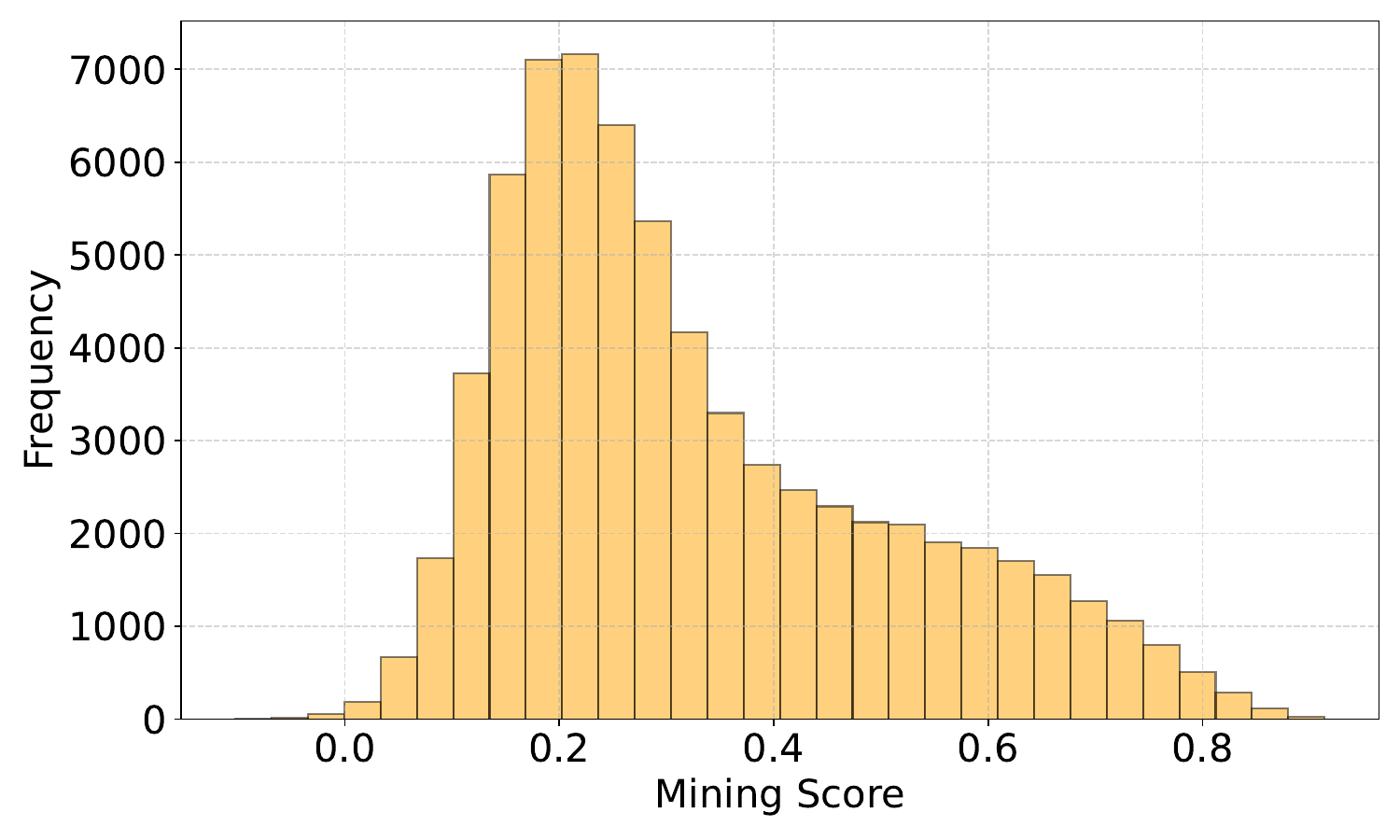}
            \caption{Seamless Align - Kannada}
            \label{fig:seamless_kannada}
        \end{subfigure}
        \hfill
        \begin{subfigure}[b]{0.48\textwidth}
            \centering
            \includegraphics[width=\textwidth]{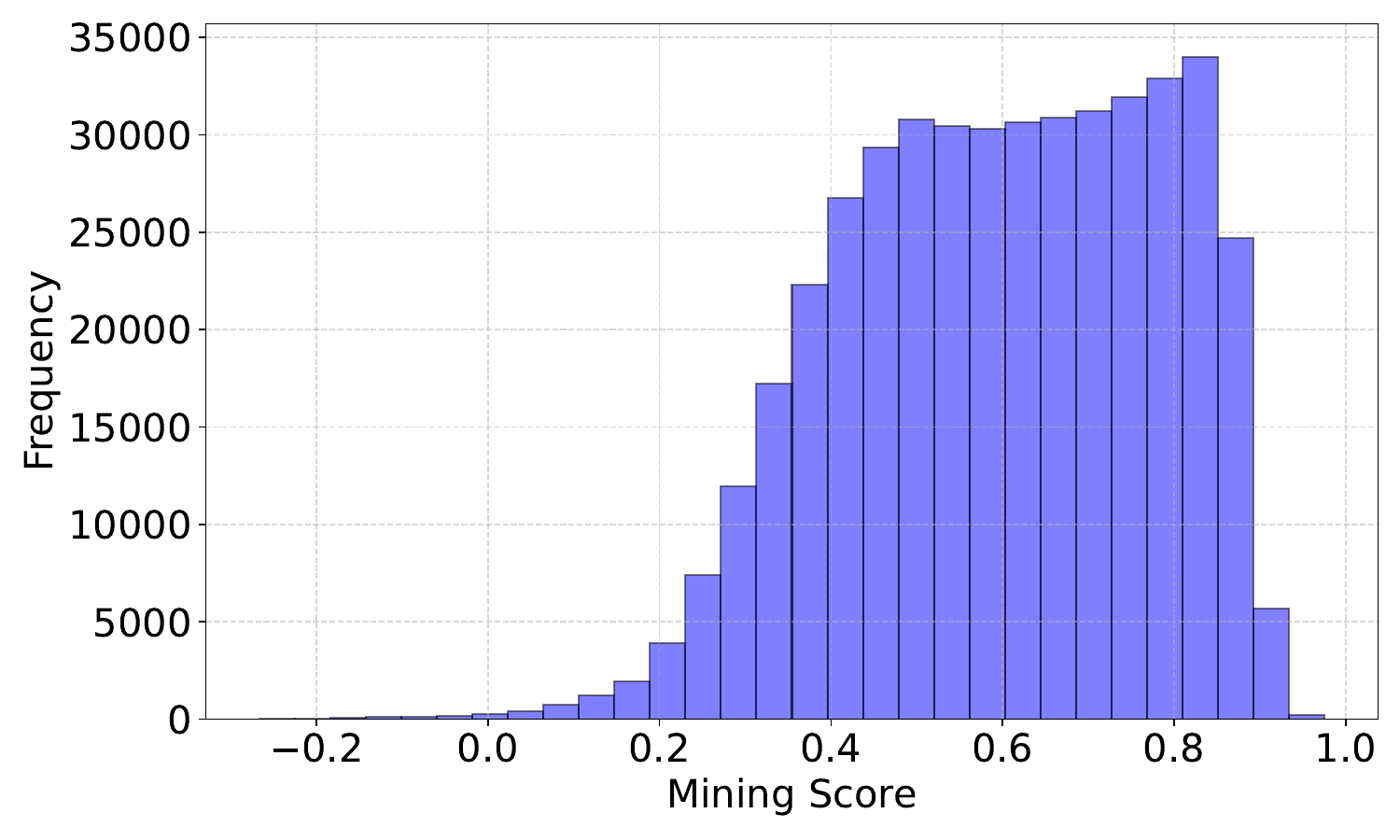}
            \caption{\dataset - Kannada}
            \label{fig:dataset_kannada}
        \end{subfigure}
    \end{minipage}
    
    
    \begin{minipage}{\textwidth}
        \centering
        \begin{subfigure}[b]{0.48\textwidth}
            \centering
            \includegraphics[width=\textwidth]{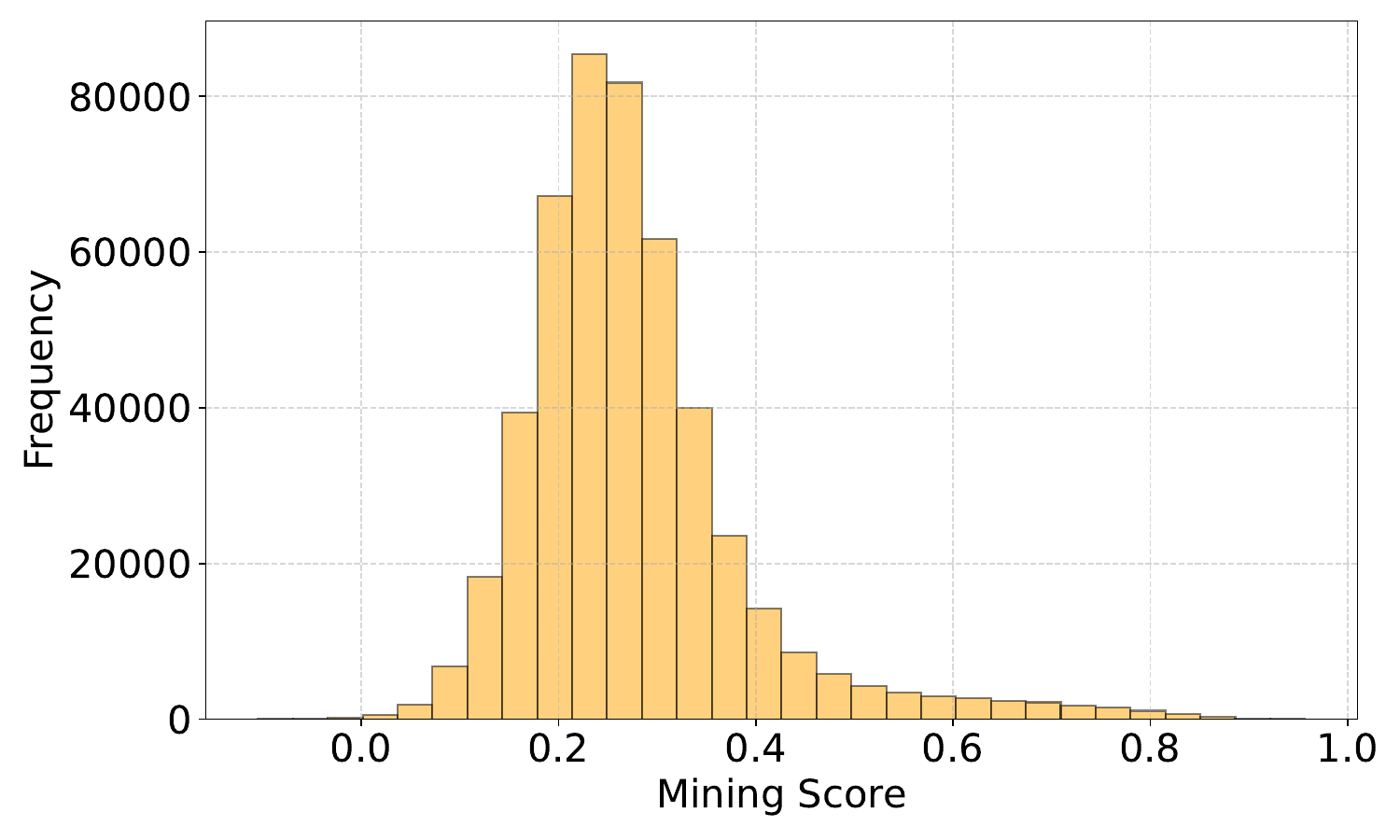}
            \caption{Seamless Align - Tamil}
            \label{fig:seamless_lang3}
        \end{subfigure}
        \hfill
        \begin{subfigure}[b]{0.48\textwidth}
            \centering
            \includegraphics[width=\textwidth]{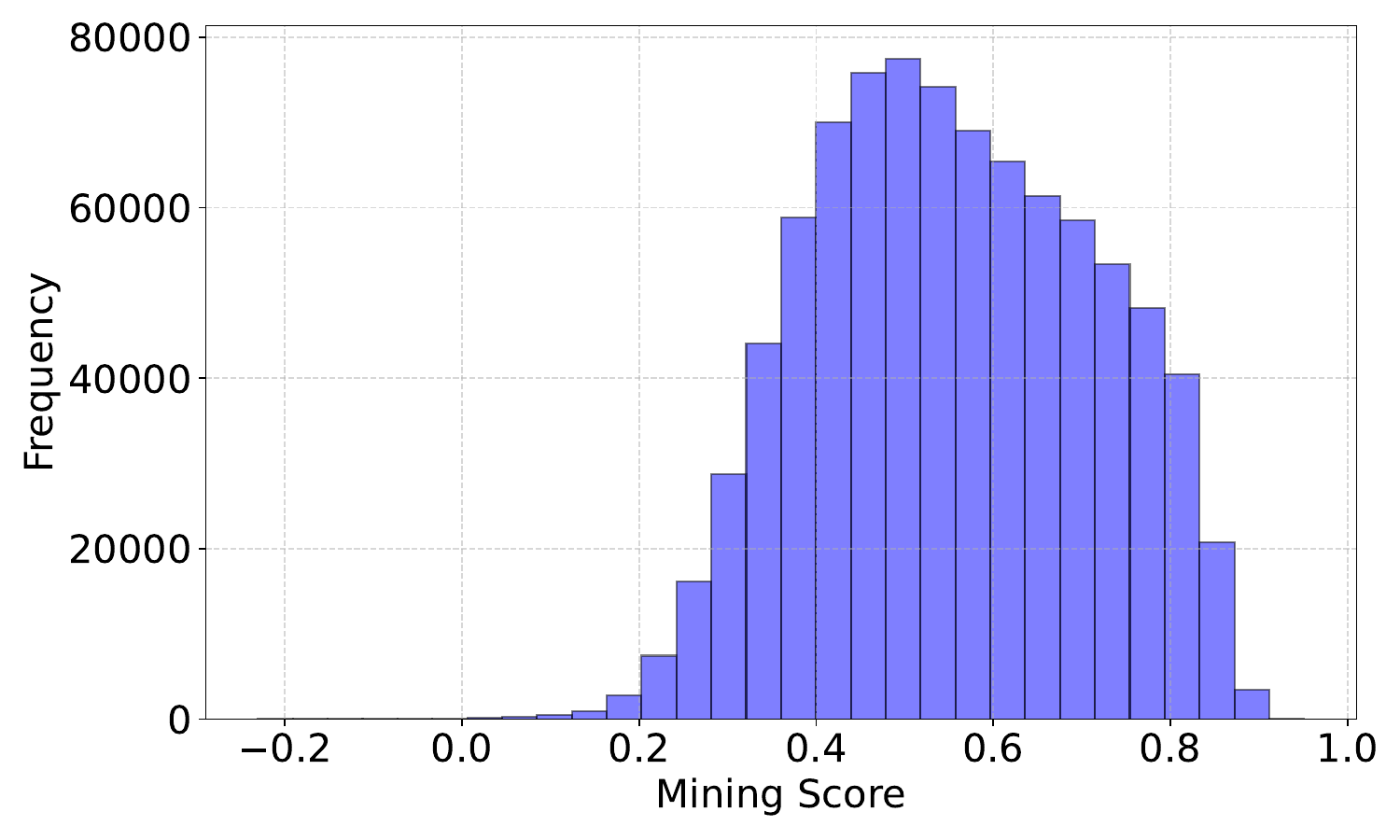}
            \caption{\dataset - Tamil}
            \label{fig:dataset_tamil}
        \end{subfigure}
    \end{minipage}
    
    \begin{minipage}{\textwidth}
        \centering
        \begin{subfigure}[b]{0.48\textwidth}
            \centering
            \includegraphics[width=\textwidth]{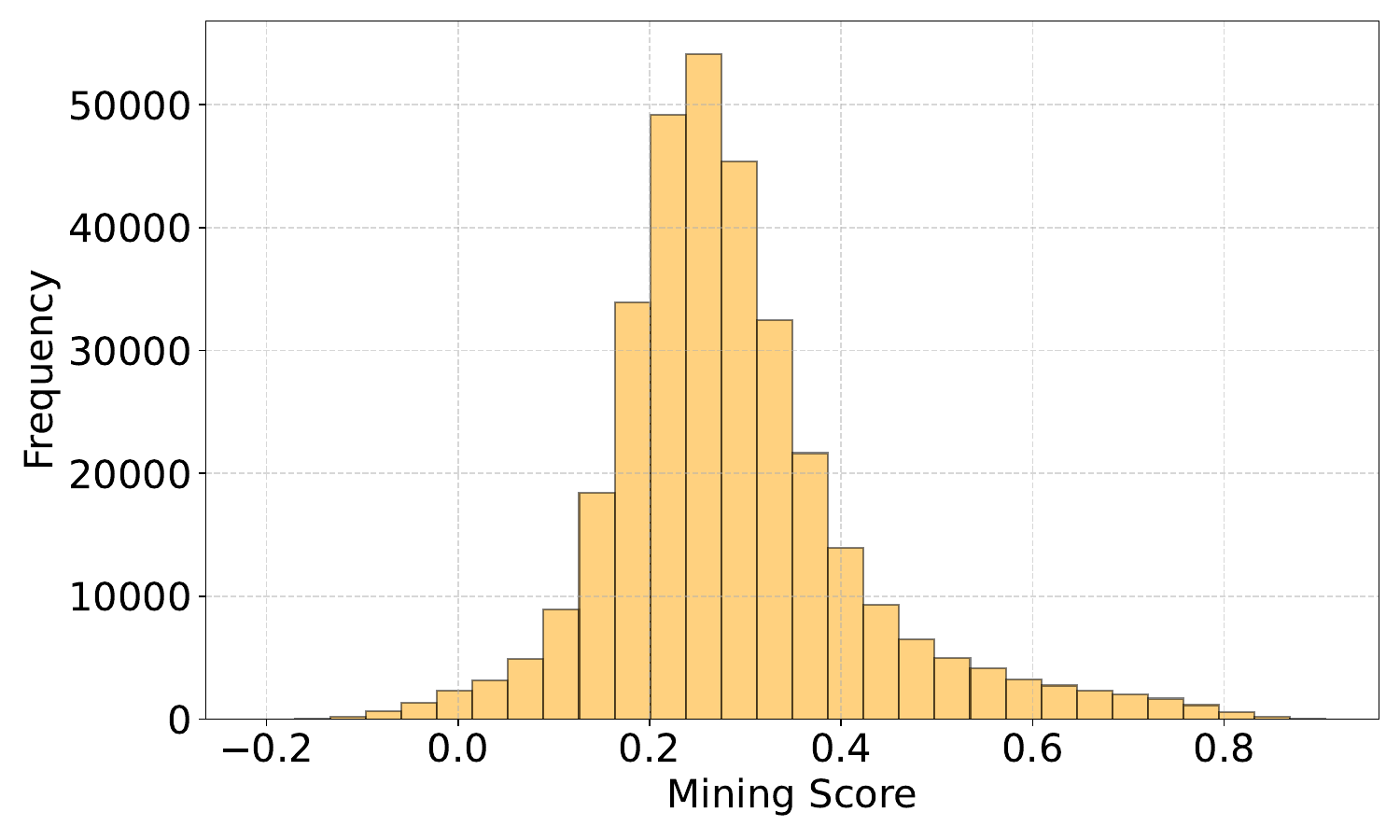}
            \caption{Seamless Align - Telugu}
            \label{fig:seamless_lang4}
        \end{subfigure}
        \hfill
        \begin{subfigure}[b]{0.48\textwidth}
            \centering
            \includegraphics[width=\textwidth]{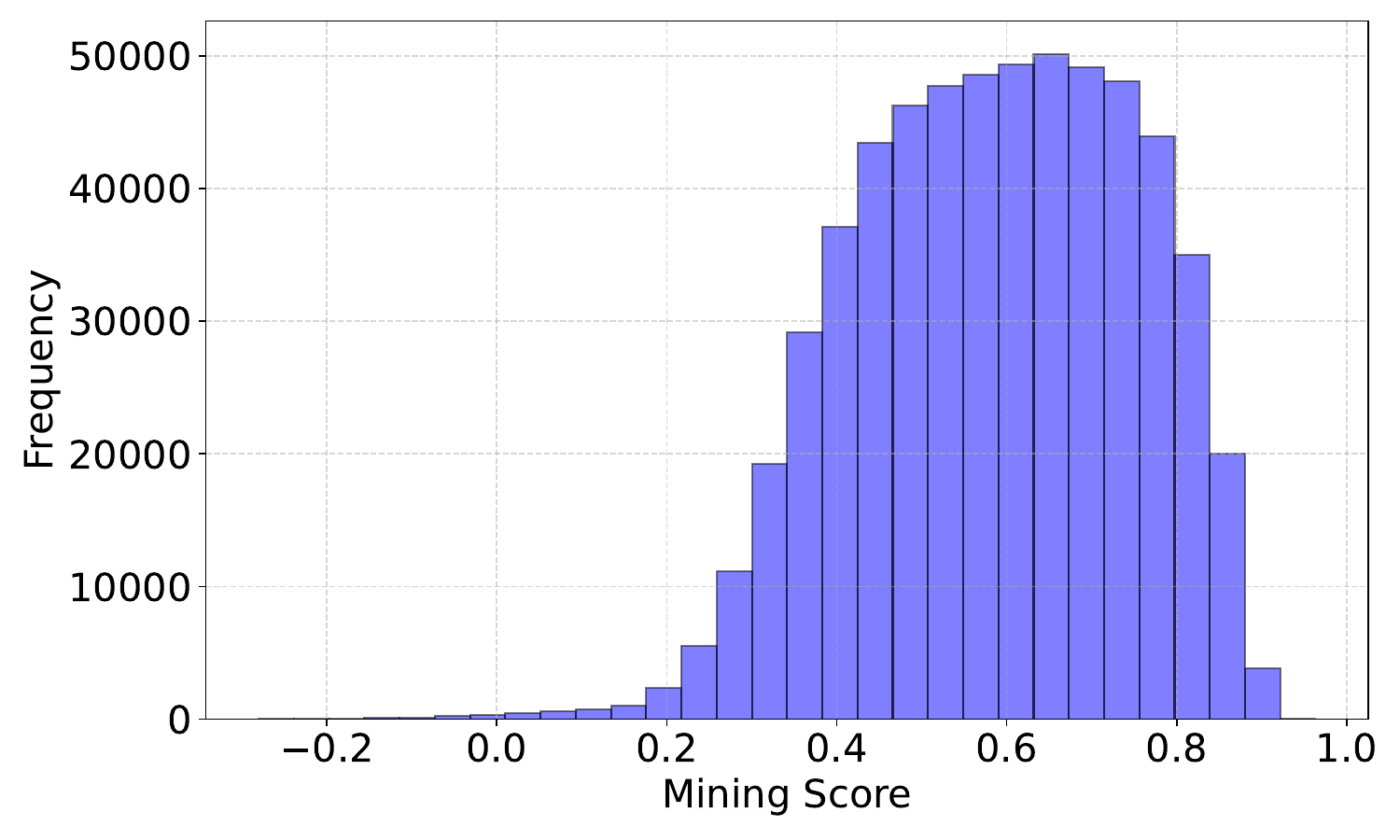}
            \caption{\dataset - Telugu}
            \label{fig:dataset_telugu}
        \end{subfigure}
    \end{minipage}
    
    
    
    \caption{Distribution of SONAR mining scores for XX $\rightarrow$ En sentence pairs from the SeamlessAlign (left) and \dataset~ (right) datasets across four languages: Hindi, Kannada, Tamil, and Telugu. \dataset~ consistently exhibits a higher concentration of high-quality alignments, with mining scores skewed toward the upper end of the scale. In contrast, SeamlessAlign distributions are centered around lower mining scores, indicating overall lower alignment quality. This comparison underscores the improved precision of alignments in \dataset~.}
    \label{fig:lang_dist_1}
\end{figure*}

\begin{figure*}[h!]    
    \begin{minipage}{\textwidth}
        \centering
        \begin{subfigure}[b]{0.48\textwidth}
            \centering
            \includegraphics[width=\textwidth]{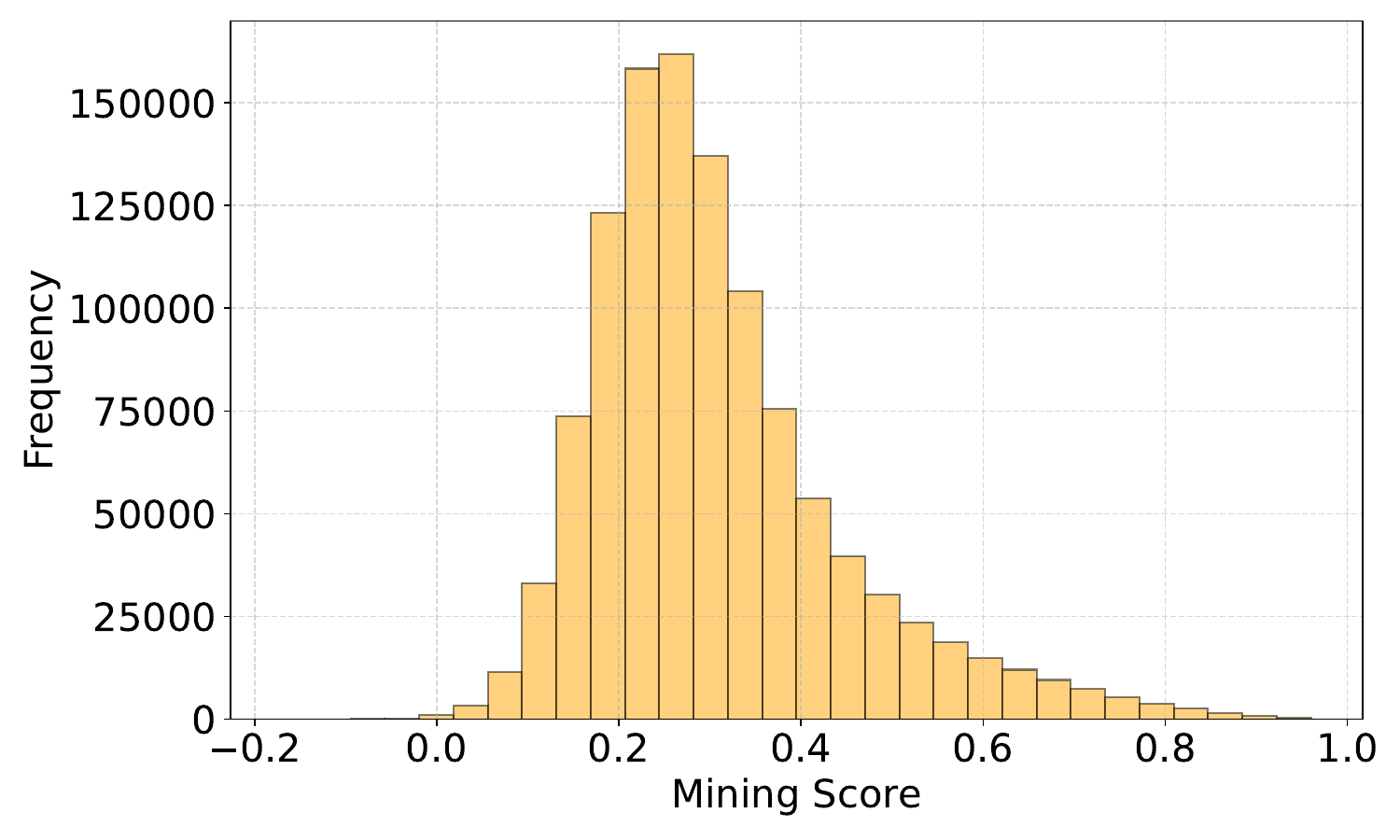}
            \caption{Seamless Align - Urdu}
            \label{fig:seamless_urdu}
        \end{subfigure}
        \hfill
        \begin{subfigure}[b]{0.48\textwidth}
            \centering
            \includegraphics[width=\textwidth]{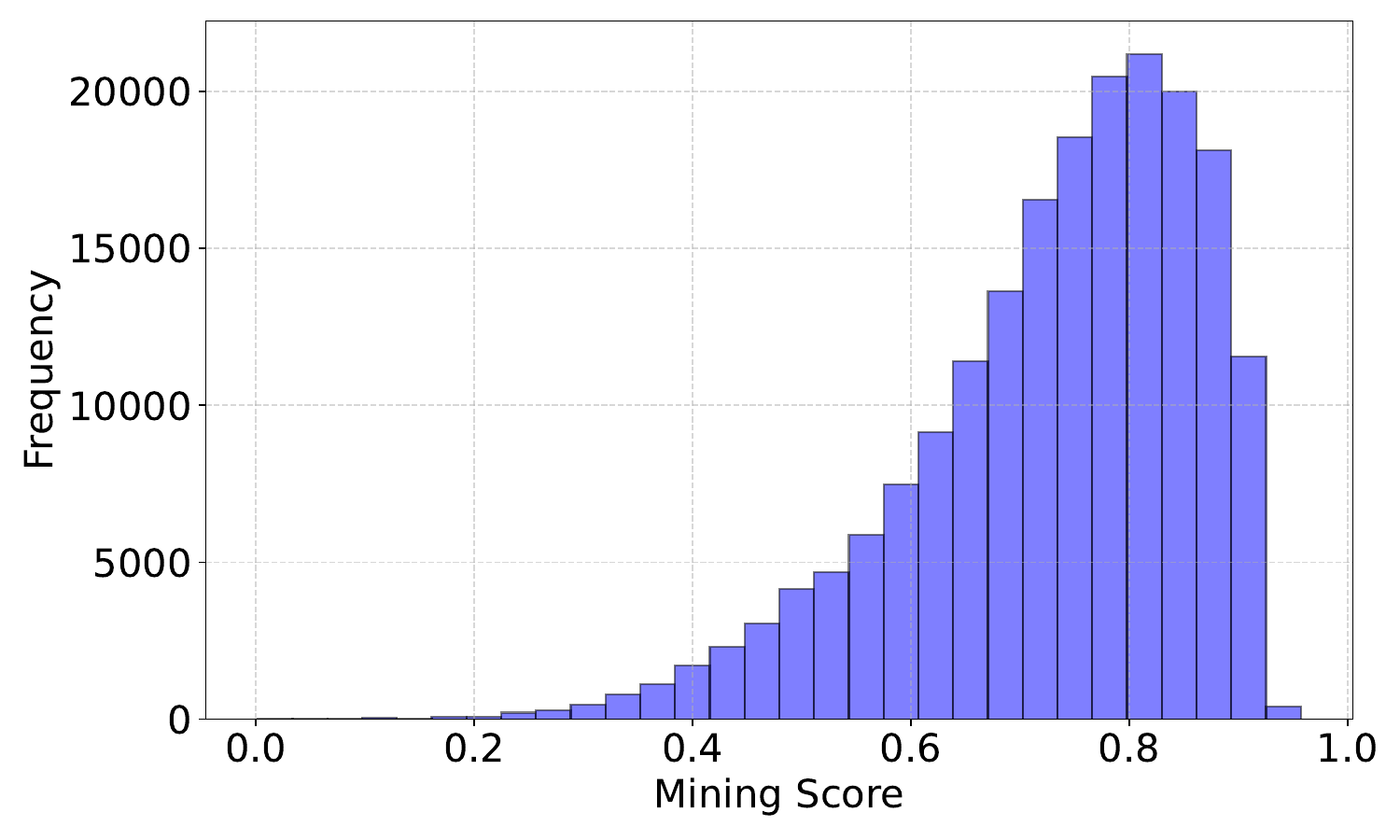}
            \caption{\dataset - Urdu}
            \label{fig:dataset_urdu}
        \end{subfigure}
    \end{minipage}
    
    \caption{Distribution of SONAR mining scores for XX $\rightarrow$ En sentence pairs from the SeamlessAlign (left) and \dataset~ (right) datasets across four languages: Hindi, Kannada, Tamil, and Telugu. \dataset~ consistently exhibits a higher concentration of high-quality alignments, with mining scores skewed toward the upper end of the scale. In contrast, SeamlessAlign distributions are centered around lower mining scores, indicating overall lower alignment quality.}
    \label{fig:lang_dist_2}

    \begin{minipage}{\textwidth}
        \centering
        \begin{subfigure}[b]{0.48\textwidth}
            \centering
            \includegraphics[width=\textwidth]{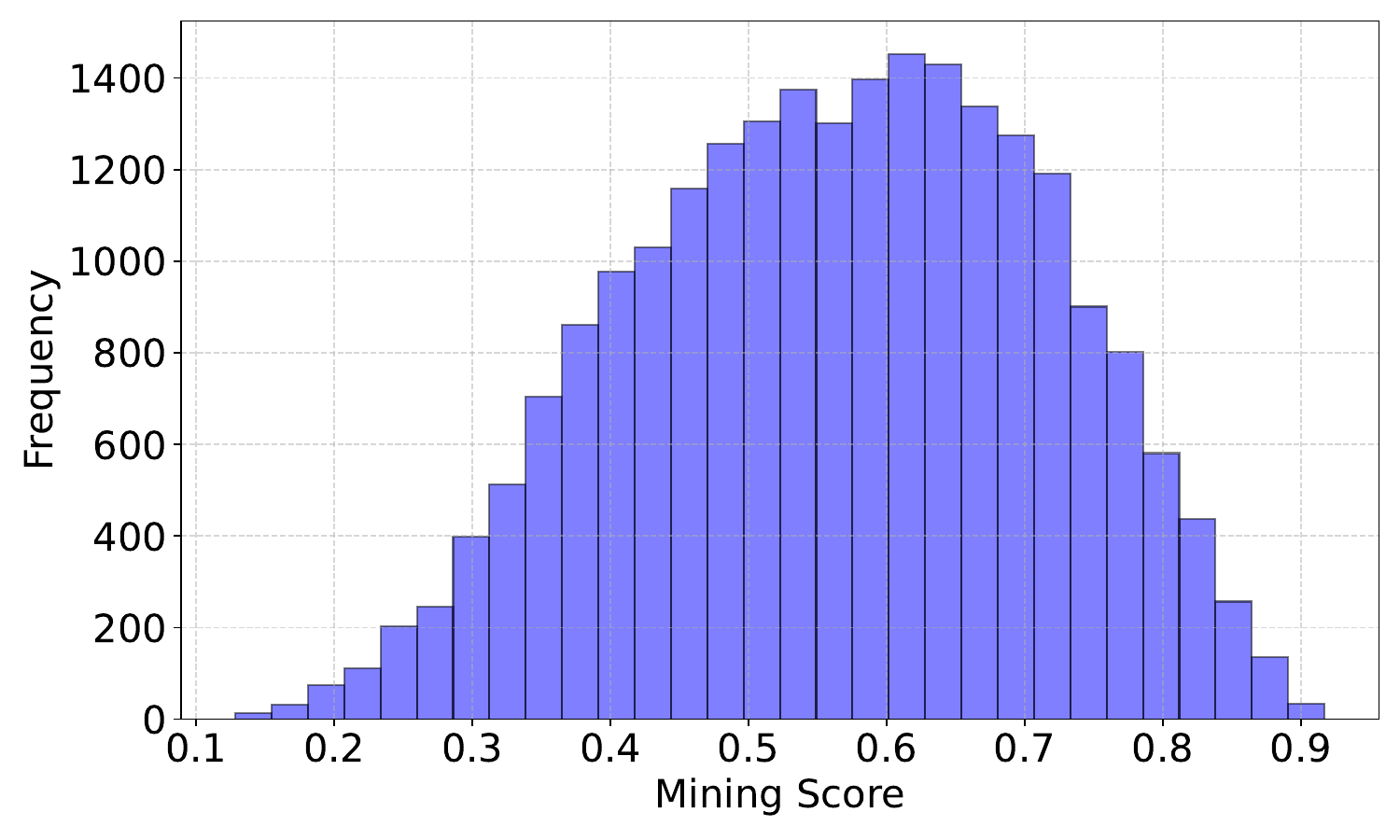}
            \caption{\centering English - Assamese}
            \label{fig:dataset_eng_as}
        \end{subfigure}
        \hfill
        \begin{subfigure}[b]{0.48\textwidth}
            \centering
            \includegraphics[width=\textwidth]{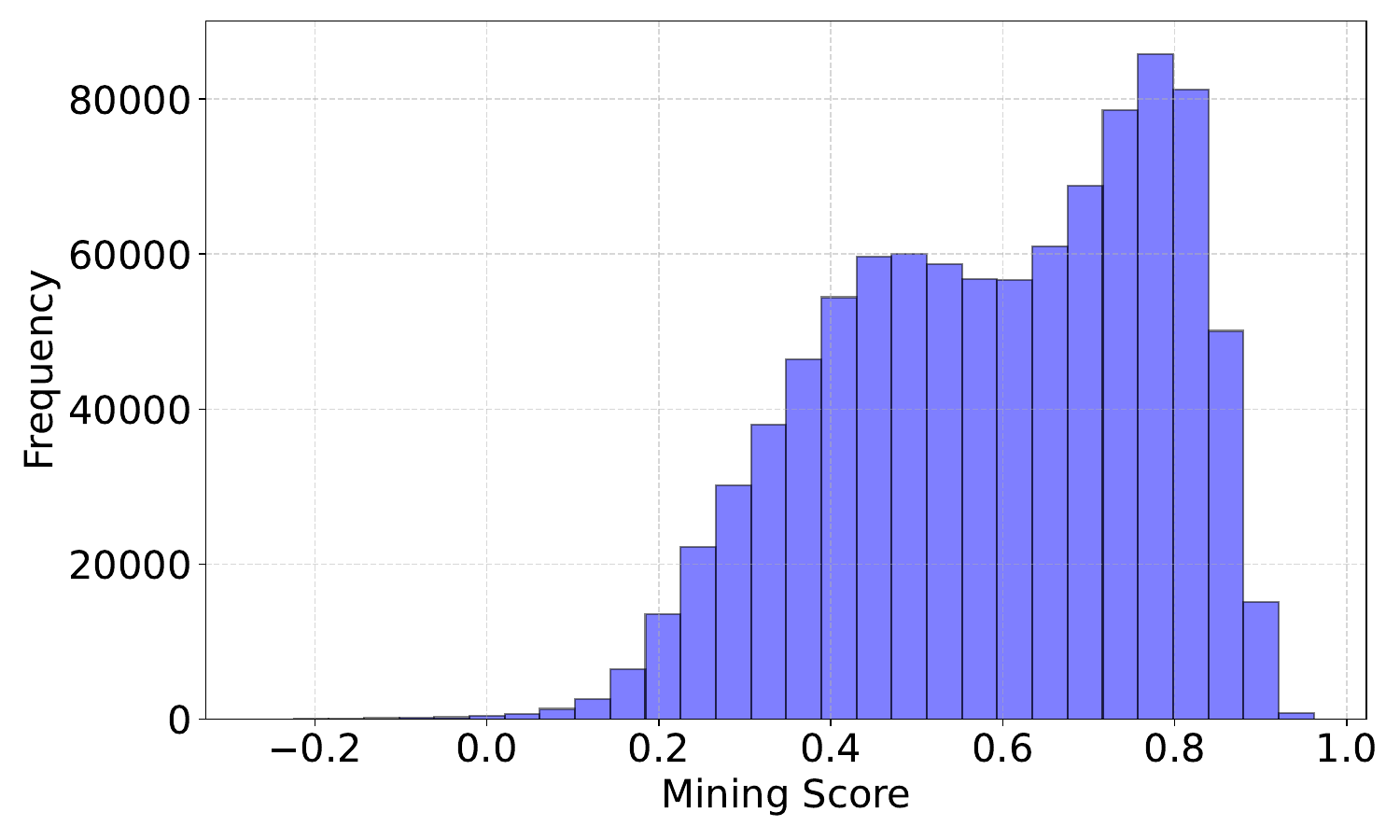}
            \caption{\centering English - Bengali}
            \label{fig:dataset_eng_bn}
        \end{subfigure}
    \end{minipage}
    \label{fig:plot_comparison_pairs_eng_to_indic}

    \begin{minipage}{\textwidth}
        \centering
        \begin{subfigure}[b]{0.48\textwidth}
            \centering
            \includegraphics[width=\textwidth]{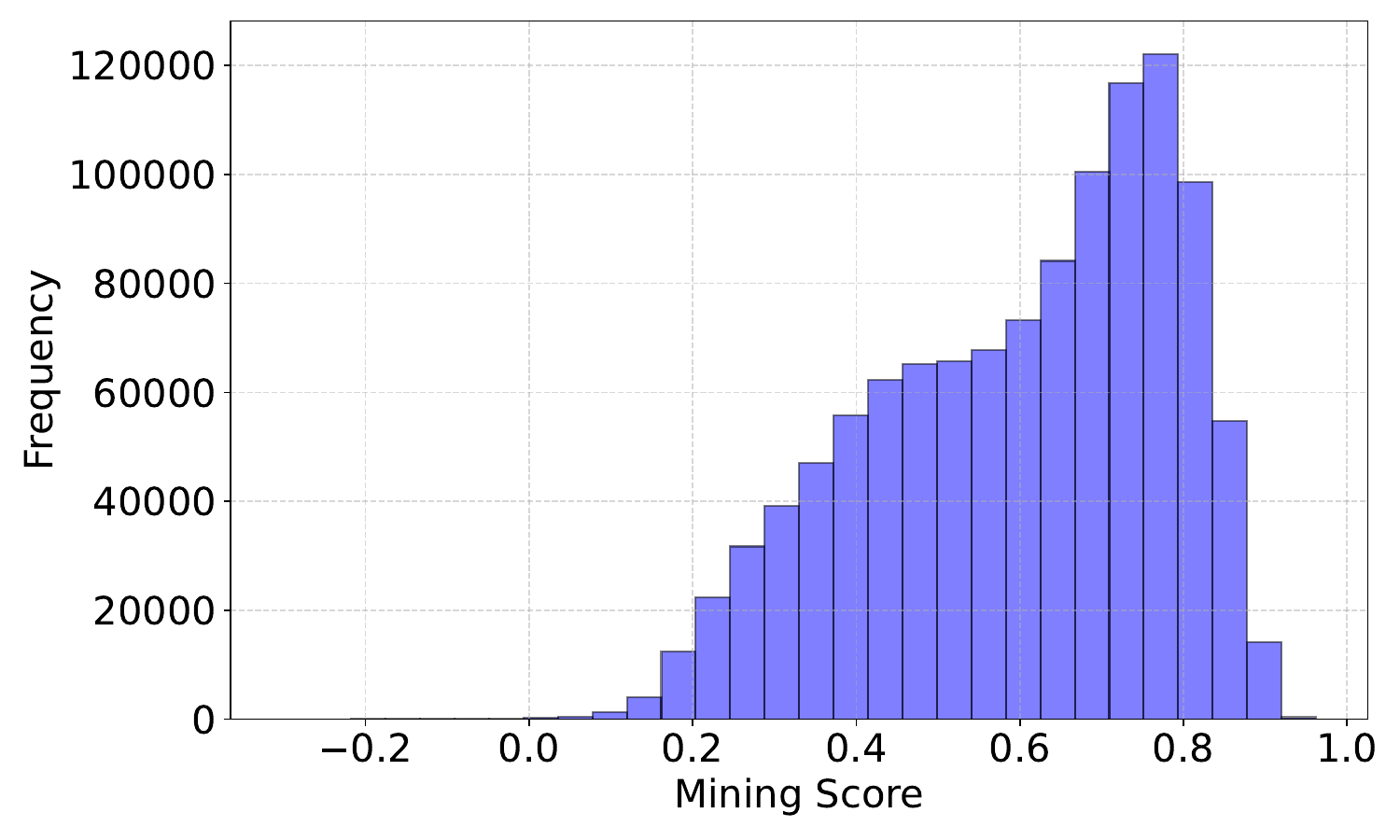}
            \caption{\centering English - Gujarati}
            \label{fig:dataset_eng_gu}
        \end{subfigure}
        \hfill
        \begin{subfigure}[b]{0.48\textwidth}
            \centering
            \includegraphics[width=\textwidth]{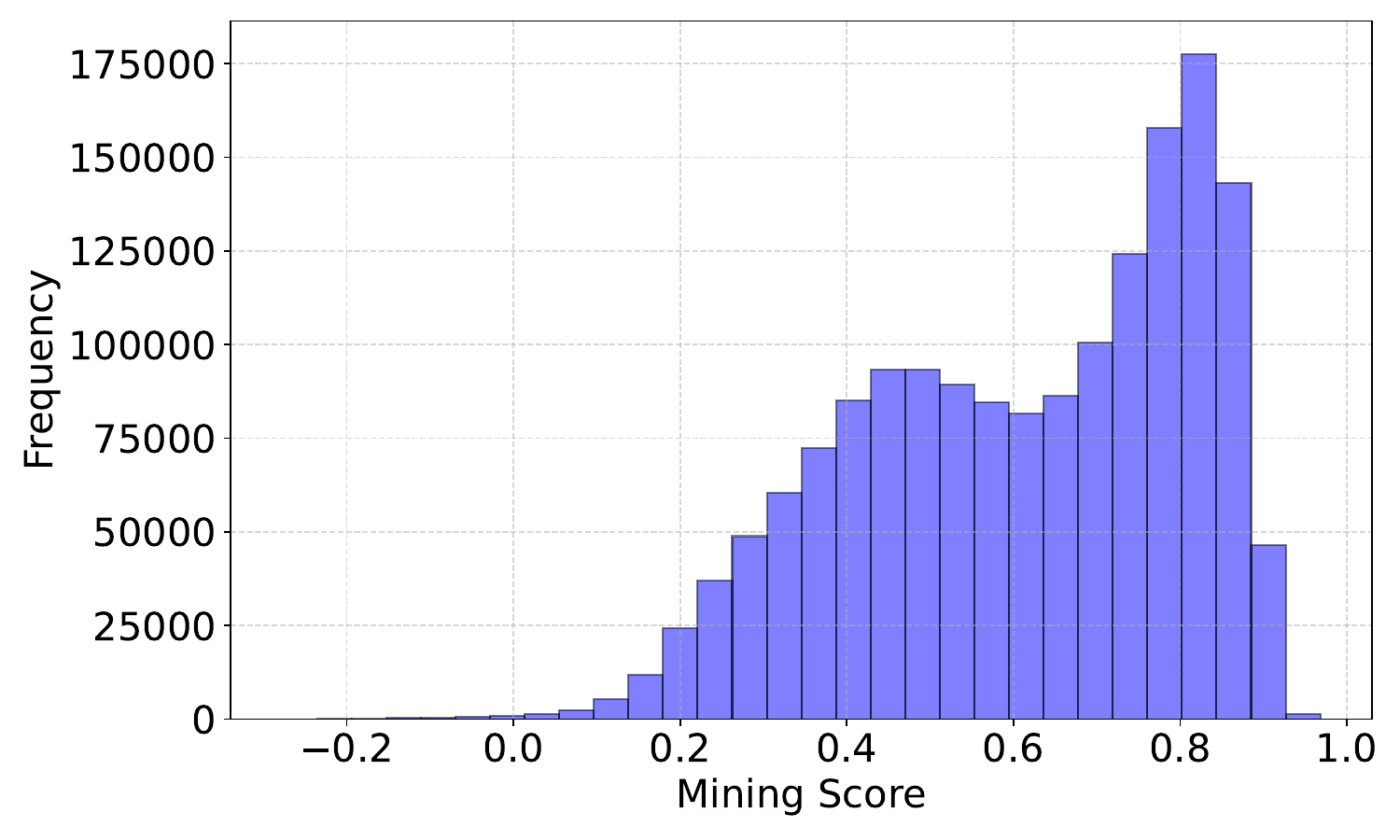}
            \caption{\centering English - Hindi}
            \label{fig:dataset_eng_hi}
        \end{subfigure}
    \end{minipage}
    
    \label{fig:plot_comparison_pairs_eng_gu_hi}
    \begin{minipage}{\textwidth}
        \centering
        \begin{subfigure}[b]{0.48\textwidth}
            \centering
            \includegraphics[width=\textwidth]{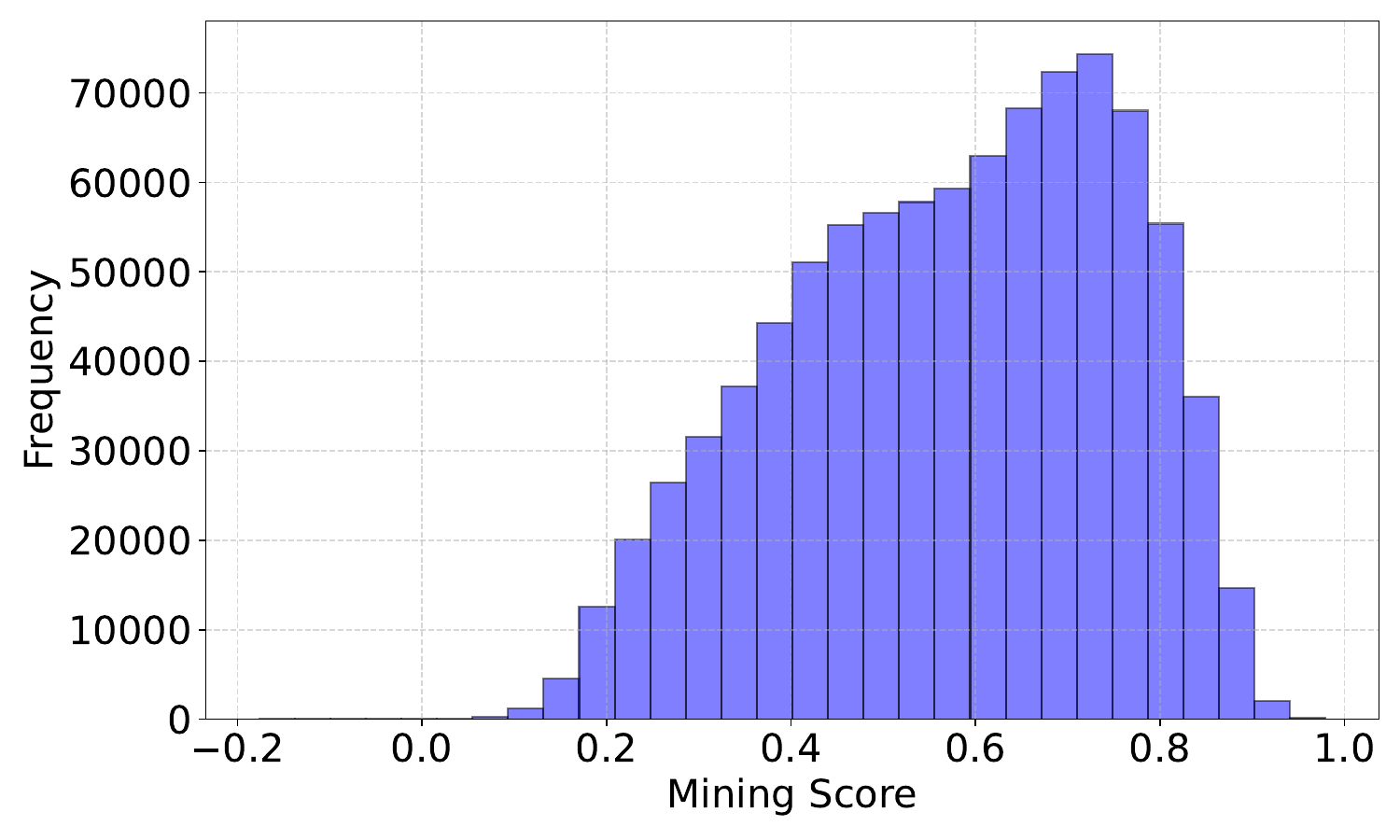}
            \caption{\centering English - Kannada}
            \label{fig:dataset_eng_kn}
        \end{subfigure}
        \hfill
        \begin{subfigure}[b]{0.48\textwidth}
            \centering
            \includegraphics[width=\textwidth]{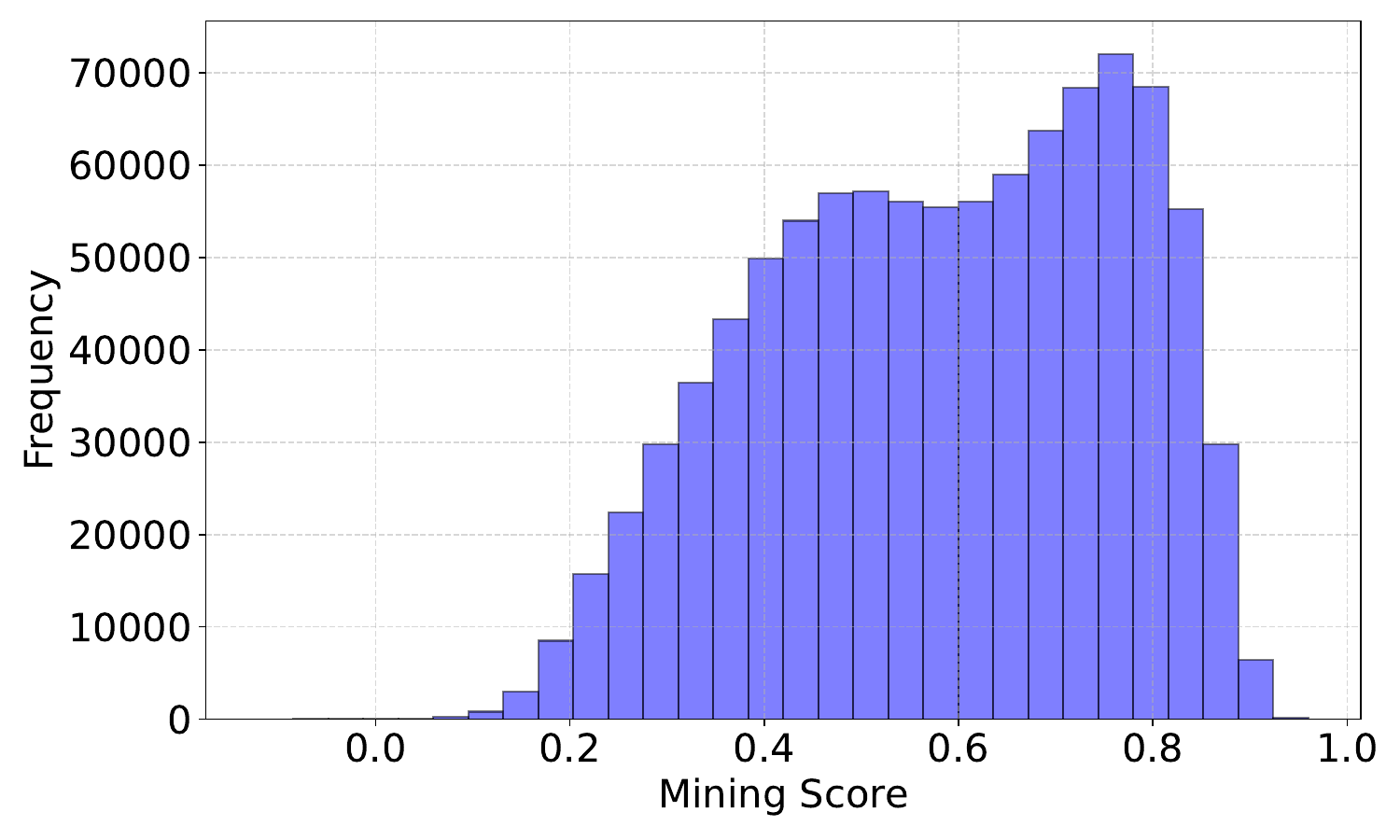}
            \caption{\centering English - Malayalam}
            \label{fig:dataset_eng_ml}
        \end{subfigure}
    \end{minipage}
    
    \caption{Histogram of SONAR mining scores for En $\rightarrow$ XX language pairs in the \dataset~ dataset. The distributions indicate a strong skew toward high-quality alignments (scores above 0.6) across most language pairs, reflecting the effectiveness of the mining and filtering pipeline used in ~\dataset.}
    \label{fig:mining_histogram_ba_1}
\end{figure*}

\begin{figure*}[h!]
    \centering

    \begin{minipage}{\textwidth}
        \centering
        \begin{subfigure}[b]{0.48\textwidth}
            \centering
            \includegraphics[width=\textwidth]{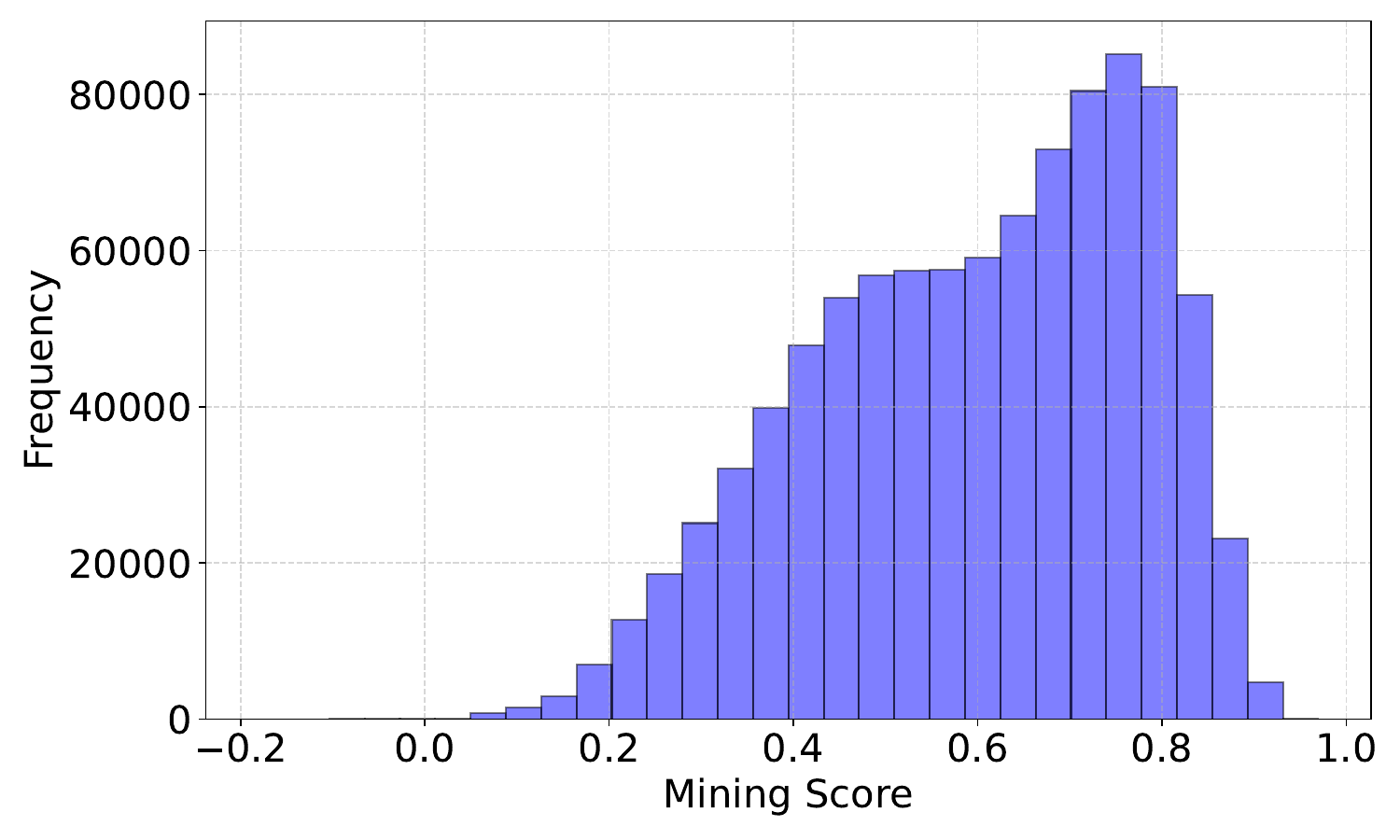}
            \caption{\centering English - Marathi}
            \label{fig:dataset_eng_mr}
        \end{subfigure}
        \hfill
        \begin{subfigure}[b]{0.48\textwidth}
            \centering
            \includegraphics[width=\textwidth]{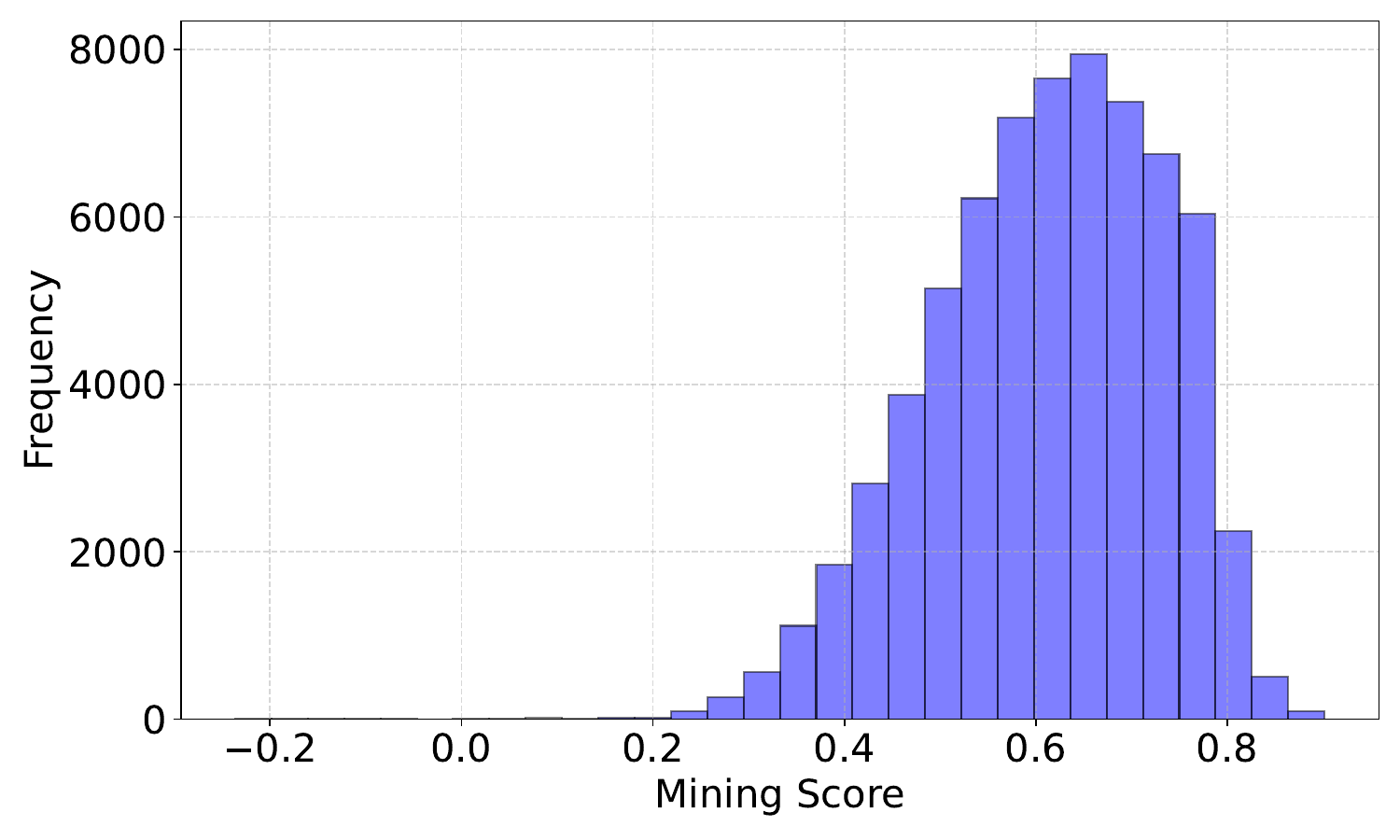}
            \caption{\centering English - Nepali}
            \label{fig:dataset_eng_ne}
        \end{subfigure}
    \end{minipage}
    
    \label{fig:plot_analysis_pairs_eng_mr_ne}

    \centering

    \begin{minipage}{\textwidth}
        \centering
        \begin{subfigure}[b]{0.48\textwidth}
            \centering
            \includegraphics[width=\textwidth]{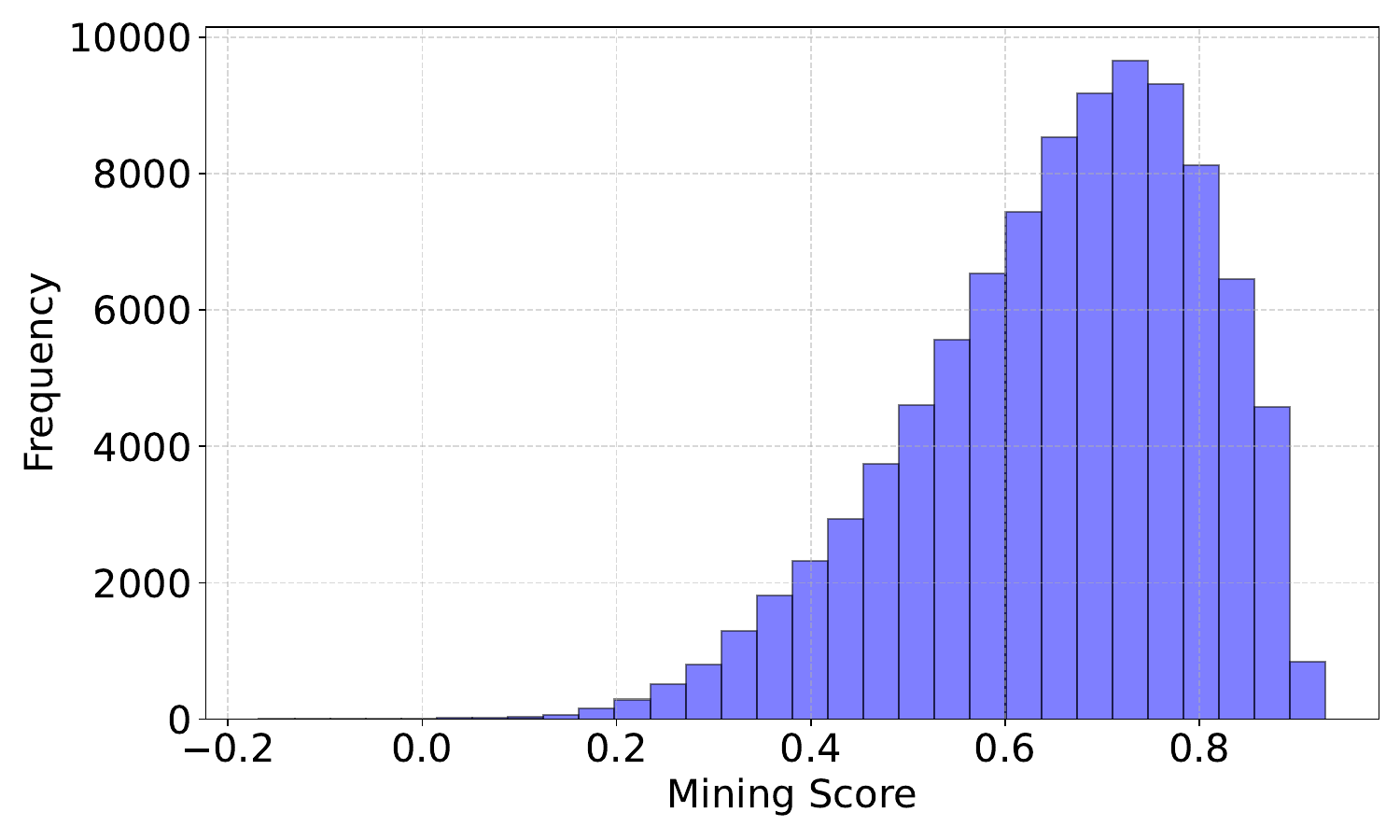}
            \caption{\centering English - Odia}
            \label{fig:dataset_eng_or}
        \end{subfigure}
        \hfill
        \begin{subfigure}[b]{0.48\textwidth}
            \centering
            \includegraphics[width=\textwidth]{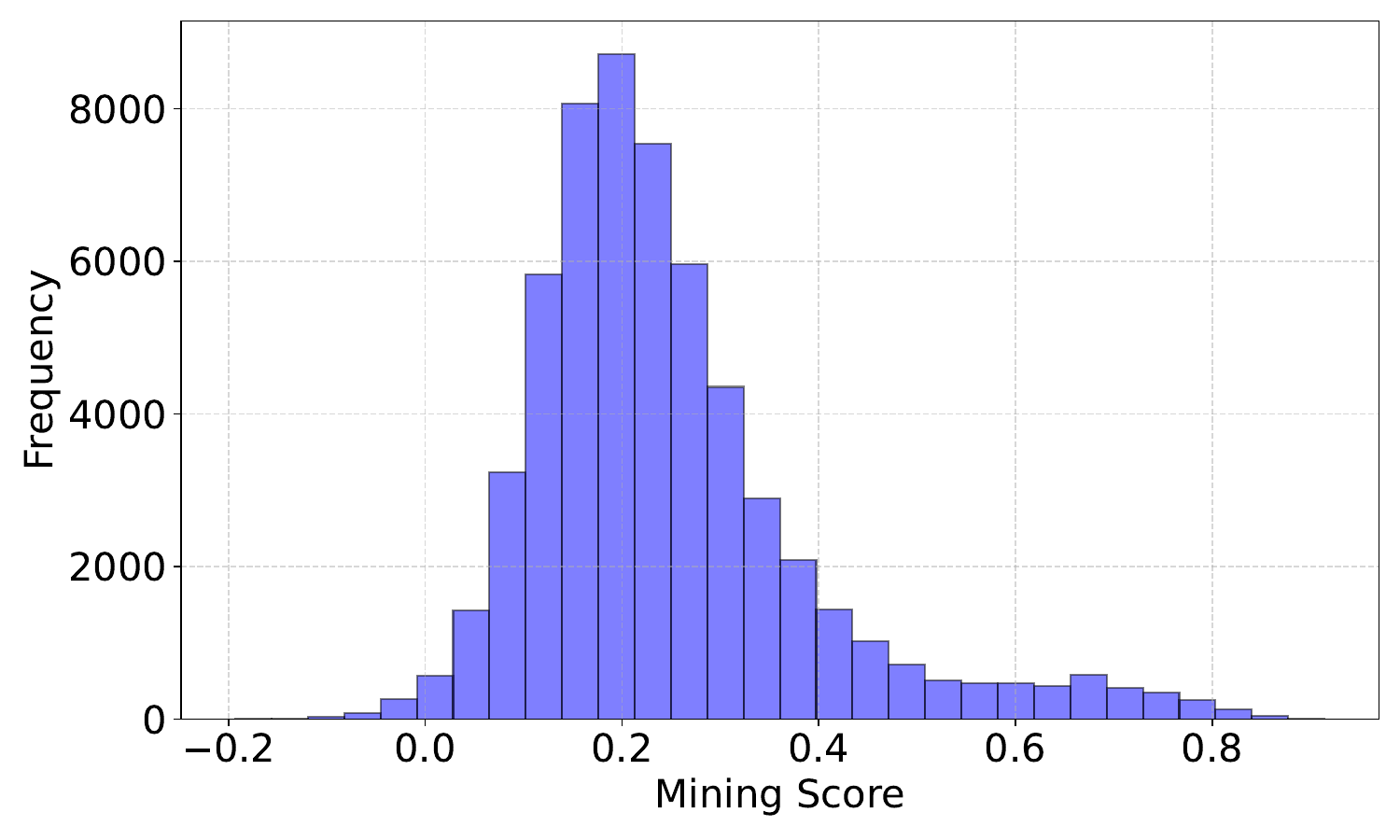}
            \caption{\centering English - Punjabi}
            \label{fig:dataset_eng_pa}
        \end{subfigure}
    \end{minipage}
    
    \label{fig:plot_analysis_pairs_eng_or_pa}

    \centering

    \begin{minipage}{\textwidth}
        \centering
        \begin{subfigure}[b]{0.48\textwidth}
            \centering
            \includegraphics[width=\textwidth]{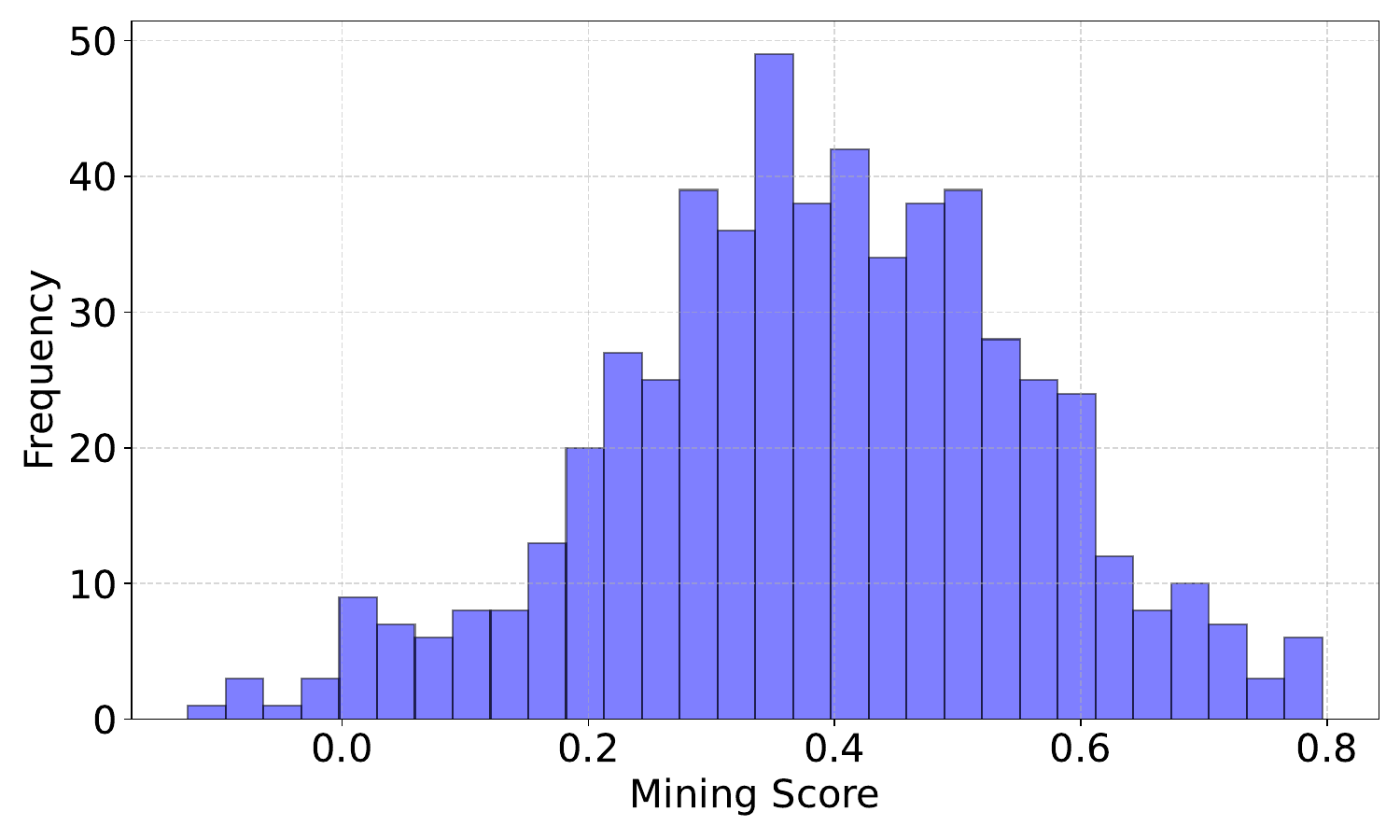}
            \caption{\centering English - Sindhi}
            \label{fig:dataset_eng_sd}
        \end{subfigure}
        \hfill
        \begin{subfigure}[b]{0.48\textwidth}
            \centering
            \includegraphics[width=\textwidth]{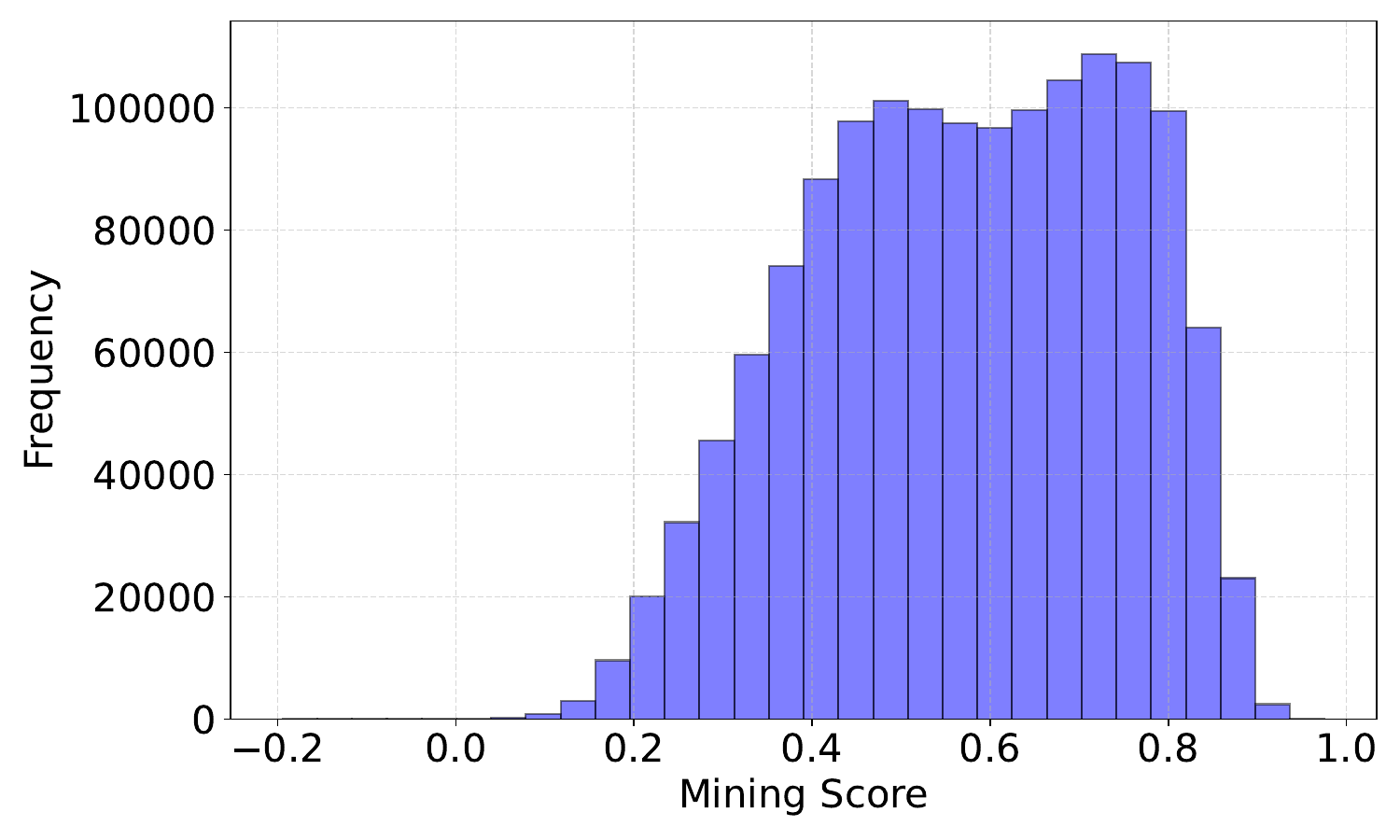}
            \caption{\centering English - Tamil}
            \label{fig:dataset_eng_ta}
        \end{subfigure}
    \end{minipage}
    
    \label{fig:plot_analysis_pairs_eng_sd_ta}

    \centering

    \begin{minipage}{\textwidth}
        \centering
        \begin{subfigure}[b]{0.48\textwidth}
            \centering
            \includegraphics[width=\textwidth]{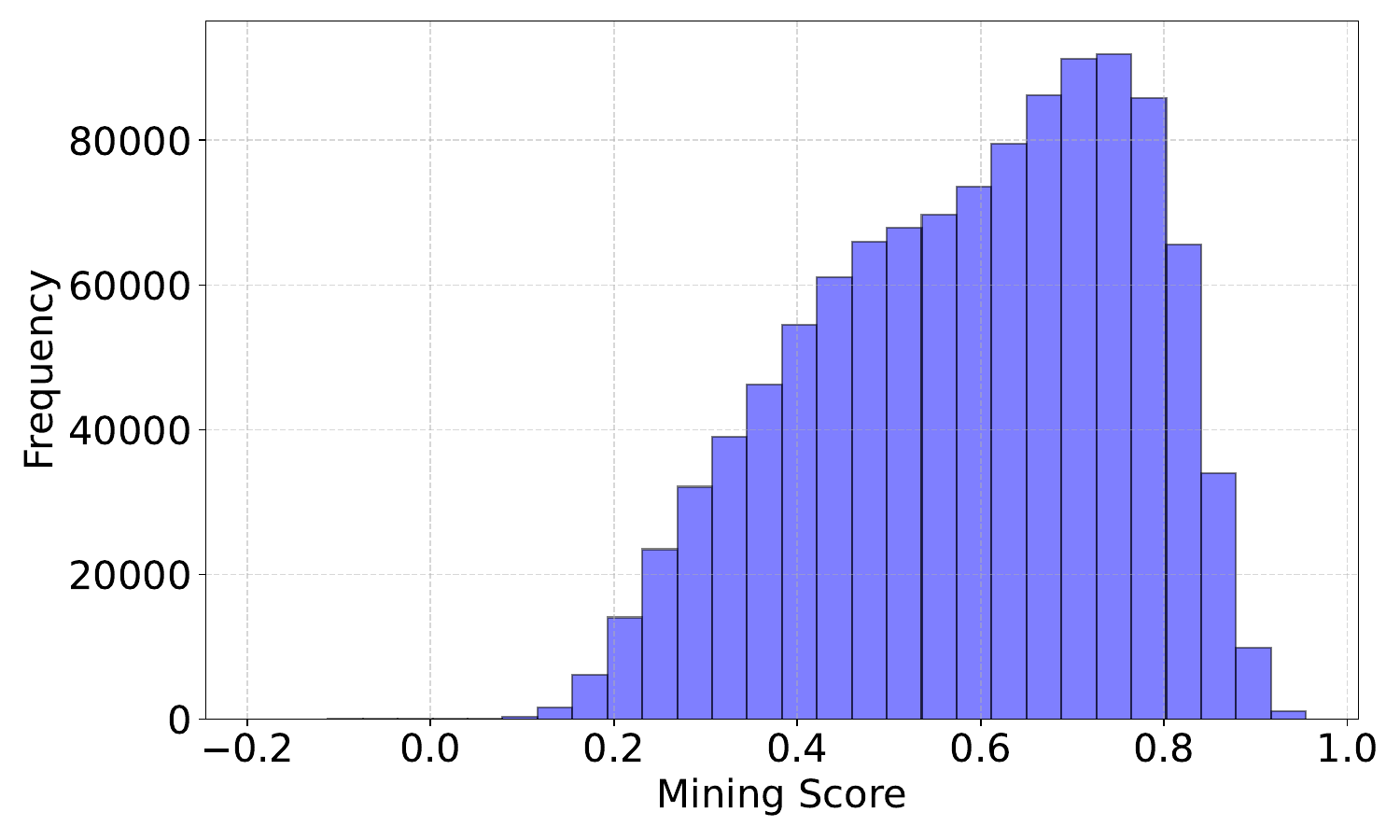}
            \caption{\centering English - Telugu}
            \label{fig:dataset_eng_te}
        \end{subfigure}
        \hfill
        \begin{subfigure}[b]{0.48\textwidth}
            \centering
            \includegraphics[width=\textwidth]{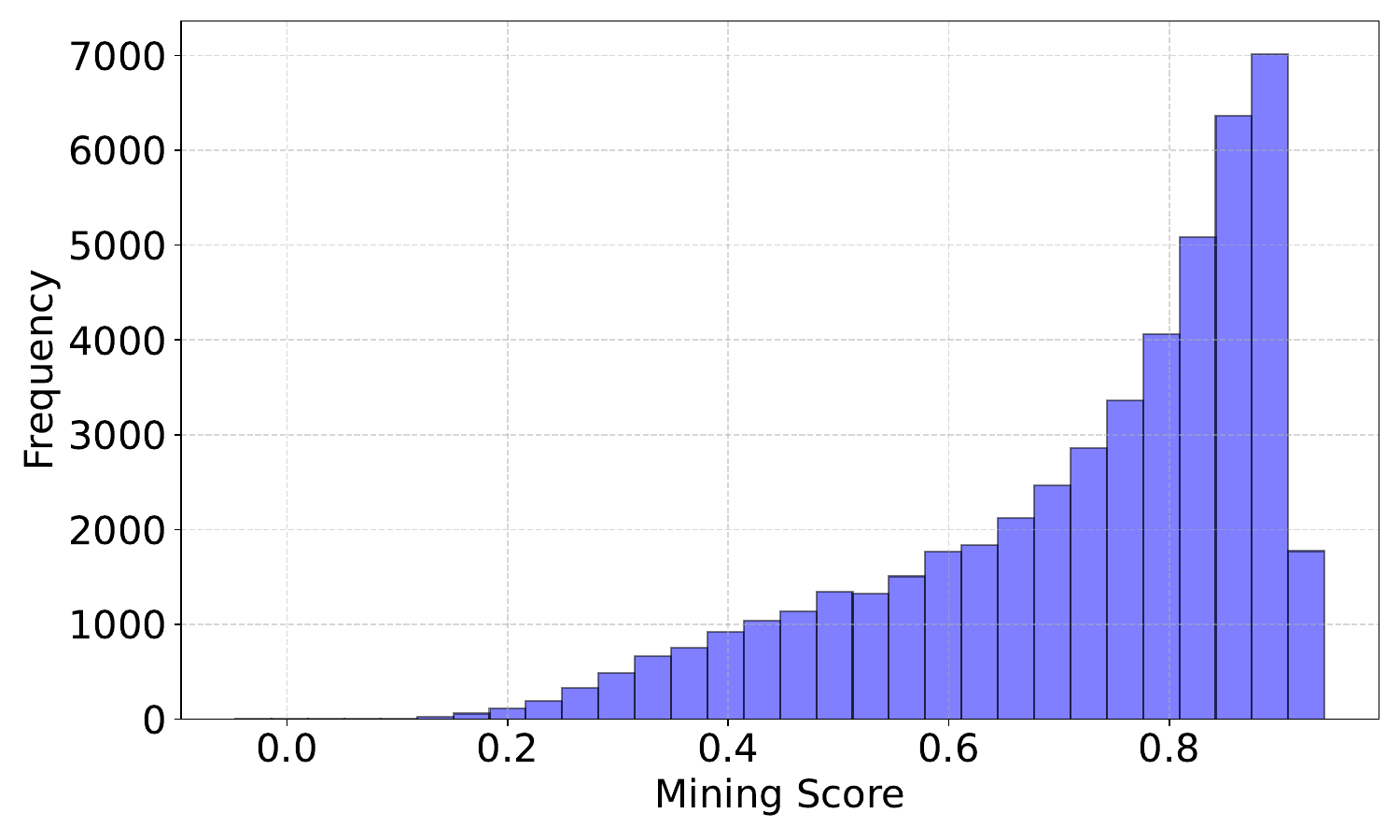}
            \caption{\centering English - Urdu}
            \label{fig:dataset_eng_ur}
        \end{subfigure}
    \end{minipage}
    
    \caption{Histogram of SONAR mining scores for En $\rightarrow$ XX language pairs in the \dataset~ dataset. The distributions indicate a strong skew toward high-quality alignments (scores above 0.6) across most language pairs, reflecting the effectiveness of the mining and filtering pipeline used in \dataset~ (contd).}
    \label{fig:mining_histogram_ba_2}

\end{figure*}

\begin{figure*}[h!]
    \centering

    \begin{minipage}{\textwidth}
        \centering
        \begin{subfigure}[b]{0.48\textwidth}
            \centering
            \includegraphics[width=\textwidth]{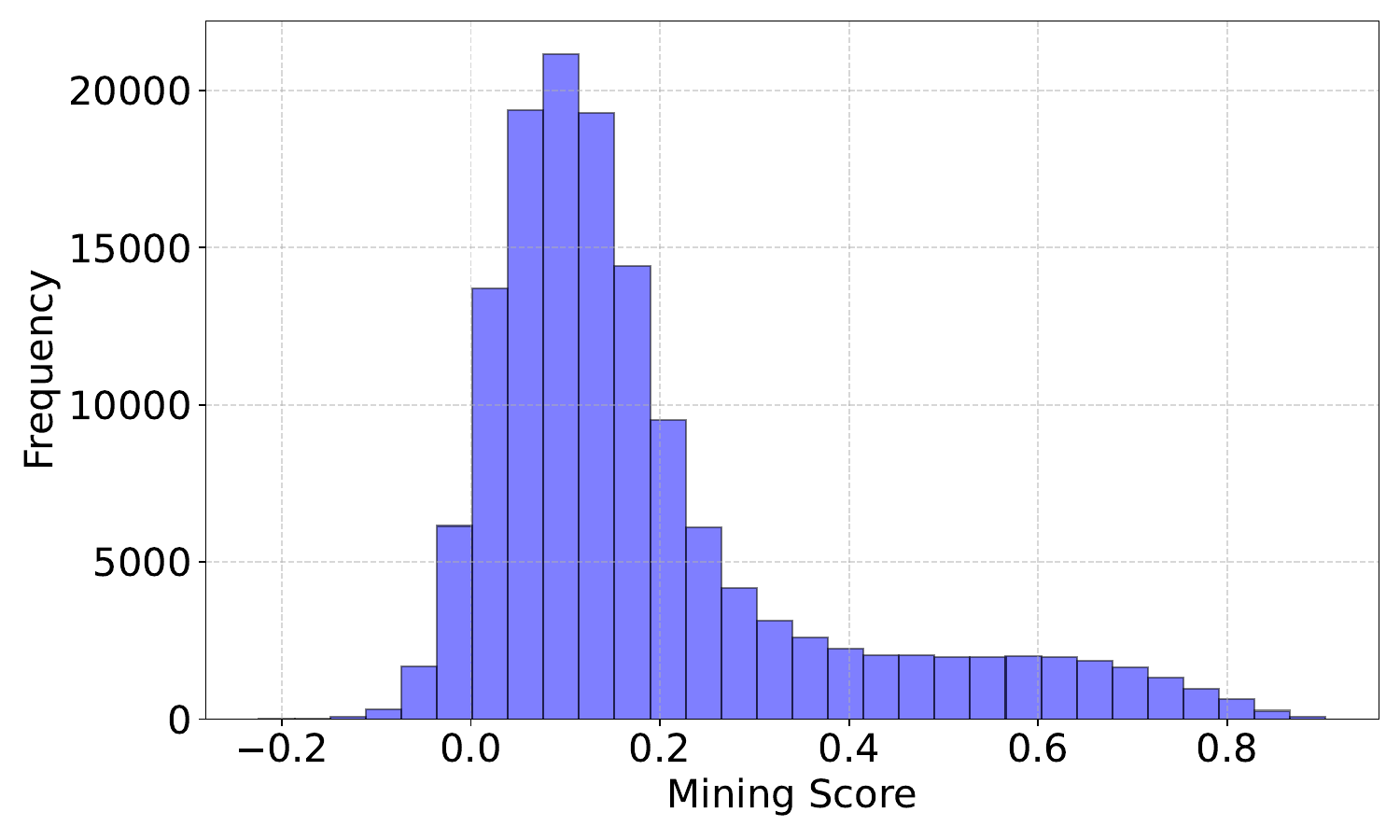}
            \caption{\centering Assamese - English}
            \label{fig:dataset_assamese_i2e}
        \end{subfigure}
        \hfill
        \begin{subfigure}[b]{0.48\textwidth}
            \centering
            \includegraphics[width=\textwidth]{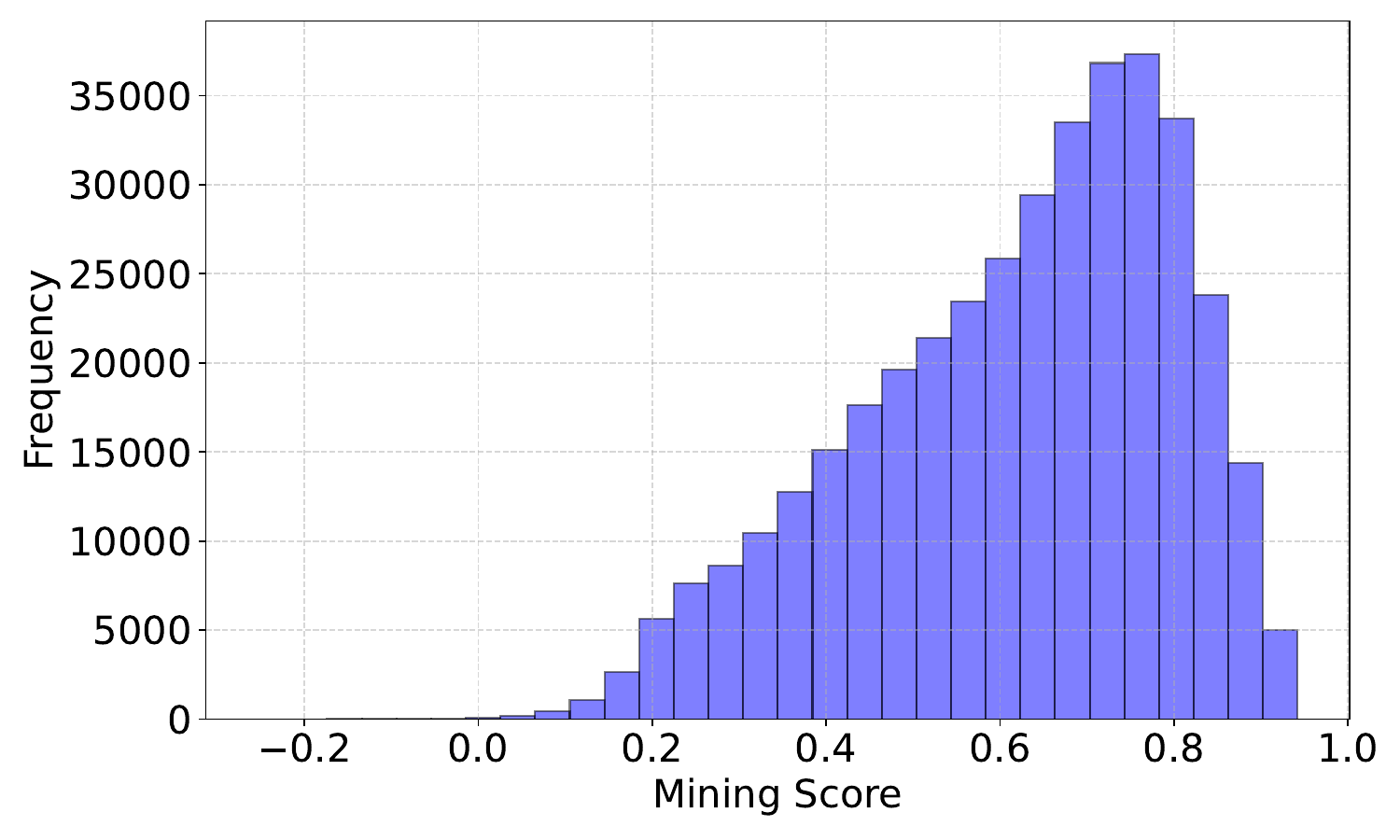}
            \caption{\centering Bengali - English}
            \label{fig:dataset_bengali_i2e}
        \end{subfigure}
    \end{minipage}
    
    \label{fig:plot_analysis_pairs_i2e_as_bn}

    \centering

    \begin{minipage}{\textwidth}
        \centering
        \begin{subfigure}[b]{0.48\textwidth}
            \centering
            \includegraphics[width=\textwidth]{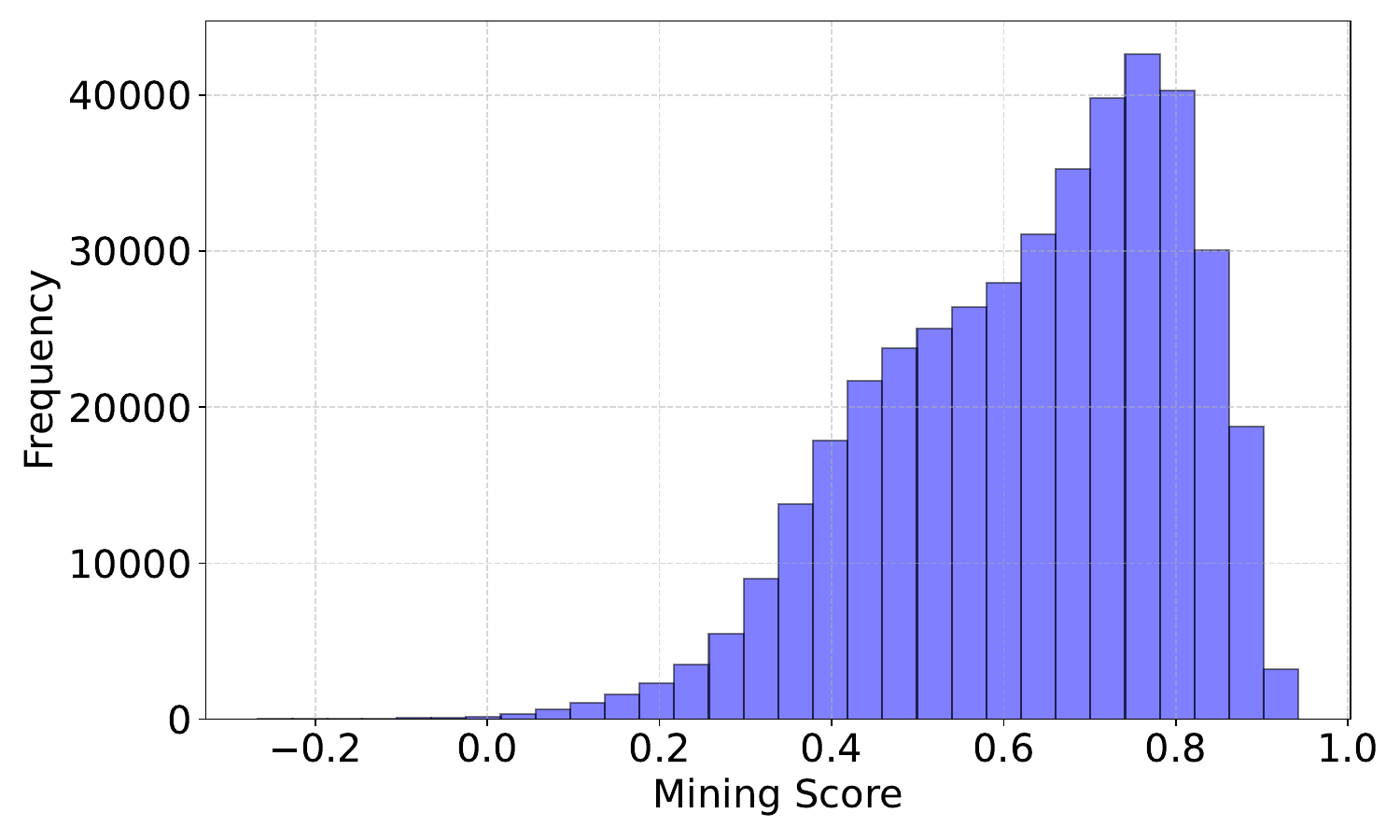}
            \caption{\centering Gujarati - English}
            \label{fig:dataset_gujarati_i2e}
        \end{subfigure}
        \hfill
        \begin{subfigure}[b]{0.48\textwidth}
            \centering
            \includegraphics[width=\textwidth]{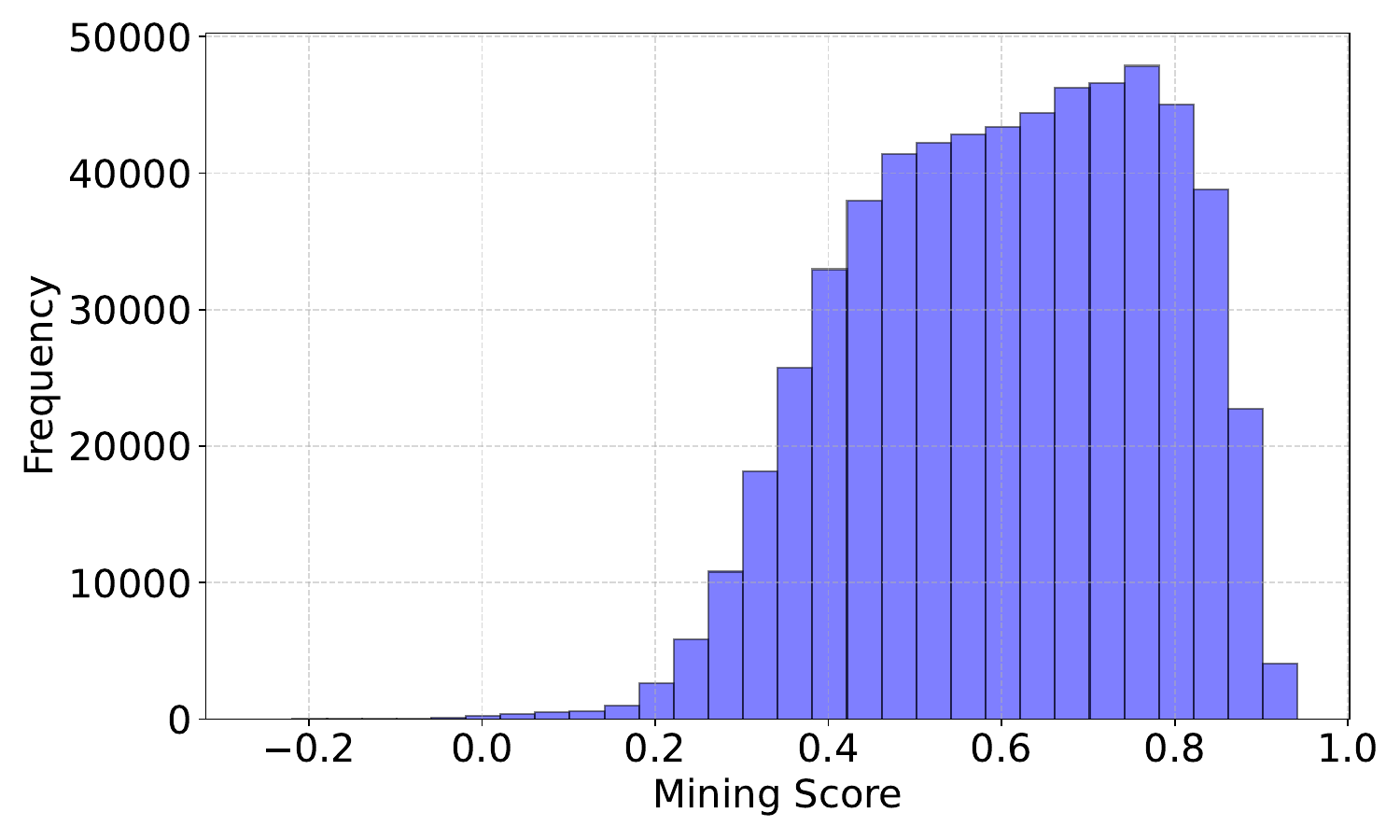}
            \caption{\centering Malayalam - English}
            \label{fig:dataset_malayalam_i2e}
        \end{subfigure}
    \end{minipage}
    
    \label{fig:plot_analysis_pairs_i2e_gu_ml}

    \centering

    \begin{minipage}{\textwidth}
        \centering
        \begin{subfigure}[b]{0.48\textwidth}
            \centering
            \includegraphics[width=\textwidth]{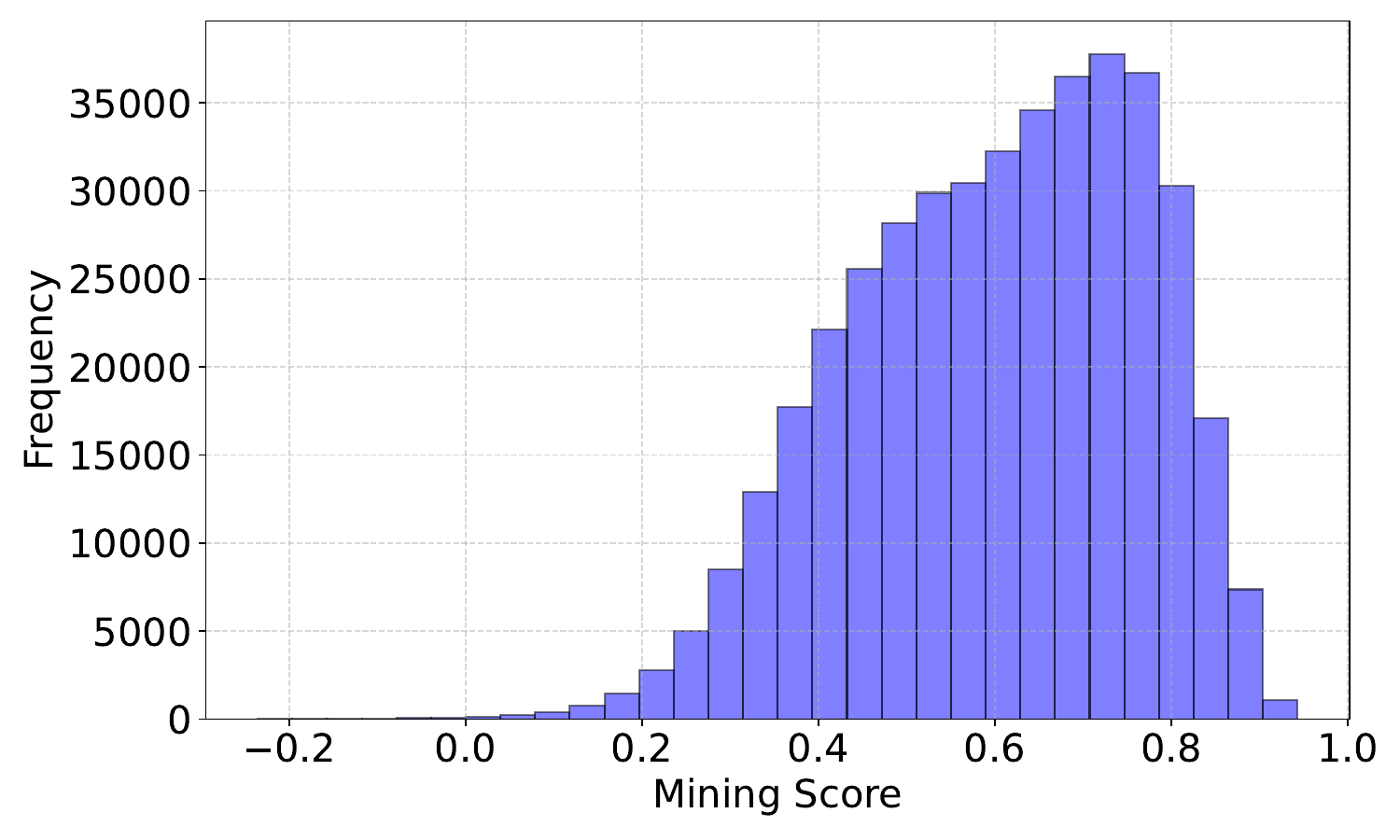}
            \caption{\centering Marathi - English}
            \label{fig:dataset_marathi_i2e}
        \end{subfigure}
        \hfill
        \begin{subfigure}[b]{0.48\textwidth}
            \centering
            \includegraphics[width=\textwidth]{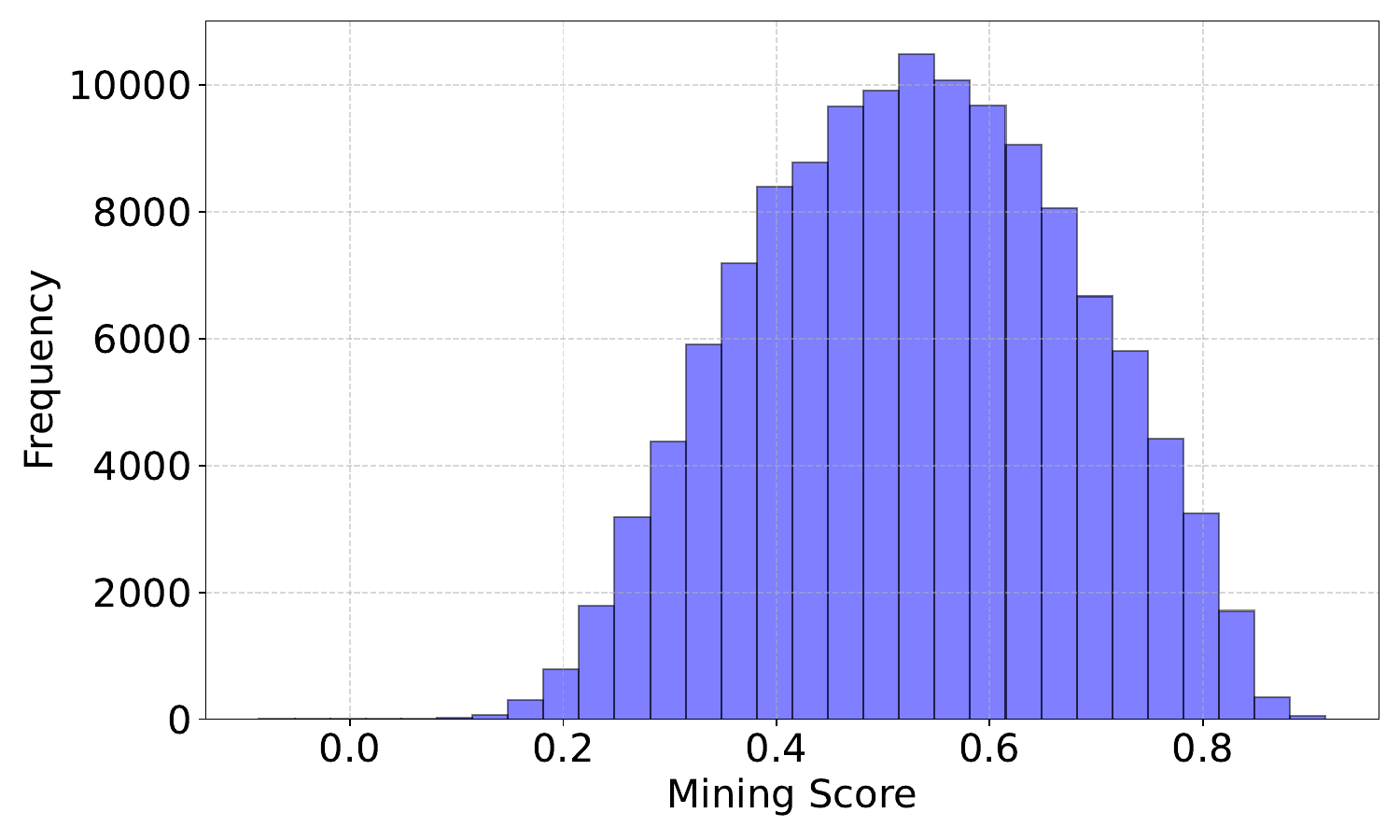}
            \caption{\centering Nepali - English}
            \label{fig:dataset_nepali_i2e}
        \end{subfigure}
    \end{minipage}
    
    \label{fig:plot_analysis_pairs_i2e_mr_ne}

    \centering

    \begin{minipage}{\textwidth}
        \centering
        \begin{subfigure}[b]{0.48\textwidth}
            \centering
            \includegraphics[width=\textwidth]{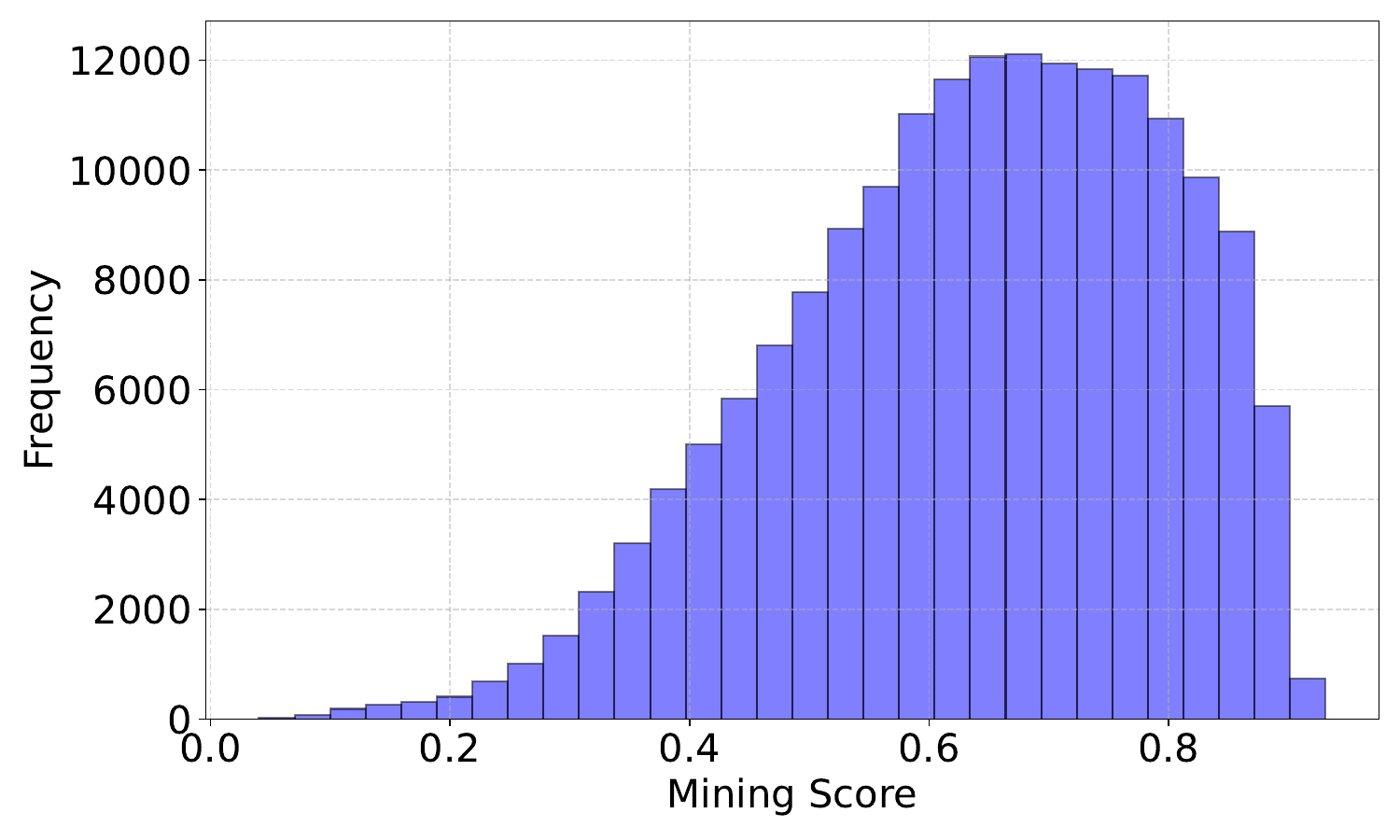}
            \caption{\centering Odia - English}
            \label{fig:dataset_odia_i2e}
        \end{subfigure}
        \hfill
        \begin{subfigure}[b]{0.48\textwidth}
            \centering
            \includegraphics[width=\textwidth]{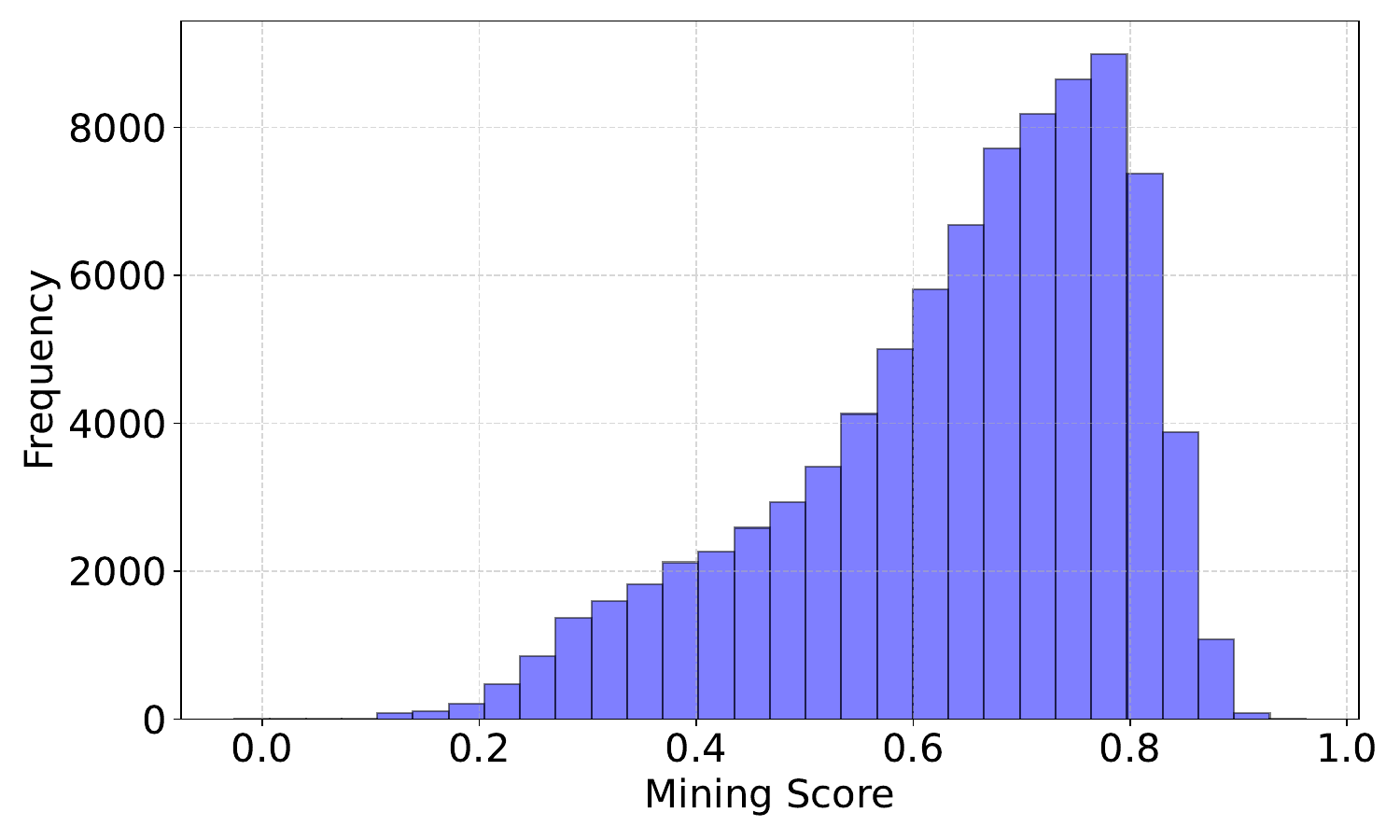}
            \caption{\centering Punjabi - English}
            \label{fig:dataset_punjabi_i2e}
        \end{subfigure}
    \end{minipage}
    
    \label{fig:plot_analysis_pairs_i2e_or_pa}

    \caption{Histogram of SONAR mining scores for XX $\rightarrow$ Any language pairs in the \dataset~ dataset. The distributions indicate a strong skew toward high-quality alignments (scores above 0.6) across most language pairs, reflecting the effectiveness of the mining and filtering pipeline used in \dataset~.}
    \label{fig:mining_histogram_ba_3}
\end{figure*}

A histogram depicting the distribution of mining scores for each language pair in the dataset is illustrated in Figure~\ref{fig:lang_dist_1}, Figure~\ref{fig:lang_dist_2}, Figure~\ref{fig:mining_histogram_ba_1}, Figure~\ref{fig:mining_histogram_ba_2}, and Figure~\ref{fig:mining_histogram_ba_3}. These visualizations highlight a significant challenge: the SONAR model's performance is considerably constrained when applied to \textbf{low-resource languages}. 

For instance, the Assamese-English language pair (Figure~\ref{fig:dataset_assamese_i2e}) exhibits a distribution of mining scores that is heavily skewed towards lower values, with a distinct peak frequency around a mere 0.2. Similarly, the English-Sindhi pair, as depicted in Figure~\ref{fig:dataset_eng_sd}, also demonstrates this limitation, with its most prominent frequency peak for mining scores occurring around 0.2 to 0.3, despite some presence of higher scores. This performance with Sindhi, a low-resource language, strongly suggests that the SONAR model struggles to generate consistently effective cross-lingual embeddings for under-represented languages.

In contrast, the model demonstrates commendable efficacy with higher-resource Indian languages. For language pairs like Bengali-English (Figure~\ref{fig:dataset_bengali_i2e}), Gujarati-English (Figure~\ref{fig:dataset_gujarati_i2e}), Malayalam-English (Figure~\ref{fig:dataset_malayalam_i2e}), and Marathi-English (Figure~\ref{fig:dataset_marathi_i2e}), the mining scores predominantly cluster at higher values, typically peaking between 0.6 and 0.8. This indicates the SONAR model's capability to capture semantic similarities effectively when ample linguistic resources are available.

\textbf{These divergent outcomes underscore a critical need: current multilingual embedding models like SONAR are not robust enough for low-resource languages.} To address this, two major steps are imperative: (1) the adoption of \textbf{language-specific thresholds} when filtering mined data, rather than applying uniform cutoffs across all language pairs, and (2) the development of \textbf{improved multilingual embedding models} that can provide more accurate and reliable cross-lingual representations, particularly for under-represented languages like Assamese and Sindhi. Without such targeted improvements, data mining efforts risk perpetuating the under-representation and degraded performance of these languages in downstream tasks.



\begin{table*}[]
\centering
\fontsize{8pt}{10pt}\selectfont
\begin{tabular}{@{}ccccccccccccc@{}}
\toprule
{\color[HTML]{31333F} }                            & \multicolumn{6}{c}{\textbf{Direct}}                                                                                                                                                                                                   & \multicolumn{6}{c}{\textbf{Cascaded}}                                                                                                                                                   \\ \cmidrule(lr){2-7} \cmidrule(lr){8-13}
\cellcolor[HTML]{FFFFFF}{\color[HTML]{31333F} }                                & \multicolumn{2}{c}{\textbf{SD}}                                                                           & \multicolumn{2}{c}{\textbf{SD\textsubscript{BA}}}                           & \multicolumn{2}{c}{\textbf{AZURE}}                                           & \multicolumn{2}{c}{\textbf{SC}}                               & \multicolumn{2}{c}{\textbf{IC + IT2}}                          & \multicolumn{2}{c}{\textbf{SC + IT2}}                         \\ \cmidrule(lr){2-3} \cmidrule(lr){4-5} \cmidrule(lr){6-7} \cmidrule(lr){8-9} \cmidrule(lr){10-11} \cmidrule(lr){12-13}
\multirow{-3}{*}{\cellcolor[HTML]{FFFFFF}{\color[HTML]{31333F} \textit{lang}}} & \textbf{FL}                                     & \textbf{BA}                                         & \textbf{FL}               & \textbf{BA}                   & \textbf{FL}                              & \textbf{BA}                   & \textbf{FL}               & \textbf{BA}                   & \textbf{FL}               & \textbf{BA}                   & \textbf{FL}               & \textbf{BA}                   \\ \midrule
{\color[HTML]{0D0D0D} \textit{asm}}                & \cellcolor[HTML]{ABDDC5}{\color[HTML]{0D0D0D} 0.77} & \cellcolor[HTML]{FFF0C3}{\color[HTML]{0D0D0D} 0.76} & \cellcolor[HTML]{B0DFC9}0.75 & \cellcolor[HTML]{FFE59C}0.83 & \cellcolor[HTML]{D4EEE2}0.59 & \cellcolor[HTML]{FFF6DA}0.70 & \cellcolor[HTML]{D7EFE3}0.58 & \cellcolor[HTML]{FFFFFF}0.60 & \cellcolor[HTML]{E0F3EA}0.54 & \cellcolor[HTML]{FFFBF0}0.64 & \cellcolor[HTML]{D7EFE3}0.58 & \cellcolor[HTML]{FFEDBB}0.78 \\
{\color[HTML]{0D0D0D} \textit{ben}}                & \cellcolor[HTML]{71C69D}{\color[HTML]{0D0D0D} 0.86} & \cellcolor[HTML]{FFE7A4}{\color[HTML]{0D0D0D} 0.82} & \cellcolor[HTML]{85CEAA}0.83 & \cellcolor[HTML]{FFDB76}0.88 & \cellcolor[HTML]{B2E0CA}0.74 & \cellcolor[HTML]{FFF7E1}0.68 & \cellcolor[HTML]{98D6B8}0.80 & \cellcolor[HTML]{FFEDBB}0.78 & \cellcolor[HTML]{ABDDC5}0.77 & \cellcolor[HTML]{FFF1C7}0.75 & \cellcolor[HTML]{98D6B8}0.80 & \cellcolor[HTML]{FFE7A4}0.82 \\
{\color[HTML]{0D0D0D} \textit{guj}}                & \cellcolor[HTML]{71C69D}{\color[HTML]{0D0D0D} 0.86} & \cellcolor[HTML]{FFE395}{\color[HTML]{0D0D0D} 0.84} & \cellcolor[HTML]{78C9A1}0.85 & \cellcolor[HTML]{FFDB76}0.88 & \cellcolor[HTML]{C4E8D7}0.66 & \cellcolor[HTML]{FFFFFF}0.59 & \cellcolor[HTML]{64C194}0.88 & \cellcolor[HTML]{FFE18D}0.85 & \cellcolor[HTML]{6BC398}0.87 & \cellcolor[HTML]{FFE59C}0.83 & \cellcolor[HTML]{64C194}0.88 & \cellcolor[HTML]{FFDF85}0.86 \\
{\color[HTML]{0D0D0D} \textit{hin}}                & \cellcolor[HTML]{6BC398}{\color[HTML]{0D0D0D} 0.87} & \cellcolor[HTML]{FFE18D}{\color[HTML]{0D0D0D} 0.85} & \cellcolor[HTML]{78C9A1}0.85 & \cellcolor[HTML]{FFDB76}0.88 & \cellcolor[HTML]{98D6B8}0.80 & \cellcolor[HTML]{FFEDBB}0.78 & \cellcolor[HTML]{B5E1CC}0.73 & \cellcolor[HTML]{FFEFBF}0.77 & \cellcolor[HTML]{B9E3CF}0.71 & \cellcolor[HTML]{FFF6DA}0.70 & \cellcolor[HTML]{B5E1CC}0.73 & \cellcolor[HTML]{FFE18D}0.85 \\
{\color[HTML]{0D0D0D} \textit{kan}}                & \cellcolor[HTML]{71C69D}{\color[HTML]{0D0D0D} 0.86} & \cellcolor[HTML]{FFE395}{\color[HTML]{0D0D0D} 0.84} & \cellcolor[HTML]{7ECBA6}0.84 & \cellcolor[HTML]{FFD96E}0.89 & \cellcolor[HTML]{B7E2CD}0.72 & \cellcolor[HTML]{FFF7DD}0.69 & \cellcolor[HTML]{ABDDC5}0.77 & \cellcolor[HTML]{FFEFBF}0.77 & \cellcolor[HTML]{AEDEC7}0.76 & \cellcolor[HTML]{FFEFBF}0.77 & \cellcolor[HTML]{ABDDC5}0.77 & \cellcolor[HTML]{FFE18D}0.85 \\
{\color[HTML]{0D0D0D} \textit{mal}}                & \cellcolor[HTML]{71C69D}{\color[HTML]{0D0D0D} 0.86} & \cellcolor[HTML]{FFEBB3}{\color[HTML]{0D0D0D} 0.80} & \cellcolor[HTML]{7ECBA6}0.84 & \cellcolor[HTML]{FFDD7E}0.87 & \cellcolor[HTML]{B9E3CF}0.71 & \cellcolor[HTML]{FFF9E9}0.66 & \cellcolor[HTML]{71C69D}0.86 & \cellcolor[HTML]{FFEFBF}0.77 & \cellcolor[HTML]{71C69D}0.86 & \cellcolor[HTML]{FFEBB3}0.80 & \cellcolor[HTML]{6BC398}0.87 & \cellcolor[HTML]{FFE9AC}0.81 \\
{\color[HTML]{0D0D0D} \textit{mar}}                & \cellcolor[HTML]{8BD0AF}{\color[HTML]{0D0D0D} 0.82} & \cellcolor[HTML]{FFE7A4}{\color[HTML]{0D0D0D} 0.82} & \cellcolor[HTML]{98D6B8}0.80 & \cellcolor[HTML]{FFDD7E}0.87 & \cellcolor[HTML]{C4E8D7}0.66 & \cellcolor[HTML]{FFF7DD}0.69 & \cellcolor[HTML]{C2E7D5}0.67 & \cellcolor[HTML]{FFF9E9}0.66 & \cellcolor[HTML]{C7E9D8}0.65 & \cellcolor[HTML]{FFF9E9}0.66 & \cellcolor[HTML]{C2E7D5}0.67 & \cellcolor[HTML]{FFE395}0.84 \\
{\color[HTML]{0D0D0D} \textit{npi}}                & \cellcolor[HTML]{85CEAA}{\color[HTML]{0D0D0D} 0.83} & \cellcolor[HTML]{FFFAEC}{\color[HTML]{0D0D0D} 0.65} & \cellcolor[HTML]{92D3B3}0.81 & \cellcolor[HTML]{FFE59C}0.83 & \cellcolor[HTML]{C4E8D7}0.66 & \cellcolor[HTML]{FFFFFF}0.55 & \cellcolor[HTML]{B0DFC9}0.75 & \cellcolor[HTML]{FFFEFC}0.61 & \cellcolor[HTML]{B9E3CF}0.71 & \cellcolor[HTML]{FFF8E5}0.67 & \cellcolor[HTML]{B0DFC9}0.75 & \cellcolor[HTML]{FFF8E5}0.67 \\
{\color[HTML]{0D0D0D} \textit{ory}}                & \cellcolor[HTML]{8BD0AF}{\color[HTML]{0D0D0D} 0.82} & \cellcolor[HTML]{FFE18D}{\color[HTML]{0D0D0D} 0.85} & \cellcolor[HTML]{8BD0AF}0.82 & \cellcolor[HTML]{FFD96E}0.89 & \cellcolor[HTML]{B5E1CC}0.73 & \cellcolor[HTML]{FFF1C7}0.75 & \cellcolor[HTML]{98D6B8}0.80 & \cellcolor[HTML]{FFEDBB}0.78 & \cellcolor[HTML]{92D3B3}0.81 & \cellcolor[HTML]{FFE9AC}0.81 & \cellcolor[HTML]{98D6B8}0.80 & \cellcolor[HTML]{FFDD7E}0.87 \\
{\color[HTML]{0D0D0D} \textit{pan}}                & \cellcolor[HTML]{9FD8BC}{\color[HTML]{0D0D0D} 0.79} & \cellcolor[HTML]{FFF5D6}{\color[HTML]{0D0D0D} 0.71} & \cellcolor[HTML]{98D6B8}0.80 & \cellcolor[HTML]{FFECB7}0.79 & \cellcolor[HTML]{D4EEE2}0.59 & \cellcolor[HTML]{FFF7E1}0.68 & \cellcolor[HTML]{ABDDC5}0.77 & \cellcolor[HTML]{FFF4D2}0.72 & \cellcolor[HTML]{A5DBC1}0.78 & \cellcolor[HTML]{FFF2CA}0.74 & \cellcolor[HTML]{ABDDC5}0.77 & \cellcolor[HTML]{FFF4D2}0.72 \\
{\color[HTML]{0D0D0D} \textit{snd}}                & \cellcolor[HTML]{F9FDFB}{\color[HTML]{0D0D0D} 0.43} & {\color[HTML]{0D0D0D} -}                            & \cellcolor[HTML]{D2EDE0}0.60 & -                            & -                            & -                            & \cellcolor[HTML]{FFFFFF}0.39 & -                            & \cellcolor[HTML]{FFFFFF}0.21 & \multicolumn{1}{l}{}         & \cellcolor[HTML]{FFFFFF}0.39 & \multicolumn{1}{l}{}         \\
{\color[HTML]{0D0D0D} \textit{tam}}                & \cellcolor[HTML]{92D3B3}{\color[HTML]{0D0D0D} 0.81} & \cellcolor[HTML]{FFF2CA}{\color[HTML]{0D0D0D} 0.74} & \cellcolor[HTML]{A5DBC1}0.78 & \cellcolor[HTML]{FFE18D}0.85 & \cellcolor[HTML]{B9E3CF}0.71 & \cellcolor[HTML]{FFF8E5}0.67 & \cellcolor[HTML]{BEE5D2}0.69 & \cellcolor[HTML]{FFF8E5}0.67 & \cellcolor[HTML]{BEE5D2}0.69 & \cellcolor[HTML]{FFF7E1}0.68 & \cellcolor[HTML]{BEE5D2}0.69 & \cellcolor[HTML]{FFEFBF}0.77 \\
{\color[HTML]{0D0D0D} \textit{tel}}                & \cellcolor[HTML]{85CEAA}{\color[HTML]{0D0D0D} 0.83} & \cellcolor[HTML]{FFEFBF}{\color[HTML]{0D0D0D} 0.77} & \cellcolor[HTML]{98D6B8}0.80 & \cellcolor[HTML]{FFDF85}0.86 & \cellcolor[HTML]{BBE4D0}0.70 & \cellcolor[HTML]{FFFAEC}0.65 & \cellcolor[HTML]{B5E1CC}0.73 & \cellcolor[HTML]{FFF5D6}0.71 & \cellcolor[HTML]{B5E1CC}0.73 & \cellcolor[HTML]{FFF5D6}0.71 & \cellcolor[HTML]{B5E1CC}0.73 & \cellcolor[HTML]{FFE9AC}0.81 \\
{\color[HTML]{0D0D0D} \textit{urd}}                & \cellcolor[HTML]{8BD0AF}{\color[HTML]{0D0D0D} 0.82} & \cellcolor[HTML]{FFE7A4}{\color[HTML]{0D0D0D} 0.82} & \cellcolor[HTML]{8BD0AF}0.82 & \cellcolor[HTML]{FFDF85}0.86 & \cellcolor[HTML]{BBE4D0}0.70 & \cellcolor[HTML]{FFF8E5}0.67 & \cellcolor[HTML]{B5E1CC}0.73 & \cellcolor[HTML]{FFEFBF}0.77 & \cellcolor[HTML]{C2E7D5}0.67 & \cellcolor[HTML]{FFF6DA}0.70 & \cellcolor[HTML]{B5E1CC}0.73 & \cellcolor[HTML]{FFE59C}0.83 \\ \midrule
\textit{total}                                     & 0.80                                                & 0.79                                                & 0.80                         & 0.86                         & 0.69                         & 0.67                         & 0.73                         & 0.73                         & 0.70                         & 0.73                         & 0.73                         & 0.81                         \\ \bottomrule
\end{tabular}
\caption{COMET ($\uparrow$) scores for different direct and cascaded speech translation models evaluated on the \textsc{Fleurs-Test} (FL; \textcolor{customgreen}{green}) and \textsc{BhasaAnuvaad-Test} (BA; \textcolor{customyellow}{yellow}) datasets in $XX \rightarrow En$ direction. The colors represent a heatmap, where darker shades indicating better translation performance.}
\label{tab:comet_xx_en}

\bigskip

\begin{tabular}{@{}ccccccccccccc@{}}
\toprule
{\color[HTML]{31333F} }                            & \multicolumn{6}{c}{\textbf{Direct}}                                                                                                                                                                                                                                                                                               & \multicolumn{6}{c}{\textbf{Cascaded}}                                                                                                                                                                                                                                              \\ \cmidrule(lr){2-7} \cmidrule(lr){8-13}
\cellcolor[HTML]{FFFFFF}{\color[HTML]{31333F} }                                & \multicolumn{2}{c}{\textbf{SD}}                                                                           & \multicolumn{2}{c}{\textbf{SD\textsubscript{BA}}}                           & \multicolumn{2}{c}{\textbf{AZURE}}                                           & \multicolumn{2}{c}{\textbf{SC}}                               & \multicolumn{2}{c}{\textbf{W + IT2}}                          & \multicolumn{2}{c}{\textbf{SC + IT2}}                         \\ \cmidrule(lr){2-3} \cmidrule(lr){4-5} \cmidrule(lr){6-7} \cmidrule(lr){8-9} \cmidrule(lr){10-11} \cmidrule(lr){12-13}
\multirow{-3}{*}{\cellcolor[HTML]{FFFFFF}{\color[HTML]{31333F} \textit{lang}}} & \textbf{FL}                                     & \textbf{BA}                                         & \textbf{FL}               & \textbf{BA}                   & \textbf{FL}                              & \textbf{BA}                   & \textbf{FL}               & \textbf{BA}                   & \textbf{FL}               & \textbf{BA}                   & \textbf{FL}               & \textbf{BA}                   \\ \midrule
{\color[HTML]{0D0D0D} \textit{asm}}                & \cellcolor[HTML]{CAEADB}{\color[HTML]{0D0D0D} 0.59} & \cellcolor[HTML]{FFF6DC}{\color[HTML]{0D0D0D} 0.49} & \cellcolor[HTML]{D3EDE1}0.56 & \cellcolor[HTML]{FFEDBB}0.62 & \cellcolor[HTML]{E6F5EE}0.49 & \cellcolor[HTML]{FFF5D9}0.50 & \cellcolor[HTML]{FAFDFC}0.42 & \cellcolor[HTML]{FFFFFF}0.31 & \cellcolor[HTML]{FAFDFC}0.42 & \cellcolor[HTML]{FFFFFF}0.35 & \cellcolor[HTML]{FAFDFC}0.42 & \cellcolor[HTML]{FFF1CA}0.56 \\
{\color[HTML]{0D0D0D} \textit{ben}}                & \cellcolor[HTML]{8ED2B1}{\color[HTML]{0D0D0D} 0.77} & \cellcolor[HTML]{FFF0C3}{\color[HTML]{0D0D0D} 0.59} & \cellcolor[HTML]{9BD7BA}0.74 & \cellcolor[HTML]{FFE59B}0.73 & \cellcolor[HTML]{AEDFC7}0.69 & \cellcolor[HTML]{FFEEBE}0.61 & \cellcolor[HTML]{8ED2B1}0.77 & \cellcolor[HTML]{FFF8E4}0.46 & \cellcolor[HTML]{8AD0AE}0.78 & \cellcolor[HTML]{FFF5D9}0.50 & \cellcolor[HTML]{8ED2B1}0.77 & \cellcolor[HTML]{FFEAAD}0.67 \\
{\color[HTML]{0D0D0D} \textit{guj}}                & \cellcolor[HTML]{68C296}{\color[HTML]{0D0D0D} 0.86} & \cellcolor[HTML]{FFEBB3}{\color[HTML]{0D0D0D} 0.65} & \cellcolor[HTML]{68C296}0.86 & \cellcolor[HTML]{FFE088}0.79 & \cellcolor[HTML]{81CCA8}0.80 & \cellcolor[HTML]{FFEBB3}0.65 & \cellcolor[HTML]{86CEAB}0.79 & \cellcolor[HTML]{FFF8E4}0.46 & \cellcolor[HTML]{81CCA8}0.80 & \cellcolor[HTML]{FFF4D4}0.52 & \cellcolor[HTML]{86CEAB}0.79 & \cellcolor[HTML]{FFE6A1}0.71 \\
{\color[HTML]{0D0D0D} \textit{hin}}                & \cellcolor[HTML]{9BD7BA}{\color[HTML]{0D0D0D} 0.74} & \cellcolor[HTML]{FFEAAD}{\color[HTML]{0D0D0D} 0.67} & \cellcolor[HTML]{9BD7BA}0.74 & \cellcolor[HTML]{FFE395}0.75 & \cellcolor[HTML]{BCE4D1}0.64 & \cellcolor[HTML]{FFEBB3}0.65 & \cellcolor[HTML]{68C296}0.86 & \cellcolor[HTML]{FFECB6}0.64 & \cellcolor[HTML]{6CC499}0.85 & \cellcolor[HTML]{FFE9AA}0.68 & \cellcolor[HTML]{68C296}0.86 & \cellcolor[HTML]{FFE6A1}0.71 \\
{\color[HTML]{0D0D0D} \textit{kan}}                & \cellcolor[HTML]{A3DAC0}{\color[HTML]{0D0D0D} 0.72} & \cellcolor[HTML]{FFF1C8}{\color[HTML]{0D0D0D} 0.57} & \cellcolor[HTML]{A3DAC0}0.72 & \cellcolor[HTML]{FFE69E}0.72 & \cellcolor[HTML]{B7E2CD}0.66 & \cellcolor[HTML]{FFEFC0}0.60 & \cellcolor[HTML]{ABDDC5}0.70 & \cellcolor[HTML]{FFF9E6}0.45 & \cellcolor[HTML]{AEDFC7}0.69 & \cellcolor[HTML]{FFF5D9}0.50 & \cellcolor[HTML]{ABDDC5}0.70 & \cellcolor[HTML]{FFECB6}0.64 \\
{\color[HTML]{0D0D0D} \textit{mal}}                & \cellcolor[HTML]{6CC499}{\color[HTML]{0D0D0D} 0.85} & \cellcolor[HTML]{FFE18F}{\color[HTML]{0D0D0D} 0.77} & \cellcolor[HTML]{71C69C}0.84 & \cellcolor[HTML]{FFDC79}0.84 & \cellcolor[HTML]{86CEAB}0.79 & \cellcolor[HTML]{FFE292}0.76 & \cellcolor[HTML]{79C9A2}0.82 & \cellcolor[HTML]{FFEDB9}0.63 & \cellcolor[HTML]{79C9A2}0.82 & \cellcolor[HTML]{FFECB6}0.64 & \cellcolor[HTML]{79C9A2}0.82 & \cellcolor[HTML]{FFDE82}0.81 \\
{\color[HTML]{0D0D0D} \textit{mar}}                & \cellcolor[HTML]{B7E2CD}{\color[HTML]{0D0D0D} 0.66} & \cellcolor[HTML]{FFF7DF}{\color[HTML]{0D0D0D} 0.48} & \cellcolor[HTML]{BCE4D1}0.64 & \cellcolor[HTML]{FFEDBB}0.62 & \cellcolor[HTML]{C7E9D9}0.60 & \cellcolor[HTML]{FFF5D7}0.51 & \cellcolor[HTML]{A3DAC0}0.72 & \cellcolor[HTML]{FFFBEE}0.42 & \cellcolor[HTML]{A7DCC3}0.71 & \cellcolor[HTML]{FFF8E4}0.46 & \cellcolor[HTML]{A3DAC0}0.72 & \cellcolor[HTML]{FFF0C5}0.58 \\
{\color[HTML]{0D0D0D} \textit{npi}}                & \cellcolor[HTML]{75C79F}{\color[HTML]{0D0D0D} 0.83} & \cellcolor[HTML]{FFF7DF}{\color[HTML]{0D0D0D} 0.48} & \cellcolor[HTML]{71C69C}0.84 & \cellcolor[HTML]{FFE6A1}0.71 & \cellcolor[HTML]{9FD8BD}0.73 & \cellcolor[HTML]{FFF5D7}0.51 & \cellcolor[HTML]{86CEAB}0.79 & \cellcolor[HTML]{FFFFFF}0.32 & \cellcolor[HTML]{81CCA8}0.8  & \cellcolor[HTML]{FFFDF8}0.38 & \cellcolor[HTML]{86CEAB}0.79 & \cellcolor[HTML]{FFF0C3}0.59 \\
{\color[HTML]{0D0D0D} \textit{ory}}                & \cellcolor[HTML]{75C79F}{\color[HTML]{0D0D0D} 0.83} & \cellcolor[HTML]{FFF0C5}{\color[HTML]{0D0D0D} 0.58} & \cellcolor[HTML]{81CCA8}0.80 & \cellcolor[HTML]{FFE59B}0.73 & \cellcolor[HTML]{9BD7BA}0.74 & \cellcolor[HTML]{FFF1CA}0.56 & \cellcolor[HTML]{B7E2CD}0.66 & \cellcolor[HTML]{FFFFFF}0.33 & \cellcolor[HTML]{B4E1CB}0.67 & \cellcolor[HTML]{FFFEFA}0.37 & \cellcolor[HTML]{B7E2CD}0.66 & \cellcolor[HTML]{FFEBB0}0.66 \\
{\color[HTML]{0D0D0D} \textit{pan}}                & \cellcolor[HTML]{79C9A2}{\color[HTML]{0D0D0D} 0.82} & \cellcolor[HTML]{FFF6DC}{\color[HTML]{0D0D0D} 0.49} & \cellcolor[HTML]{79C9A2}0.82 & \cellcolor[HTML]{FFF1C8}0.57 & \cellcolor[HTML]{92D3B4}0.76 & \cellcolor[HTML]{FFECB6}0.64 & \cellcolor[HTML]{9BD7BA}0.74 & \cellcolor[HTML]{FFFFFF}0.29 & \cellcolor[HTML]{9BD7BA}0.74 & \cellcolor[HTML]{FFFDF5}0.39 & \cellcolor[HTML]{9BD7BA}0.74 & \cellcolor[HTML]{FFF5D9}0.50 \\
{\color[HTML]{0D0D0D} \textit{snd}}                & \cellcolor[HTML]{AEDFC7}{\color[HTML]{0D0D0D} 0.69} & {\color[HTML]{0D0D0D} -}                            & \cellcolor[HTML]{A3DAC0}0.72 & -                            & -                            & -                            & \cellcolor[HTML]{B9E3CF}0.65 & -                            & \cellcolor[HTML]{B7E2CD}0.66 & \cellcolor[HTML]{FFFFFF}0.32 & \cellcolor[HTML]{B9E3CF}0.65 & --                           \\
{\color[HTML]{0D0D0D} \textit{tam}}                & \cellcolor[HTML]{B1E0C9}{\color[HTML]{0D0D0D} 0.68} & \cellcolor[HTML]{FFF2CD}{\color[HTML]{0D0D0D} 0.55} & \cellcolor[HTML]{B4E1CB}0.67 & \cellcolor[HTML]{FFEAAD}0.67 & \cellcolor[HTML]{C2E7D5}0.62 & \cellcolor[HTML]{FFF0C3}0.59 & \cellcolor[HTML]{A7DCC3}0.71 & \cellcolor[HTML]{FFF8E4}0.46 & \cellcolor[HTML]{A3DAC0}0.72 & \cellcolor[HTML]{FFF5D9}0.50 & \cellcolor[HTML]{A7DCC3}0.71 & \cellcolor[HTML]{FFEDB9}0.63 \\
{\color[HTML]{0D0D0D} \textit{tel}}                & \cellcolor[HTML]{96D5B7}{\color[HTML]{0D0D0D} 0.75} & \cellcolor[HTML]{FFF1CA}{\color[HTML]{0D0D0D} 0.56} & \cellcolor[HTML]{9BD7BA}0.74 & \cellcolor[HTML]{FFE6A1}0.71 & \cellcolor[HTML]{A7DCC3}0.71 & \cellcolor[HTML]{FFEDBB}0.62 & \cellcolor[HTML]{9FD8BD}0.73 & \cellcolor[HTML]{FFF9E6}0.45 & \cellcolor[HTML]{9BD7BA}0.74 & \cellcolor[HTML]{FFF6DC}0.49 & \cellcolor[HTML]{9FD8BD}0.73 & \cellcolor[HTML]{FFEDBB}0.62 \\
{\color[HTML]{0D0D0D} \textit{urd}}                & \cellcolor[HTML]{9FD8BD}{\color[HTML]{0D0D0D} 0.73} & \cellcolor[HTML]{FFEFC0}{\color[HTML]{0D0D0D} 0.60} & \cellcolor[HTML]{9FD8BD}0.73 & \cellcolor[HTML]{FFE69E}0.72 & \cellcolor[HTML]{B7E2CD}0.66 & \cellcolor[HTML]{FFF0C3}0.59 & \cellcolor[HTML]{8AD0AE}0.78 & \cellcolor[HTML]{FFF6DC}0.49 & \cellcolor[HTML]{8AD0AE}0.78 & \cellcolor[HTML]{FFF3CF}0.54 & \cellcolor[HTML]{8AD0AE}0.78 & \cellcolor[HTML]{FFEBB3}0.65 \\ \midrule
\textit{total}                                     & 0.75                                                & 0.58                                                & 0.74                         & 0.71                         & 0.68                         & 0.60                         & 0.72                         & 0.44                         & 0.73                         & 0.47                         & 0.72                         & 0.64                        \\\bottomrule
\end{tabular}
\caption{COMET ($\uparrow$) scores for different direct and cascaded speech translation models evaluated on the \textsc{Fleurs-Test} (FL; \textcolor{customgreen}{green}) and \textsc{BhasaAnuvaad-Test} (BA; \textcolor{customyellow}{yellow}) datasets in $En \rightarrow XX$ direction. The colors represent a heatmap, where darker shades indicating better translation performance.}
\label{tab:comet_en_xx}
\end{table*}

\begin{table*}[]
\centering
\fontsize{8pt}{10pt}\selectfont

\end{table*}

\begin{table*}[]
\centering
\fontsize{8pt}{10pt}\selectfont
\begin{tabular}{ccccccccccccc}
\toprule
{\color[HTML]{31333F} }                            & \multicolumn{6}{c}{\textbf{Direct}}                                                                                                                                                                                                                                                                                               & \multicolumn{6}{c}{\textbf{Cascaded}}                                                                                                                                                                                                                                              \\ \cmidrule(lr){2-7} \cmidrule(lr){8-13}
\cellcolor[HTML]{FFFFFF}{\color[HTML]{31333F} }                                & \multicolumn{2}{c}{\textbf{SD}}                                                                           & \multicolumn{2}{c}{\textbf{SD\textsubscript{BA}}}                           & \multicolumn{2}{c}{\textbf{AZURE}}                                           & \multicolumn{2}{c}{\textbf{SC}}                               & \multicolumn{2}{c}{\textbf{W + IT2}}                          & \multicolumn{2}{c}{\textbf{SC + IT2}}                         \\ \cmidrule(lr){2-3} \cmidrule(lr){4-5} \cmidrule(lr){6-7} \cmidrule(lr){8-9} \cmidrule(lr){10-11} \cmidrule(lr){12-13}
\multirow{-3}{*}{\cellcolor[HTML]{FFFFFF}{\color[HTML]{31333F} \textit{lang}}} & \textbf{FL}                                     & \textbf{BA}                                         & \textbf{FL}               & \textbf{BA}                   & \textbf{FL}                              & \textbf{BA}                   & \textbf{FL}               & \textbf{BA}                   & \textbf{FL}               & \textbf{BA}                   & \textbf{FL}               & \textbf{BA}                   \\ \midrule
{\color[HTML]{0D0D0D} \textit{asm}}                & \cellcolor[HTML]{FCFEFD}{\color[HTML]{0D0D0D} 8.20}  & \cellcolor[HTML]{FFFAED}{\color[HTML]{0D0D0D} 8.70}  & \cellcolor[HTML]{F3FAF7}9.50  & \cellcolor[HTML]{FFEFC1}15.50 & \cellcolor[HTML]{FCFEFD}8.20  & \cellcolor[HTML]{FFF4D5}12.40 & \cellcolor[HTML]{FFFFFF}7.70  & \cellcolor[HTML]{FFFAEA}9.10  & \cellcolor[HTML]{F4FBF7}9.40  & \cellcolor[HTML]{FFF0C7}14.60 & \cellcolor[HTML]{F5FBF8}9.20  & \cellcolor[HTML]{FFFAEA}9.10  \\
{\color[HTML]{0D0D0D} \textit{ben}}                & \cellcolor[HTML]{CEEBDD}{\color[HTML]{0D0D0D} 14.80} & \cellcolor[HTML]{FFFBEF}{\color[HTML]{0D0D0D} 8.30}  & \cellcolor[HTML]{BDE4D1}17.20 & \cellcolor[HTML]{FFEDBC}16.30 & \cellcolor[HTML]{C7E8D8}15.80 & \cellcolor[HTML]{FFF5D8}11.90 & \cellcolor[HTML]{D2EDE0}14.20 & \cellcolor[HTML]{FFF7DF}10.90 & \cellcolor[HTML]{B7E2CD}18.00 & \cellcolor[HTML]{FFF0C5}14.80 & \cellcolor[HTML]{BAE3CF}17.60 & \cellcolor[HTML]{FFF7DF}10.90 \\
{\color[HTML]{0D0D0D} \textit{guj}}                & \cellcolor[HTML]{C0E6D3}{\color[HTML]{0D0D0D} 16.80} & \cellcolor[HTML]{FFF5D9}{\color[HTML]{0D0D0D} 11.80} & \cellcolor[HTML]{ADDEC6}19.50 & \cellcolor[HTML]{FFE59D}21.00 & \cellcolor[HTML]{BFE5D2}16.90 & \cellcolor[HTML]{FFF0C6}14.70 & \cellcolor[HTML]{C2E7D5}16.40 & \cellcolor[HTML]{FFF0C6}14.70 & \cellcolor[HTML]{B1E0C9}18.90 & \cellcolor[HTML]{FFE7A4}20.00 & \cellcolor[HTML]{B4E1CB}18.50 & \cellcolor[HTML]{FFF0C6}14.70 \\
{\color[HTML]{0D0D0D} \textit{hin}}                & \cellcolor[HTML]{6CC499}{\color[HTML]{0D0D0D} 28.70} & \cellcolor[HTML]{FFF6DB}{\color[HTML]{0D0D0D} 11.50} & \cellcolor[HTML]{57BB8A}31.60 & \cellcolor[HTML]{FFD666}29.50 & \cellcolor[HTML]{67C295}29.40 & \cellcolor[HTML]{FFDE82}25.20 & \cellcolor[HTML]{6CC499}28.70 & \cellcolor[HTML]{FFE18F}23.30 & \cellcolor[HTML]{61BF91}30.30 & \cellcolor[HTML]{FFDA73}27.50 & \cellcolor[HTML]{61BF91}30.30 & \cellcolor[HTML]{FFE18F}23.30 \\
{\color[HTML]{0D0D0D} \textit{kan}}                & \cellcolor[HTML]{E2F4EB}{\color[HTML]{0D0D0D} 11.90} & \cellcolor[HTML]{FFF9E7}{\color[HTML]{0D0D0D} 9.60}  & \cellcolor[HTML]{CEEBDD}14.80 & \cellcolor[HTML]{FFEAB0}18.10 & \cellcolor[HTML]{D4EEE1}13.90 & \cellcolor[HTML]{FFF4D4}12.50 & \cellcolor[HTML]{E5F5ED}11.40 & \cellcolor[HTML]{FFF5D8}11.90 & \cellcolor[HTML]{CCEBDC}15.00 & \cellcolor[HTML]{FFEDBA}16.50 & \cellcolor[HTML]{D0ECDE}14.50 & \cellcolor[HTML]{FFF5D8}11.90 \\
{\color[HTML]{0D0D0D} \textit{mal}}                & \cellcolor[HTML]{F3FAF7}{\color[HTML]{0D0D0D} 9.50}  & \cellcolor[HTML]{FFFBEF}{\color[HTML]{0D0D0D} 8.40}  & \cellcolor[HTML]{DAF0E6}13.00 & \cellcolor[HTML]{FFEBB2}17.80 & \cellcolor[HTML]{DEF2E8}12.50 & \cellcolor[HTML]{FFF2CE}13.50 & \cellcolor[HTML]{F2FAF6}9.60  & \cellcolor[HTML]{FFF6DC}11.30 & \cellcolor[HTML]{D2EDE0}14.20 & \cellcolor[HTML]{FFEEBD}16.10 & \cellcolor[HTML]{DAF0E6}13.00 & \cellcolor[HTML]{FFF6DC}11.30 \\
{\color[HTML]{0D0D0D} \textit{mar}}                & \cellcolor[HTML]{E8F6EF}{\color[HTML]{0D0D0D} 11.10} & \cellcolor[HTML]{FFFCF4}{\color[HTML]{0D0D0D} 7.60}  & \cellcolor[HTML]{DFF2E9}12.30 & \cellcolor[HTML]{FFEFC1}15.50 & \cellcolor[HTML]{EAF7F1}10.70 & \cellcolor[HTML]{FFF7DF}10.80 & \cellcolor[HTML]{EDF8F2}10.40 & \cellcolor[HTML]{FFF7DF}10.90 & \cellcolor[HTML]{D8F0E4}13.30 & \cellcolor[HTML]{FFF0C7}14.60 & \cellcolor[HTML]{DCF1E7}12.80 & \cellcolor[HTML]{FFF7DF}10.90 \\
{\color[HTML]{0D0D0D} \textit{npi}}                & \cellcolor[HTML]{D3EDE0}{\color[HTML]{0D0D0D} 14.10} & \cellcolor[HTML]{FFFEF8}{\color[HTML]{0D0D0D} 6.90}  & \cellcolor[HTML]{C5E8D7}16.00 & \cellcolor[HTML]{FFF0C4}15.00 & \cellcolor[HTML]{DCF1E7}12.70 & \cellcolor[HTML]{FFFFFF}5.80  & \cellcolor[HTML]{DAF0E6}13.00 & \cellcolor[HTML]{FFF9E6}9.70  & \cellcolor[HTML]{BCE4D1}17.3  & \cellcolor[HTML]{FFF3CF}13.30 & \cellcolor[HTML]{BFE5D2}16.90 & \cellcolor[HTML]{FFF9E6}9.70  \\
{\color[HTML]{0D0D0D} \textit{ory}}                & \cellcolor[HTML]{D8F0E4}{\color[HTML]{0D0D0D} 13.30} & \cellcolor[HTML]{FFFDF5}{\color[HTML]{0D0D0D} 7.50}  & \cellcolor[HTML]{CAEADA}15.30 & \cellcolor[HTML]{FFF4D4}12.50 & \cellcolor[HTML]{D1EDDF}14.30 & \cellcolor[HTML]{FFFFFE}6.10  & \cellcolor[HTML]{E5F5ED}11.40 & \cellcolor[HTML]{FFFAEC}8.80  & \cellcolor[HTML]{E0F3E9}12.20 & \cellcolor[HTML]{FFF9E9}9.30  & \cellcolor[HTML]{E2F4EB}11.90 & \cellcolor[HTML]{FFFAEC}8.80  \\
{\color[HTML]{0D0D0D} \textit{pan}}                & \cellcolor[HTML]{8DD1B0}{\color[HTML]{0D0D0D} 24.00} & \cellcolor[HTML]{FFF5DA}{\color[HTML]{0D0D0D} 11.60} & \cellcolor[HTML]{7ECBA5}26.10 & \cellcolor[HTML]{FFECB5}17.30 & \cellcolor[HTML]{95D4B5}22.90 & \cellcolor[HTML]{FFF8E3}10.20 & \cellcolor[HTML]{95D5B6}22.80 & \cellcolor[HTML]{FFF4D4}12.60 & \cellcolor[HTML]{81CCA7}25.70 & \cellcolor[HTML]{FFEEBC}16.20 & \cellcolor[HTML]{84CEAA}25.20 & \cellcolor[HTML]{FFF4D4}12.60 \\
{\color[HTML]{0D0D0D} \textit{snd}}                & \cellcolor[HTML]{A6DBC1}{\color[HTML]{0D0D0D} 20.50} & {\color[HTML]{0D0D0D} ---}                          & \cellcolor[HTML]{90D2B2}23.60 & {\color[HTML]{0D0D0D} ---}   & {\color[HTML]{0D0D0D} ---}   & {\color[HTML]{0D0D0D} ---}   & \cellcolor[HTML]{AADDC4}19.80 & {\color[HTML]{0D0D0D} ---}   & \cellcolor[HTML]{C0E6D3}16.80 & {\color[HTML]{0D0D0D} ---}   & \cellcolor[HTML]{C4E7D6}16.20 & {\color[HTML]{0D0D0D} ---}   \\
{\color[HTML]{0D0D0D} \textit{tam}}                & \cellcolor[HTML]{B5E1CC}{\color[HTML]{0D0D0D} 18.30} & \cellcolor[HTML]{FFFFFC}{\color[HTML]{0D0D0D} 6.30}  & \cellcolor[HTML]{CDEBDC}14.90 & \cellcolor[HTML]{FFEDBA}16.60 & \cellcolor[HTML]{DBF1E6}12.90 & \cellcolor[HTML]{FFF2CC}13.80 & \cellcolor[HTML]{E8F6EF}11.00 & \cellcolor[HTML]{FFF9E6}9.80  & \cellcolor[HTML]{D6EFE3}13.60 & \cellcolor[HTML]{FFEDBA}16.60 & \cellcolor[HTML]{D8F0E4}13.30 & \cellcolor[HTML]{FFF9E6}9.80  \\
{\color[HTML]{0D0D0D} \textit{tel}}                & \cellcolor[HTML]{C7E8D8}{\color[HTML]{0D0D0D} 15.80} & \cellcolor[HTML]{FFFCF4}{\color[HTML]{0D0D0D} 7.60}  & \cellcolor[HTML]{B2E0CA}18.70 & \cellcolor[HTML]{FFECB7}17.00 & \cellcolor[HTML]{CBEADB}15.20 & \cellcolor[HTML]{FFF7E1}10.60 & \cellcolor[HTML]{CAEADA}15.30 & \cellcolor[HTML]{FFF9E6}9.80  & \cellcolor[HTML]{B1E0C9}18.80 & \cellcolor[HTML]{FFEFC2}15.40 & \cellcolor[HTML]{B8E2CE}17.90 & \cellcolor[HTML]{FFF9E6}9.80  \\
{\color[HTML]{0D0D0D} \textit{urd}}                & \cellcolor[HTML]{9CD7BA}{\color[HTML]{0D0D0D} 21.80} & \cellcolor[HTML]{FFE7A6}{\color[HTML]{0D0D0D} 19.70} & \cellcolor[HTML]{94D4B5}23.00 & \cellcolor[HTML]{FFDE84}24.90 & \cellcolor[HTML]{9AD7B9}22.10 & \cellcolor[HTML]{FFE6A1}20.50 & \cellcolor[HTML]{A1D9BD}21.20 & \cellcolor[HTML]{FFE394}22.50 & \cellcolor[HTML]{90D2B2}23.60 & \cellcolor[HTML]{FFDA75}27.30 & \cellcolor[HTML]{8ED1B0}23.90 & \cellcolor[HTML]{FFE394}22.50 \\ \midrule
\textit{total}                                     & 16.34                                                & 9.65                                                 & 18.25                         & 18.23                         & 15.96                         & 12.92                         & 15.21                         & 12.72                         & 17.65                         & 17.09                         & 17.23                         & 12.72                        \\\bottomrule
\end{tabular}
\caption{BLEU ($\uparrow$) scores for different direct and cascaded speech translation models evaluated on the \textsc{Fleurs-Test} (FL; \textcolor{customgreen}{green}) and \textsc{BhasaAnuvaad-Test} (BA; \textcolor{customyellow}{yellow}) datasets in $En \rightarrow XX$ direction. The colors represent a heatmap, where darker shades indicating better translation performance.}
\label{tab:bleu_en_xx}

\bigskip

\begin{tabular}{ccccccccccccc}
\toprule
{\color[HTML]{31333F} }                            & \multicolumn{6}{c}{\textbf{Direct}}                                                                                                                                                                                                                                                                                               & \multicolumn{6}{c}{\textbf{Cascaded}}                                                                                                                                                                                                                                              \\ \cmidrule(lr){2-7} \cmidrule(lr){8-13}
\cellcolor[HTML]{FFFFFF}{\color[HTML]{31333F} }                                & \multicolumn{2}{c}{\textbf{SD}}                                                                           & \multicolumn{2}{c}{\textbf{SD\textsubscript{BA}}}                           & \multicolumn{2}{c}{\textbf{AZURE}}                                           & \multicolumn{2}{c}{\textbf{SC}}                               & \multicolumn{2}{c}{\textbf{W + IT2}}                          & \multicolumn{2}{c}{\textbf{SC + IT2}}                         \\ \cmidrule(lr){2-3} \cmidrule(lr){4-5} \cmidrule(lr){6-7} \cmidrule(lr){8-9} \cmidrule(lr){10-11} \cmidrule(lr){12-13}
\multirow{-3}{*}{\cellcolor[HTML]{FFFFFF}{\color[HTML]{31333F} \textit{lang}}} & \textbf{FL}                                     & \textbf{BA}                                         & \textbf{FL}               & \textbf{BA}                   & \textbf{FL}                              & \textbf{BA}                   & \textbf{FL}               & \textbf{BA}                   & \textbf{FL}               & \textbf{BA}                   & \textbf{FL}               & \textbf{BA}                   \\ \midrule
{\color[HTML]{0D0D0D} \textit{asm}}                & \cellcolor[HTML]{FFFFFF}{\color[HTML]{0D0D0D} 14.70} & \cellcolor[HTML]{FFF6DA}{\color[HTML]{0D0D0D} 21.40} & \cellcolor[HTML]{A9DDC4}22.40 & \cellcolor[HTML]{FFE9AB}29.50 & \cellcolor[HTML]{FFFFFF}12.80 & \cellcolor[HTML]{FFFDF6}16.70 & \cellcolor[HTML]{E7F6EF}17.00 & \cellcolor[HTML]{FFF2CD}23.60 & \cellcolor[HTML]{DDF1E7}17.90 & \cellcolor[HTML]{FFEEBD}26.40 & \cellcolor[HTML]{E9F6F0}16.90 & \cellcolor[HTML]{FFF2CD}23.60 \\
{\color[HTML]{0D0D0D} \textit{ben}}                & \cellcolor[HTML]{D7EFE3}{\color[HTML]{0D0D0D} 18.40} & \cellcolor[HTML]{FFF5D8}{\color[HTML]{0D0D0D} 21.80} & \cellcolor[HTML]{8BD0AF}27.40 & \cellcolor[HTML]{FFE089}35.20 & \cellcolor[HTML]{CBEADB}19.40 & \cellcolor[HTML]{FFFAEC}18.30 & \cellcolor[HTML]{B7E2CE}21.00 & \cellcolor[HTML]{FFF3CF}23.30 & \cellcolor[HTML]{AADDC5}22.20 & \cellcolor[HTML]{FFE9A9}29.80 & \cellcolor[HTML]{ABDDC5}22.00 & \cellcolor[HTML]{FFF3CF}23.30 \\
{\color[HTML]{0D0D0D} \textit{guj}}                & \cellcolor[HTML]{A7DCC2}{\color[HTML]{0D0D0D} 22.80} & \cellcolor[HTML]{FFEDB8}{\color[HTML]{0D0D0D} 27.20} & \cellcolor[HTML]{66C295}33.50 & \cellcolor[HTML]{FFDC7A}37.70 & \cellcolor[HTML]{CBEADB}19.40 & \cellcolor[HTML]{FFFFFF}12.60 & \cellcolor[HTML]{96D5B7}25.50 & \cellcolor[HTML]{FFE9AB}29.50 & \cellcolor[HTML]{89D0AD}27.70 & \cellcolor[HTML]{FFE18E}34.30 & \cellcolor[HTML]{8FD2B2}26.70 & \cellcolor[HTML]{FFE9AB}29.50 \\
{\color[HTML]{0D0D0D} \textit{hin}}                & \cellcolor[HTML]{C0E6D4}{\color[HTML]{0D0D0D} 20.30} & \cellcolor[HTML]{FFF0C3}{\color[HTML]{0D0D0D} 25.40} & \cellcolor[HTML]{67C295}33.40 & \cellcolor[HTML]{FFDC7B}37.50 & \cellcolor[HTML]{A3DAC0}23.40 & \cellcolor[HTML]{FFF1C7}24.70 & \cellcolor[HTML]{A1D9BE}23.70 & \cellcolor[HTML]{FFECB5}27.80 & \cellcolor[HTML]{8FD2B2}26.70 & \cellcolor[HTML]{FFE089}35.20 & \cellcolor[HTML]{8ED2B1}26.90 & \cellcolor[HTML]{FFECB5}27.80 \\
{\color[HTML]{0D0D0D} \textit{kan}}                & \cellcolor[HTML]{E4F4EC}{\color[HTML]{0D0D0D} 17.30} & \cellcolor[HTML]{FFF6DC}{\color[HTML]{0D0D0D} 21.10} & \cellcolor[HTML]{8AD0AE}27.60 & \cellcolor[HTML]{FFE498}32.70 & \cellcolor[HTML]{E1F3EB}17.50 & \cellcolor[HTML]{FFFFFD}15.40 & \cellcolor[HTML]{C6E8D8}19.80 & \cellcolor[HTML]{FFEFC1}25.70 & \cellcolor[HTML]{AEDEC7}21.80 & \cellcolor[HTML]{FFE6A0}31.30 & \cellcolor[HTML]{BDE5D2}20.50 & \cellcolor[HTML]{FFEFC1}25.70 \\
{\color[HTML]{0D0D0D} \textit{mal}}                & \cellcolor[HTML]{D9F0E5}{\color[HTML]{0D0D0D} 18.20} & \cellcolor[HTML]{FFFAE9}{\color[HTML]{0D0D0D} 18.80} & \cellcolor[HTML]{83CDA9}28.70 & \cellcolor[HTML]{FFE6A1}31.20 & \cellcolor[HTML]{D3EEE1}18.70 & \cellcolor[HTML]{FFFFFF}12.50 & \cellcolor[HTML]{ABDDC5}22.10 & \cellcolor[HTML]{FFF5D9}21.50 & \cellcolor[HTML]{9CD7BB}24.60 & \cellcolor[HTML]{FFEDBB}26.80 & \cellcolor[HTML]{A1D9BE}23.80 & \cellcolor[HTML]{FFF5D9}21.50 \\
{\color[HTML]{0D0D0D} \textit{mar}}                & \cellcolor[HTML]{D8EFE4}{\color[HTML]{0D0D0D} 18.30} & \cellcolor[HTML]{FFF6DB}{\color[HTML]{0D0D0D} 21.20} & \cellcolor[HTML]{86CEAB}28.30 & \cellcolor[HTML]{FFE394}33.40 & \cellcolor[HTML]{DEF2E8}17.80 & \cellcolor[HTML]{FFFFFF}14.10 & \cellcolor[HTML]{AEDEC7}21.80 & \cellcolor[HTML]{FFF1C7}24.70 & \cellcolor[HTML]{9CD7BB}24.50 & \cellcolor[HTML]{FFE8A8}29.90 & \cellcolor[HTML]{ABDDC5}22.00 & \cellcolor[HTML]{FFF1C7}24.70 \\
{\color[HTML]{0D0D0D} \textit{npi}}                & \cellcolor[HTML]{B3E0CA}{\color[HTML]{0D0D0D} 21.40} & \cellcolor[HTML]{FFFFFF}{\color[HTML]{0D0D0D} 14.80} & \cellcolor[HTML]{78C9A2}30.50 & \cellcolor[HTML]{FFE8A8}29.90 & \cellcolor[HTML]{C1E6D4}20.20 & \cellcolor[HTML]{FFFFFF}10.90 & \cellcolor[HTML]{9AD6B9}24.90 & \cellcolor[HTML]{FFFEF8}16.20 & \cellcolor[HTML]{91D3B3}26.50 & \cellcolor[HTML]{FFF4D5}22.20 & \cellcolor[HTML]{8DD1B0}27.00 & \cellcolor[HTML]{FFFEF8}16.20 \\
{\color[HTML]{0D0D0D} \textit{ory}}                & \cellcolor[HTML]{D8EFE4}{\color[HTML]{0D0D0D} 18.30} & \cellcolor[HTML]{FFF0C4}{\color[HTML]{0D0D0D} 25.20} & \cellcolor[HTML]{8CD1B0}27.20 & \cellcolor[HTML]{FFE18C}34.70 & \cellcolor[HTML]{C5E8D7}19.90 & \cellcolor[HTML]{FFF8E3}19.90 & \cellcolor[HTML]{A4DBC0}23.20 & \cellcolor[HTML]{FFE9A9}29.70 & \cellcolor[HTML]{94D4B5}25.90 & \cellcolor[HTML]{FFE394}33.40 & \cellcolor[HTML]{A0D9BD}24.00 & \cellcolor[HTML]{FFE9A9}29.70 \\
{\color[HTML]{0D0D0D} \textit{pan}}                & \cellcolor[HTML]{C0E6D4}{\color[HTML]{0D0D0D} 20.30} & \cellcolor[HTML]{FFF1CA}{\color[HTML]{0D0D0D} 24.10} & \cellcolor[HTML]{7AC9A3}30.30 & \cellcolor[HTML]{FFEAAC}29.20 & \cellcolor[HTML]{F8FDFB}15.60 & \cellcolor[HTML]{FFF5D9}21.60 & \cellcolor[HTML]{A1D9BE}23.70 & \cellcolor[HTML]{FFEAAF}28.80 & \cellcolor[HTML]{95D5B6}25.70 & \cellcolor[HTML]{FFF0C6}24.90 & \cellcolor[HTML]{95D5B6}25.70 & \cellcolor[HTML]{FFEAAF}28.80 \\
{\color[HTML]{0D0D0D} \textit{snd}}                & \cellcolor[HTML]{FFFFFF}{\color[HTML]{0D0D0D} 6.30}  & {\color[HTML]{0D0D0D} ----}                          & \cellcolor[HTML]{FCFEFD}15.30 & {\color[HTML]{0D0D0D} ----}   & {\color[HTML]{0D0D0D} ----}   & {\color[HTML]{0D0D0D} ----}   & \cellcolor[HTML]{FFFFFF}8.00  & {\color[HTML]{0D0D0D} ----}   & \cellcolor[HTML]{FFFFFF}12.70 & {\color[HTML]{0D0D0D} ----}   & \cellcolor[HTML]{FFFFFF}10.30 & {\color[HTML]{0D0D0D} ----}   \\
{\color[HTML]{0D0D0D} \textit{tam}}                & \cellcolor[HTML]{F9FDFB}{\color[HTML]{0D0D0D} 15.50} & \cellcolor[HTML]{FFFBEF}{\color[HTML]{0D0D0D} 17.80} & \cellcolor[HTML]{9ED8BC}24.20 & \cellcolor[HTML]{FFE9AB}29.50 & \cellcolor[HTML]{E1F3EB}17.50 & \cellcolor[HTML]{FFFFFD}15.50 & \cellcolor[HTML]{DAF0E6}18.10 & \cellcolor[HTML]{FFF6DC}21.00 & \cellcolor[HTML]{B9E3CF}20.90 & \cellcolor[HTML]{FFEFBF}26.00 & \cellcolor[HTML]{C7E9D9}19.70 & \cellcolor[HTML]{FFF6DC}21.00 \\
{\color[HTML]{0D0D0D} \textit{tel}}                & \cellcolor[HTML]{DAF0E6}{\color[HTML]{0D0D0D} 18.10} & \cellcolor[HTML]{FFF8E3}{\color[HTML]{0D0D0D} 19.90} & \cellcolor[HTML]{8BD0AF}27.40 & \cellcolor[HTML]{FFE6A0}31.30 & \cellcolor[HTML]{C8E9D9}19.60 & \cellcolor[HTML]{FFFEFB}15.80 & \cellcolor[HTML]{A7DCC2}22.80 & \cellcolor[HTML]{FFF3D1}22.90 & \cellcolor[HTML]{8ED2B1}26.90 & \cellcolor[HTML]{FFEAAD}29.10 & \cellcolor[HTML]{99D6B8}25.10 & \cellcolor[HTML]{FFF3D1}22.90 \\
{\color[HTML]{0D0D0D} \textit{urd}}                & \cellcolor[HTML]{E1F3EB}{\color[HTML]{0D0D0D} 17.50} & \cellcolor[HTML]{FFEDB9}{\color[HTML]{0D0D0D} 27.00} & \cellcolor[HTML]{8CD1AF}27.30 & \cellcolor[HTML]{FFD666}41.00 & \cellcolor[HTML]{D5EEE2}18.50 & \cellcolor[HTML]{FFF5D8}21.70 & \cellcolor[HTML]{BBE4D0}20.70 & \cellcolor[HTML]{FFE9AC}29.30 & \cellcolor[HTML]{A8DCC3}22.50 & \cellcolor[HTML]{FFDF86}35.60 & \cellcolor[HTML]{A1D9BE}23.80 & \cellcolor[HTML]{FFE9AC}29.30 \\
\textit{total}                                     & 17.67                                                & 21.98                                                & 27.39                         & 33.29                         & 18.48                         & 16.90                         & 20.88                         & 24.92                         & 23.32                         & 29.61                         & 22.46                         & 24.92                        \\\bottomrule
\end{tabular}
\caption{BLEU ($\uparrow$) scores for different direct and cascaded speech translation models evaluated on the \textsc{Fleurs-Test} (FL; \textcolor{customgreen}{green}) and \textsc{BhasaAnuvaad-Test} (BA; \textcolor{customyellow}{yellow}) datasets in $XX \rightarrow En$ direction. The colors represent a heatmap, where darker shades indicating better translation performance.}
\label{tab:bleu_xx_en}
\end{table*}


\end{document}